\documentclass{article}

\usepackage{microtype}
\usepackage{graphicx}
\usepackage{booktabs} %

\usepackage[hidelinks]{hyperref}

\usepackage[accepted]{icml2024}

\usepackage{amsmath}
\usepackage{amssymb}
\usepackage{mathtools}
\usepackage{amsthm}

\usepackage[capitalize,noabbrev]{cleveref}

\usepackage[utf8]{inputenc} %
\usepackage[T1]{fontenc}    %
\usepackage{url}            %

\usepackage{booktabs}       %
\usepackage{amsfonts}       %
\usepackage{nicefrac}       %
\usepackage{bm}
\usepackage{derivative}
\usepackage{subcaption}
\usepackage{tikzscale}
\usepackage{tabularx}
\usepackage{pgfplots}
\pgfplotsset{compat=1.12,
            label style={font=\scriptsize},
            tick label style={font=\tiny},  }
\usetikzlibrary{pgfplots.groupplots}
\usetikzlibrary{patterns}
\usetikzlibrary{positioning}
\usepackage[acronym,nohypertypes={acronym,notation}]{glossaries}
\newacronym{gns}{GNS}{Graph Network Simulator}
\newacronym{mgn}{MGN}{MeshGraphNet}
\newacronym{mpc}{MPC}{Model Predictive Control}
\newacronym{mbrl}{MBRL}{Model-Based Reinforcement Learning}
\newacronym{gnn}{GNN}{Graph Neural Network}
\newacronym{sofa}{SOFA}{Simulation Open Framework Architecture}
\newacronym{mpn}{MPN}{Message Passing Network}
\newacronym{mlp}{MLP}{Multilayer Perceptron}
\newacronym{cnn}{CNN}{Convolutional Neural Network}
\newacronym{mse}{MSE}{Mean Squared Error}
\newacronym{iou}{IoU}{Intersection over Union}
\newacronym{ggns}{GGNS}{Grounding Graph Network Simulator}
\newacronym{amber}{AMBER}{Adaptive Meshing By Expert Reconstruction}
\newacronym{pde}{PDE}{Partial Differential Equation}
\newacronym{gmm}{GMM}{Gaussian Mixture Model}
\newacronym{mp}{MP}{Movement Primitive}
\newacronym{fem}{FEM}{Finite Element Method}
\newacronym{amr}{AMR}{Adaptive Mesh Refinement}
\newacronym{amg}{AMG}{Adaptive Mesh Generation}
\newacronym{ml}{ML}{Machine Learning}
\newacronym{rl}{RL}{Reinforcement Learning}
\newacronym{il}{IL}{Imitation Learning}
\newacronym{dcd}{DCD}{Density-Aware Chamfer Distance}
\newacronym{unet}{U-Net}{U-Net}

\usepackage{xcolor}  %

\theoremstyle{plain}

\theoremstyle{definition}

\theoremstyle{remark}

\icmltitlerunning{Iterative Sizing Field Prediction for Adaptive Mesh Generation From Expert Demonstrations}

\begin{document}

\twocolumn[
\icmltitle{Iterative Sizing Field Prediction for Adaptive Mesh Generation From Expert Demonstrations}

\icmlsetsymbol{equal}{*}

\begin{icmlauthorlist}
\icmlauthor{Niklas Freymuth}{alr}
\icmlauthor{Philipp Dahlinger}{alr}
\icmlauthor{Tobias Würth}{fast}
\icmlauthor{Philipp Becker}{alr}
\icmlauthor{Aleksandar Taranovic}{alr}
\icmlauthor{Onno Grönheim}{evago}
\icmlauthor{Luise Kärger}{fast}
\icmlauthor{Gerhard Neumann}{alr}

\end{icmlauthorlist}

\icmlaffiliation{fast}{Institute of Vehicle Systems Technology, Karlsruhe Institute of Technology, Karlsruhe}
\icmlaffiliation{alr}{Autonomous Learning Robots, Karlsruhe Institute of Technology, Karlsruhe, Germany}
\icmlaffiliation{evago}{EVAGO GmbH, Leonberg, Germany}

\icmlcorrespondingauthor{Niklas Freymuth}{niklas.freymuth@kit.edu}

\icmlkeywords{Geometric Deep Learning, Graph Network Simulation, Meta-Learning}

\vskip 0.3in
]

\printAffiliationsAndNotice{}  %

\begin{abstract}
Many engineering systems require accurate simulations of complex physical systems. 
Yet, analytical solutions are only available for simple problems, necessitating numerical approximations such as the Finite Element Method (FEM). 
The cost and accuracy of the FEM scale with the resolution of the underlying computational mesh. 
To balance computational speed and accuracy meshes with adaptive resolution are used, allocating more resources to critical parts of the geometry. 
Currently, practitioners often resort to hand-crafted meshes, which require extensive expert knowledge and are thus costly to obtain.
Our approach, Adaptive Meshing By Expert Reconstruction (AMBER), views mesh generation as an imitation learning problem. 
AMBER combines a graph neural network with an online data acquisition scheme to predict the projected sizing field of an expert mesh on a given intermediate mesh, creating a more accurate subsequent mesh. 
This iterative process ensures efficient and accurate imitation of expert mesh resolutions on arbitrary new geometries during inference.
We experimentally validate AMBER on heuristic $2$D meshes and $3$D meshes provided by a human expert, closely matching the provided demonstrations and outperforming a single-step CNN baseline.

\end{abstract}

\glsunset{unet}

\section{Introduction}
\definecolor{gray}{RGB}{150, 150, 150}
\begin{figure*}[t]
    \centering

    \begin{tikzpicture}
        \definecolor{gray}{RGB}{150, 150, 150}
        \coordinate (A) at (-6.6, 2.65);
        \coordinate (B) at (0.04, -2.5);

        \draw[gray, ultra thick] (A) rectangle (B);
        \node (img1) at (-0.5,0) {\includegraphics[width=\textwidth]{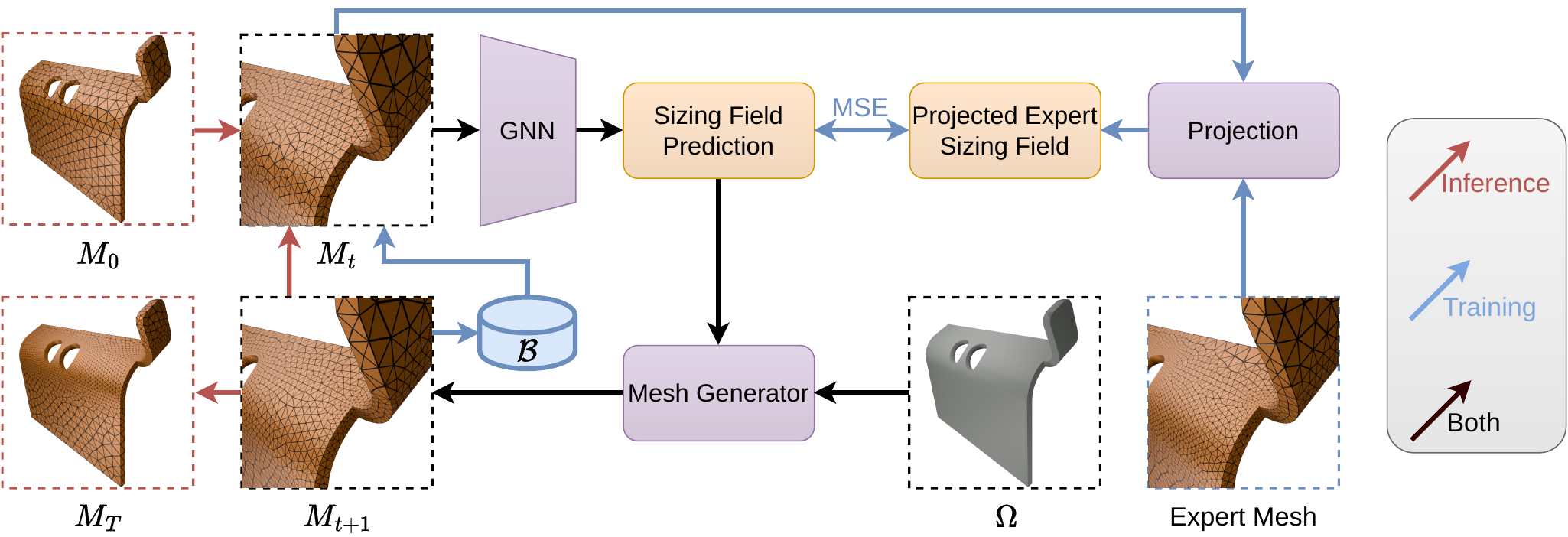}};
    \end{tikzpicture}
    \vspace{-0.5cm}
    \caption{Schematic overview of~\gls{amber}. 
    During \textcolor{purple}{inference},~\gls{amber} takes an initial mesh $M_0$, predicts a sizing field per element, and combines this with the underlying geometry in a mesh generator which produces an improved mesh. 
    This process is \textcolor{gray}{repeated} until a final mesh $M_T$ is obtained.
    When \textcolor{blue}{training},~\gls{amber} is tasked to predict the projected sizing field of an expert mesh for samples from a replay buffer, adding samples to this buffer to maintain a large and accurate distribution of training meshes.
    } 
    \label{fig:amber_schematic}
    \vspace{-0.2cm}
\end{figure*}

Simulating complex physical systems is an integral part of most engineering disciplines.
These simulations build on fundamental laws of physics, such as conservation of mass and energy, and are often expressed through~\glspl{pde}.
Due to their complexity, these \glspl{pde} are analytically intractable, requiring numerical approximations such as the~\gls{fem}~\citep{brenner2008mathematical, reddy2019introduction, anderson2021mfem}.
In the~\gls{fem}, a continuous problem geometry is partitioned into a mesh, i.e., a set of finite and simple elements. 
Such a mesh allows efficient approximations of the \gls{pde} solution, with cost and accuracy scaling with the number of elements.

For problems of realistic complexity, naively constructed meshes are often insufficient and adaptive meshes are required. 
Those can have varying element sizes allowing for a finer resolution in regions of interest and coarser resolution in areas that are simpler to simulate.
To find appropriate meshes \gls{amr} methods can refine existing meshes by allocating elements based on heuristics. 
Similarly, \gls{amg} creates meshes based on the problem geometry and initial conditions, often omitting the need for intermediate solutions.
Yet, both approaches are limited in efficiency and adaptability~\citep{mukherjee1996adaptive, kita2001error, yano2012optimization, cerveny2019nonconforming, wallwork2021mesh} and the required heuristics are often costly~\citep{giles2000introduction, giannakoglou2008adjoint}. 
Due to these challenges of existing tools, generating meshes for real-world applications still requires tedious manual labor and extensive domain knowledge. 

Several recent approaches have explored~\gls{amr} with machine learning.
One popular paradigm uses~\gls{rl}~\citep{sutton2018reinforcement} to frame~\gls{amr} as a sequential decision-making process~\citep{yang2023reinforcement, foucart2023deep, freymuth2023swarm}.
While promising, these~\gls{rl} approaches rely on application-specific and complex reward functions~\citep{foucart2023deep} or expensive, fine-grained uniform reference meshes~\cite{freymuth2023swarm}.
These factors, together with other constraints, such as a maximum refinement depth of the reference meshes~\cite{freymuth2023swarm} limit the practical applicability of these approaches.
Alternatively, recent supervised approaches aim to learn from expert meshes, using standard \glspl{cnn}~\citep{huang2021machine} or \glspl{mlp}~\citep{zhang2020meshingnet, lockPredictingNearOptimalMesh2024} to predict the target mesh.
However, these approaches are limited by the inherent properties of \glspl{cnn}, such as a fixed resolution or missing rotational equivariance.

In this work, we instead frame~\gls{amg} as an imitation learning problem, aiming to create a mesh that matches expert behavior by iteratively predicting a sizing field on an intermediate mesh to create a more accurate next mesh.
Our approach,~\gls{amber}, uses a~\gls{gnn} that consumes a graph representation of the current mesh and outputs a piecewise-constant sizing field on the mesh elements.
During inference, it then takes a geometry and a coarse initial mesh and iteratively improves this mesh by predicting a sizing field that is used to create the next mesh.
During training, this predicted sizing field is trained to be close to a sizing field that is projected from the expert mesh using a simple regression loss.
As the algorithm can create arbitrary intermediate meshes during inference, we maintain a replay buffer~\citep{lin1992self, fedus2020revisiting} during training, regularly adding meshes produced by our algorithm to this buffer and training on them.
Training on these buffered meshes instead of only using the initial meshes ensures there is no distribution shift between the simple regression during training and the iterative mesh generation procedure during inference.
We automatically label these intermediate meshes by projecting the expert sizing field onto them.
This process resembles the DAgger~\citep{ross2011reduction} in which an expert relabels out of distribution samples. 
In ~\gls{amber}, we implement a similar approach, however, we use an oracle for re-labeling. 
Figure~\ref{fig:amber_schematic} provides a schematic overview.

We experimentally validate our approach against a~\gls{cnn} baseline on heuristic expert meshes generated on a series of $2$D Poisson Equations on randomly generated L-Shaped geometries.
Additionally, we experiment on a set of complex real-world $3$D geometries with meshes created by a human expert.
We find that~\gls{amber} is particularly well-suited for highly adaptive meshes and produces accurate imitations of both heuristic and human expert behavior, regardless of the underlying mesh resolution.
We further show the effectiveness of~\gls{amber}'s individual design choices through a series of ablations.%
\footnote{Code and datasets are available at\\
~\quad\url{https://github.com/NiklasFreymuth/AMBER}.}

To summarize our contributions, we
\mbox{\textbf{(i)} propose}~\gls{amber}, a novel imitation learning approach that generates expert-like meshes through a series of sizing field predictions,
\mbox{\textbf{(ii)} introduce} challenging $2$D and $3$D datasets consisting of heuristic and human expert meshes and corresponding geometries, and
\mbox{\textbf{(iii)} evaluate} our method on these datasets, demonstrating high similarity to the provided expert meshes and outperforming a strong~\gls{cnn}-based baseline.
\section{Related Work}
\textbf{Meshing for Simulation.}
The~\glsreset{fem}\gls{fem} is a well-established method to numerically approximate physical systems, especially on complex, irregular geometric problem domains~\citep{brenner2008mathematical, reddy2019introduction}.
The~\gls{fem} solves a system of equations by dividing the geometry into a mesh consisting of smaller, simple elements.
Here, the simulation cost and accuracy directly depend on the mesh size, often necessitating an adaptive mesh that focuses more elements on important parts of the geometry for efficient simulations~\citep{plewa2005adaptive, huang2010adaptive}.  
Modern meshing approaches can be categorized into~\glsreset{amr}\gls{amr}~\citep{plewa2005adaptive, fidkowski2011review}, which adapts a given mesh, and~\glsreset{amg}\gls{amg}~\citep{yano2012optimization, remacle2013frontal, si2008adaptive}, which generates a new mesh from an estimated sizing field or related properties of the geometry.
These traditional~\gls{amr} approaches rely on heuristics~\citep{zienkiewicz1992superconvergent} or error estimates~\citep{nemecAdjointBasedAdaptiveMesh, bangerth2013adaptive}, which can quickly become inaccurate, unreliable or computationally expensive~\citep{bangerth2013adaptive, cerveny2019nonconforming, wallwork2021mesh}.

\textbf{Learning Based Approaches for Meshing.}
In recent years, \glspl{gnn}~\citep{bronstein2021geometric}, and particularly~\glspl{mpn}, have become popular architectures for learning to simulate on meshes.  
These~\glspl{gns} have been applied to mesh-based deformable~\citep{pfaff2020learning, linkerhagner2023grounding} and rigid~\citep{allen2022graph, allen2023learning, lopez2024scaling} object simulations.
We also employ~\glspl{mpn} acting on meshes to improve the speed and accuracy of physical simulations.
Yet, instead of directly learning to simulate, we generate application-specific efficient meshes for downstream classical~\gls{fem} solvers, which is more robust and safe as the actual simulation is done numerically.
Here, a recent suite of \gls{rl}-based \gls{amr} approaches has emerged for creating adaptive meshes~\citep{foucart2023deep, freymuth2023swarm, yang2023reinforcement}. 
These methods use an \gls{rl} policy to iteratively determine which mesh elements to subdivide. 
They typically depend on complex and costly reward functions, which need careful design to handle complex problems~\citep{freymuth2023swarm}. 
Despite this, the reward function requires a concrete underlying system of equations, restricts the maximum mesh resolution~\citep{yang2023reinforcement, freymuth2023swarm} or encodes a specific refinement criterion~\citep{foucart2023deep}.
These limitations are sub-optimal for real-world applications, where meshes often need to be highly locally refined, e.g., due to localized large gradients~\citep{larsonFiniteElementMethod2013}. 
We instead learn to refine based on expert meshes, omitting the need for a reward function and thus alleviating its limitations.

Other works consider supervised learning for mesh refinement. 
These approaches use recurrent neural networks to learn mesh refinement strategies~\citep{bohn2021recurrent},
optimize element stretching ratio and direction from a given error estimation~\citep{fidkowski2021metric}, and employ hand-crafted features to directly compute error estimates to solve an adjoint problem~\citep{roth2022neural, wallwork2022e2n}.
Further work trains a surrogate model with supervised learning for predicting the solution error~\citep{chen2021output, zhang2020meshingnet, zhangMeshingNet3DEfficientGeneration2021}, which in turn can be used for~\gls{amg} by setting the sizing field to the scaled inverse of the error estimation~\citep{zhangMeshingNet3DEfficientGeneration2021}.
Other approaches directly predict a sizing field using~\glspl{mlp} for simulations of magnetic devices~\citep{dyckDeterminingApproximateFinite1992a}, and~\glspl{cnn} for fluid dynamics problem~\citep{huang2021machine}.

\textbf{Interactive Imitation Learning.}
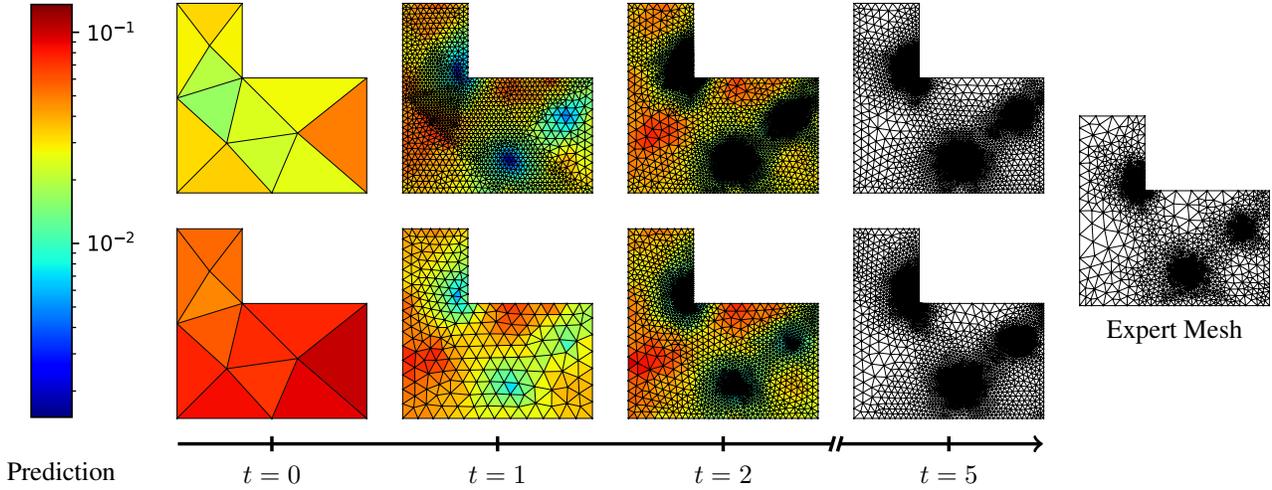
\begin{figure*}[t]
    \centering
    \begin{tikzpicture}
        \scalebox{1.0}{ 
        \node (pdf1) at (0, 4) {\includegraphics[width=3cm, height=3cm]{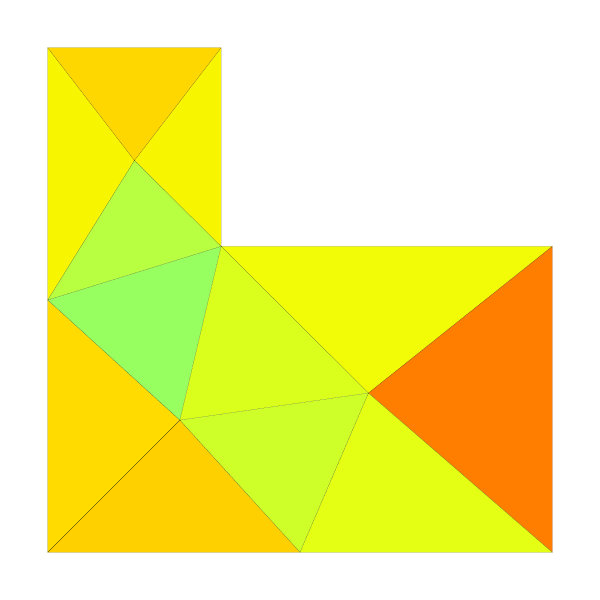}};
        \node (pdf2) at (3, 4) {\includegraphics[width=3cm, height=3cm]{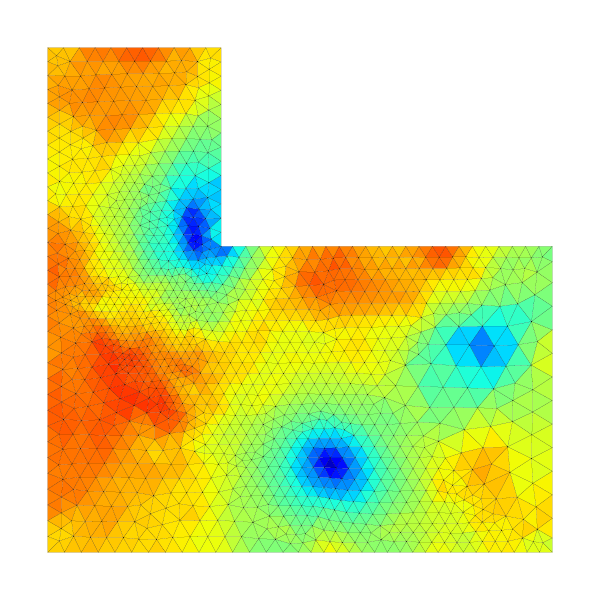}};
        \node (pdf3) at (6, 4) {\includegraphics[width=3cm, height=3cm]{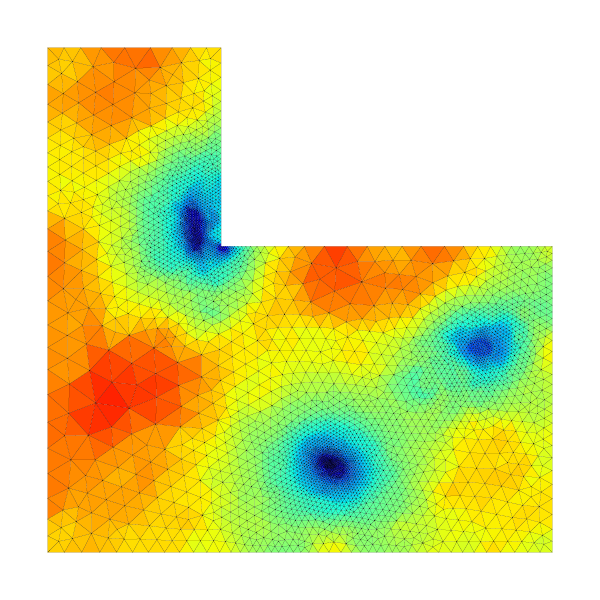}};
        \node (pdf4) at (9, 4) {\includegraphics[width=3cm, height=3cm]{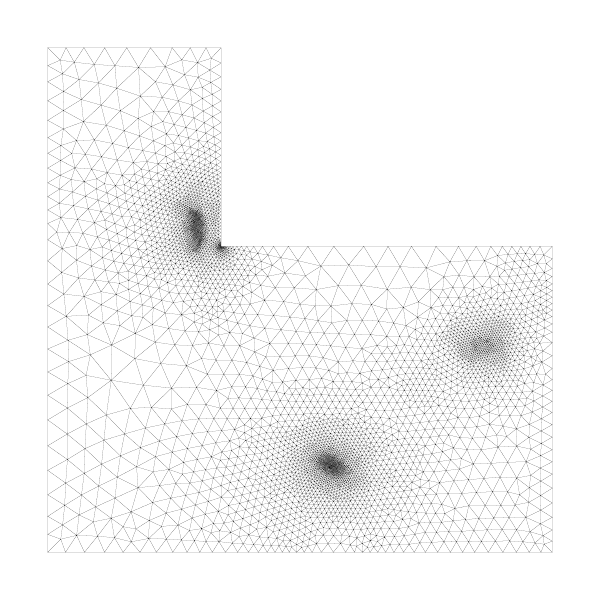}};

        \node (pdf5) at (0, 1) {\includegraphics[width=3cm, height=3cm]{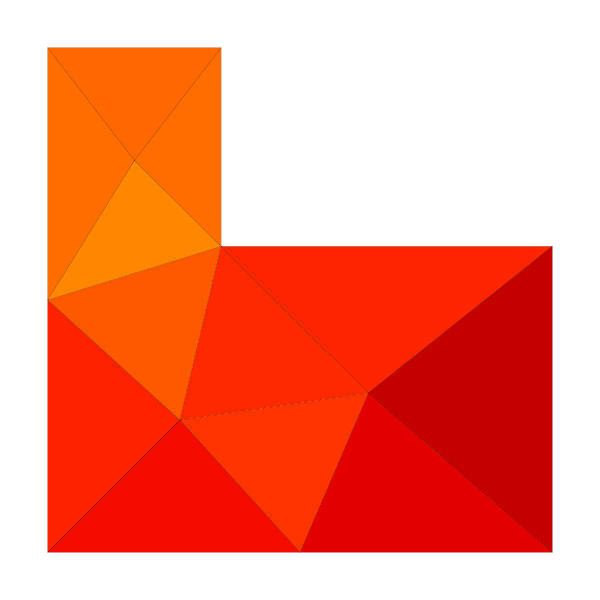}};
        \node (pdf6) at (3, 1) {\includegraphics[width=3cm, height=3cm]{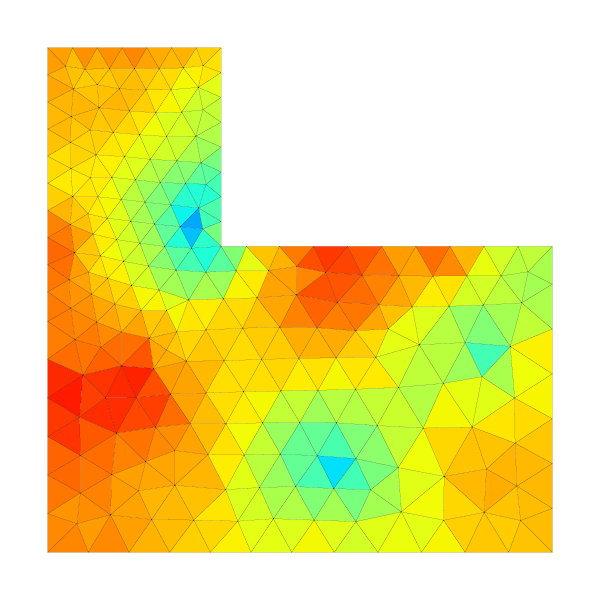}};
        \node (pdf7) at (6, 1) {\includegraphics[width=3cm, height=3cm]{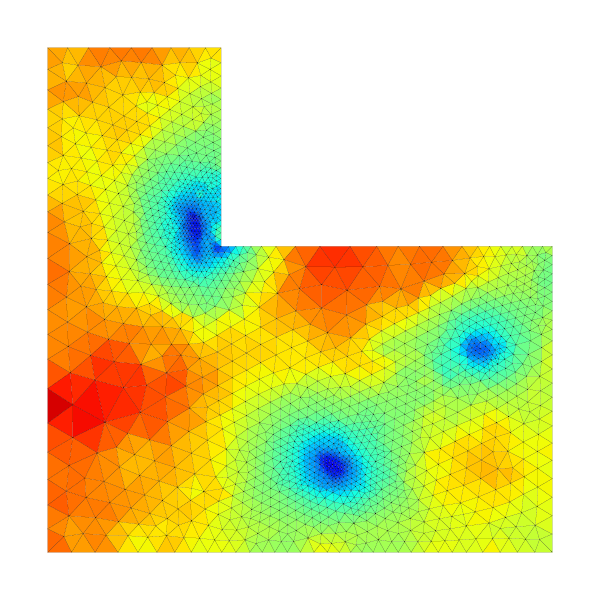}};
        \node (pdf8) at (9, 1) {\includegraphics[width=3cm, height=3cm]{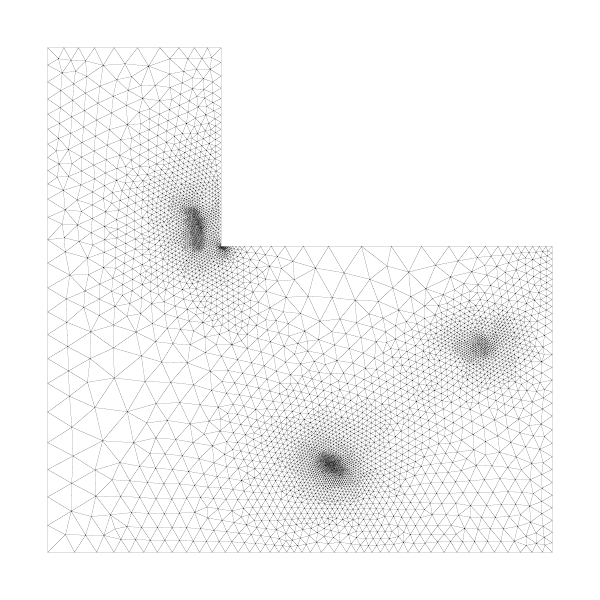}};

        \node[below=0.15cm of pdf5] {$t=0$};
        \node[below=0.15cm of pdf6] {$t=1$};
        \node[below=0.15cm of pdf7] {$t=2$};
        \node[below=0.15cm of pdf8] {$t=5$};

        \node (pdf9) at (12, 2.5) {\includegraphics[width=3cm, height=3cm]{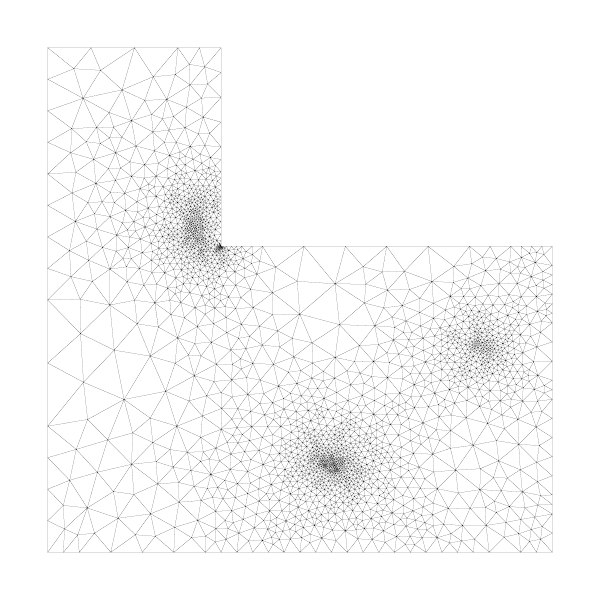}};
        \node[below=-0.3cm of pdf9] {Expert Mesh};

        \node (pdf10) at (-2.5, 2.5) {\includegraphics[width=2cm, height=6.1cm]{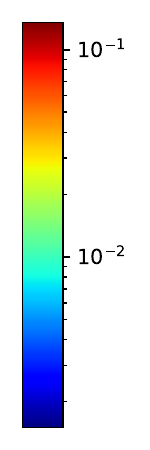}};
        \node[below=0.042cm of pdf10, xshift=-0.3cm] {Prediction};

        \draw[line width=1.3pt] (-1.25,-0.6) -- (7.45,-0.6);
        \draw[line width=1.3pt, ->] (7.55,-0.6) -- (10.25,-0.6);

        \draw[line width=1.3pt] (7.48,-0.5) -- (7.42,-0.7);
        \draw[line width=1.3pt] (7.58,-0.5) -- (7.52,-0.7);
        
        \foreach \x in {0,3,6,9}
            \draw[line width=1.3pt] (\x,-0.5) -- (\x,-0.7);
        
        }

    \end{tikzpicture}
    \caption{Intermediate and final~\gls{amber} meshes on Poisson's equation for an expert mesh with $50$ reference refinements. The colorbar on the left denotes the predicted sizing field per element for each intermediate mesh. 
    This prediction is given to a mesh generator to produce the next mesh. 
    \textbf{Top:} \gls{amber} (Mean) converges in a few generation steps at the cost of additional elements in intermediate steps.
    \textbf{Bottom:} \gls{amber} (Max) instead yields more conservative predictions, which take longer to converge but produce less total mesh elements. 
    In both cases, the mesh generator and the~\gls{mpn} architecture favor smooth solutions, generating a mesh that closely matches the expert but has less abrupt variations in local element size. 
    }
    \label{fig:qualitative_amber_mean_max}
\end{figure*}

\gls{il} is a common paradigm to learn behavior from expert demonstrations~\cite{pomerleau1988ALVINN, osa2018algorithmic, shafiullah2023BTransformer}, which is particularly applicable in scenarios where a reward function is difficult to define or can lead to undesired behavior~\citep{skalse22RewardHacking}.
A major challenge in \gls{il} occurs when the model is confronted with data absent from the expert data as it diverges from the expert behavior. 
Interactive \gls{il} methods, such as DAgger~\citep{ross2011reduction}, address this issue by allowing experts to intervene and label new data.
In some variants of this approach~\cite{menda2019ensembledagger,hoque22ThriftyDAgger}, the model actively requests new data from the expert when needed to improve efficiency further.
Similarly, \gls{amber} also actively acquires new data during learning.
However, instead of querying a human for new expert meshes, we maintain a static expert dataset and automatically generate and label new intermediate meshes based on the induced expert sizing fields.

\section{Adaptive Meshing by Expert Reconstruction (AMBER)}

\gls{amber} views mesh generation as an iterative imitation problem aiming to match an expert mesh based on a series of intermediate meshes.
In contrast to previous~\gls{amr} methods~\citep{zienkiewicz1992superconvergent, yang2023reinforcement, freymuth2023swarm}, we learn directly from meshes optimized for their respective downstream applications by experts.
This view greatly simplifies the mesh generation process, as our problem consists of matching an expert rather than optimizing for problem-specific and potentially complex adaptive mesh criteria~\citep{larsonFiniteElementMethod2013, huang2021machine,dolejsiAnisotropicHpMeshAdaptation2022}.
We approach this problem by training a~\gls{mpn} to iteratively predict sizing fields on a series of intermediate meshes, generating a new mesh from each prediction.
These projections are regressed against a projected expert sizing field, which is dynamically produced for meshes that are added to a replay buffer during training~\citep{ross2011reduction, kelly2019hgdagger}.
Figure~\ref{fig:amber_schematic} provides a schematic overview of~\gls{amber}, while the following sections describe its different aspects in more detail.

\subsection{Preliminaries}
\definecolor{redish}{RGB}{255, 116, 111}
\begin{figure*}[t]
    \centering
    \begin{tikzpicture}
        \node (img1) at (-0.5,0) {\includegraphics[width=0.29\textwidth]{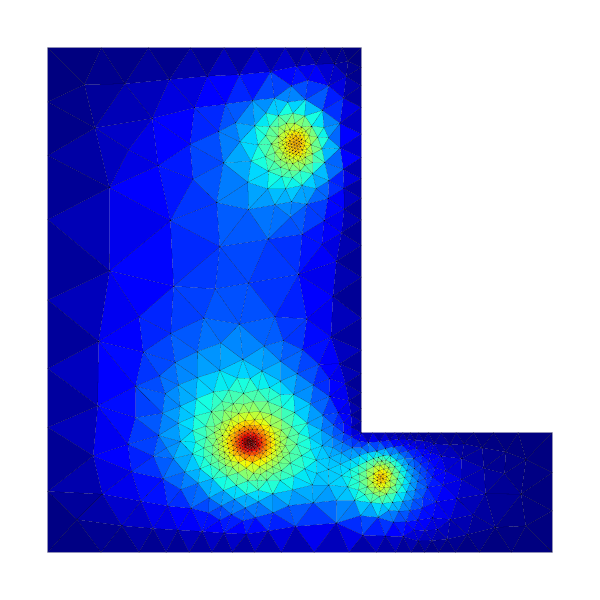}};
        \node[below=-0.2cm of img1] {Easy (25 Refs.)};
        
        \node (img2) at (4.0,0) {\includegraphics[width=0.29\textwidth]{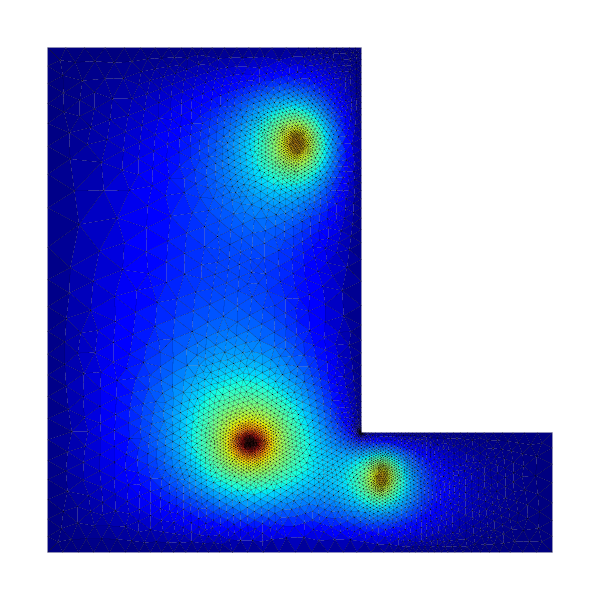}};
        \node[below=-0.2cm of img2] {Medium (50 Refs.)};
        
        \node (img3) at (8.5,0) {\includegraphics[width=0.29\textwidth]{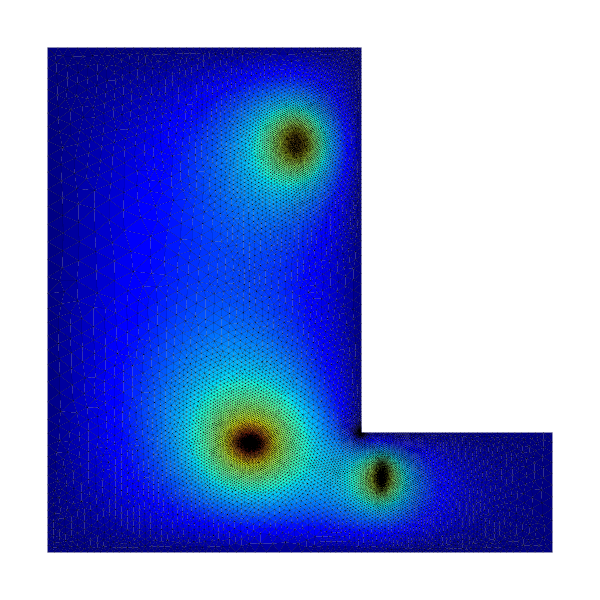}};
        \node[below=-0.2cm of img3] {Hard (75 Refs.)};

        \node (img4) at (11, 0.5) {\includegraphics[width=0.18\textwidth]{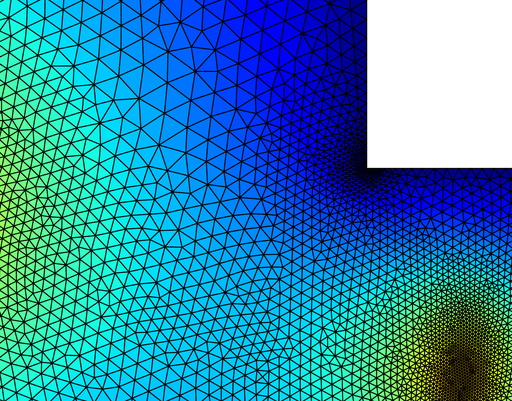}};

        \definecolor{redish}{RGB}{255, 116, 111}
        \coordinate (A) at (9.45, 1.7);
        \coordinate (B) at (12.55, -0.7);

        \coordinate (C) at (8.4, -.8);
        \coordinate (D) at (9.23, -1.5);
    
        \draw[redish, thick] (A) rectangle (B);
        \draw[redish, thick] (C) rectangle (D);
        
        \draw[redish, thick] (C) -- (A);

        \draw[redish, thick] (D) -- (B);
        
        \node (colorbar) at (-3,0) {\includegraphics[width=0.09\textwidth]
        {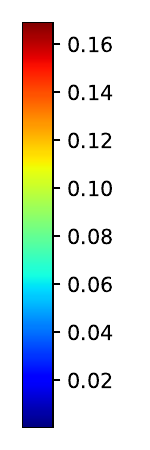}};
        \node[below=-0.045cm of colorbar, xshift=-0.35cm] {Solution};
    \end{tikzpicture}
    \caption{
    Exemplary \gls{amber} (Mean) refinements for Poisson's Equation trained on expert meshes with (\textbf{Left}) 25, (\textbf{Middle}) 50, and (\textbf{Right}) 75 refinement steps. 
    We plot the solution of Poisson's Equation as a color plot. The right figure includes a \textcolor{redish}{zoom} on the concave edge of the domain. }
    \label{fig:quantitative_amber_poisson_25_50_75}
    \vspace{-0.2cm}
\end{figure*}

We are given a training dataset $\mathcal{D}=\{(\Omega, M^*)\}_{n=1}^N$ consisting of pairs of a geometry $\Omega$ and a corresponding expert mesh $M^*$.
Each geometry describes a physical body in $2$D or $3$D, and the mesh discretizes this body into a number of simplical elements $M_j^*$.
We aim to generate a mesh $M^T$ that imitates the expert mesh $M^*$, i.e., that minimizes a distance $d(M^T, M^*)$ for unseen meshes during inference.

This generation is done using out-of-the-box approaches.
In this work, we use the Frontal Delaunay algorithm implemented in \textit{gmsh}~\citep{geuzaine2009gmsh}.
The mesh generator consumes a sizing field and a geometry and returns a mesh that conforms to this sizing field.
It additionally optimizes the mesh with respect to certain element criteria, such as the element's aspect ratio, which results in a smooth transition between element sizes.
In turn, the sizing field is a function $\Omega\rightarrow\mathbb{R}$ that describes the average desired edge length of the mesh's elements in space.
We can thus precisely a given mesh by the piece-wise constant sizing field that is induced by its elements.
Given the volume $V(M_i)$ of the $d$-dimensional simplical element $M_i$, its corresponding sizing field is given as the average edge length
\begin{equation*}
    f(M_i)=\left(V(M_i) \frac{d!}{\sqrt{d+1}}\right)^{\frac{1}{d}}\text{.}
\end{equation*}

To obtain a graph representation, we view the mesh elements as nodes $\mathcal{V}$ and their neighborhood relations as edges $\mathcal{E} \subseteq \mathcal{V} \times \mathcal{V}$. 
We then construct a bidirectional graph $\mathcal{G}_{\Omega^t} = \mathcal{G} = (\mathcal{V}, \mathcal{E}, \mathbf{x}_{\mathcal{V}}, \mathbf{x}_{\mathcal{E}})$ using node features $\mathbf{x}_{\mathcal{V}}$ and edge features $\mathbf{x}_{\mathcal{E}}$.
We process the graph using a~\gls{mpn}~\citep{pfaff2020learning}, a type of~\gls{gnn}~\citep{bronstein2021geometric}. 
\glspl{mpn} encompass the function class of several classical~\gls{pde} solvers~\citep{brandstetter2022message}, making them a popular choice for learning representations on meshes~\citep{pfaff2020learning, linkerhagner2023grounding, wurth2024physics}.
Appendix~\ref{app_sec:mpn} provides details for the~\gls{mpn} architecture.

\subsection{Mesh generation with AMBER}
When generating a mesh for a given geometry,~\gls{amber} starts with a coarse uniform initial mesh $M^t$ for $t=0$.
It then encodes the current mesh $M^t$ as a graph and feeds it through an~\gls{mpn} to predict a target size per mesh element.
The combination of these target sizes forms a piece-wise constant sizing field, which together with the underlying geometry is given to a mesh generator, which produces a new mesh $M^{t+1}$.
This process is iterated over $T$ steps, resulting in a final mesh $M^T$. 

This iteration is necessary, as the current mesh limits the resolution of the predicted sizing field in every step.
As we can only learn one target size for each current element of $M^t$, the corresponding finer resolution elements in $M^{t+1}$ will all have roughly the same size.
However, by iterating this process, the final mesh converges to an appropriate mesh, even for problems with highly non-uniform meshes.

\subsection{Training \gls{amber}}
\label{ssec:training}
\gls{amber}'s iterative mesh generation process necessitates a specialized training procedure.
To compute the target sizing field for an element $M_i^t\subseteq M^t$ of an intermediate mesh $M^t$, we first determine all expert elements $M_j^*\in M^*$ whose midpoints $p_{M_j^*}$ lie in $M_i^t$, and then aggregate over the sizing fields induced by these elements, i.e.,
\begin{equation*}
    y(M_i^t) = \bigotimes_{j} \mathbb{I}(p_{M_j^*}\in M_i^t) f(M_j^*)\text{.}
\end{equation*}
Here, $\bigotimes$ is a permutation-invariant aggregation function and we consider using either a maximum or a mean operator.
This results in two variants, called \gls{amber} (Mean) and~\gls{amber} (Max.).
Intuitively, we thus set each element's target element size to either the average or the maximum size of all of the expert elements it contains.
As the mean target size is always smaller or equal to the maximum target size, \gls{amber} (Mean) tends to refine more aggressively than \gls{amber} (Max) resulting in faster convergence while producing more and potentially unnecessary elements in the process.
In both cases, the target sizes for each element of the intermediate mesh will be smaller than their actual size, and iterating this process converges to the expert mesh.

If an element in $M^t$ to contains no midpoint $p_{M_j^*}$, we set its target sizing field to that of the element in the expert mesh that contains their midpoint.
Further, different meshes may approximate the same underlying geometry differently, leading to elements in one mesh that lie outside of the other.
We thus assign elements in the expert mesh that lie outside of the intermediate mesh to their nearest neighbor in the intermediate mesh.
This approximation ensures that all elements in the fine mesh contribute to the target sizing field of the intermediate mesh, and becomes more accurate the more closely the two meshes align.
Once the targets sizing field are obtained, we use them to train on a simple node-level~\gls{mse} loss.
This training scheme allows us to equally apply~\gls{amber} to both problem-specific meshes optimized for a given~\gls{pde}, and general-purpose meshes intended for different downstream analyses.

\definecolor{redish}{RGB}{255, 116, 111}
\definecolor{blueish}{RGB}{51, 130, 240}

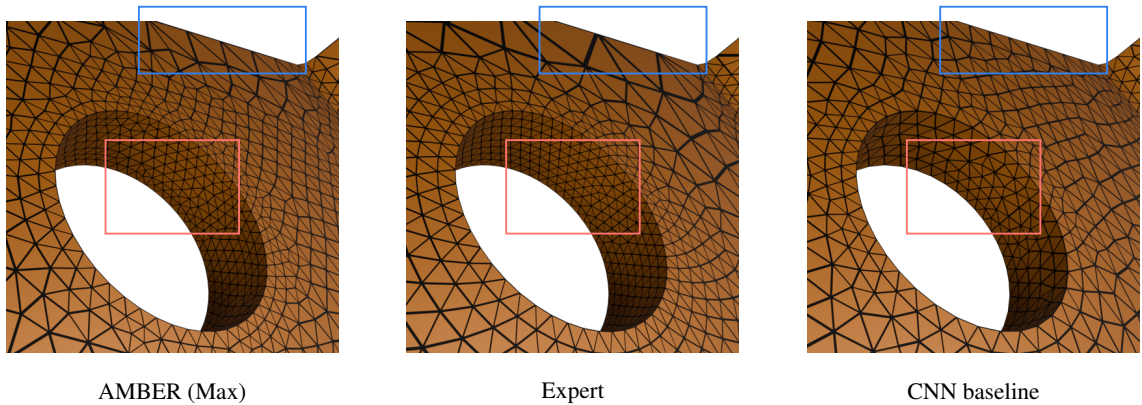
\begin{figure*}[t]
    \centering
    \resizebox{0.9\textwidth}{!}{
    \begin{tikzpicture}

        \node (img1) at (0,0) {\includegraphics[width=0.29\textwidth]{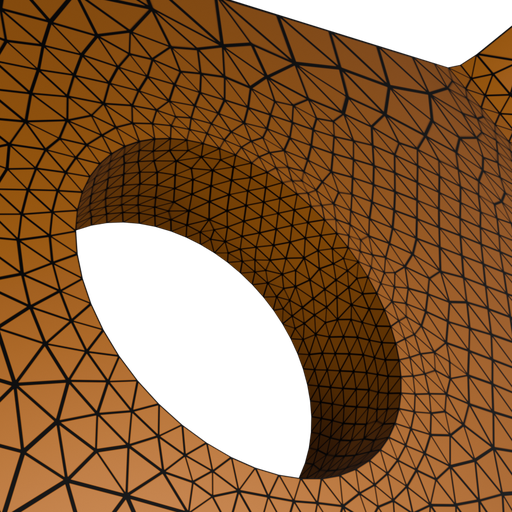}};
        \node[below=0.2cm of img1] {\gls{amber} (Max)};
        
        \node (img2) at (6,0) {\includegraphics[width=0.29\textwidth]{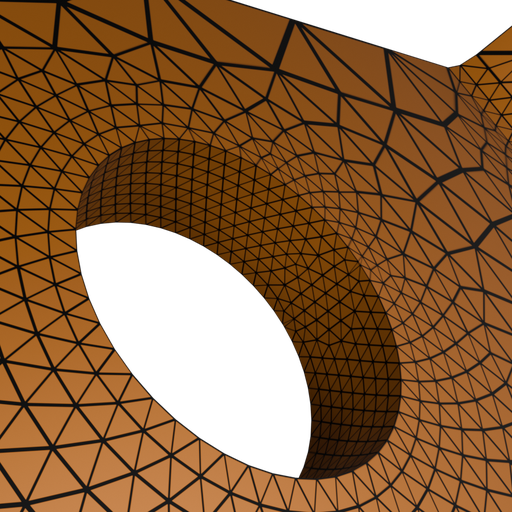}};
        \node[below=0.2cm of img2] {Expert};
        
        \node (img3) at (12,0) {\includegraphics[width=0.29\textwidth]{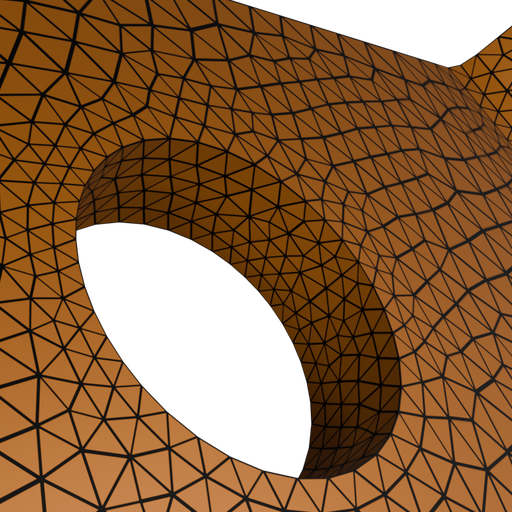}};
        \node[below=0.2cm of img3] {CNN baseline};
        
        \draw[redish, thick] (-1,-0.7) rectangle (1,0.7);
        \draw[redish, thick] (5,-0.7) rectangle (7,0.7);
        \draw[redish, thick] (11,-0.7) rectangle (13,0.7);

        \draw[blueish, thick] (-0.5, 1.7) rectangle (2.0, 2.7);
        \draw[blueish, thick] (5.5, 1.7) rectangle (8.0, 2.7);
        \draw[blueish, thick] (11.5, 1.7) rectangle (14.0, 2.7);
    \end{tikzpicture}
    }
    \caption{
        Zoom-in of a final generated meshes for an unseen test domain on the Console task.
        \gls{amber} (\textbf{Left}) closely matches the expert mesh (\textbf{Middle}), producing finer elements \textcolor{redish}{near the hole} and coarser elements \textcolor{blueish}{on the upper border} of the mesh. In comparison, the~\gls{cnn} baseline (\textbf{Right}), which acts on a $64\times 64\times 64$ $3$D image of the domain, has less variation in the element size and matches the expert less closely. 
    }
    \label{fig:qualitative_amber_expert_cnn}
    \vspace{-0.2cm}
\end{figure*}

\gls{amber} produces a series of intermediate meshes in an auto-regressive fashion during inference, but only trains on individual sizing field predictions. 
To prevent distribution shifts during the iterative mesh generation, we thus maintain a replay buffer~\citep{lin1992self, fedus2020revisiting} over \gls{amber}-generated intermediate meshes during training.
This replay buffer contains a fixed set of one coarse initial mesh per provided training sample.
Every $k$ training steps, we sample a mesh from the replay buffer, predict its projected sizing field, generate a new mesh from this prediction, project the expert field onto this new mesh to generate its labels, and store this new pair of meshes and labels in the buffer.
To stabilize the training process, we assign each intermediate mesh a `depth' that corresponds to the number of generating steps it has gone through, and store an equal amount of meshes for each depth up to a maximum.
Each training step consists of randomly drawing a number of meshes from the buffer. 
Since the meshes greatly vary in size, we fill a batch until a number of nodes is reached, rather than using a fixed number of graphs. 
Appendix~\ref{app_sec:training_details} provides details on the stratified replay buffer and the batch generation process. 

\section{Experiments}
\label{sec:exp_setup}

\textbf{Setup.}
To use an \gls{mpn} for meshes, we set the node features to the corresponding element's volume and the edge features $\mathbf{x}_{\mathcal{E}}$ are to the Euclidean distance between element midpoints.
\glspl{mpn} are permutation-equivariant by design and as we only provide Euclidean distances as spatial information, we additionally obtain equivariance to the Euclidean group~\citep{bronstein2021geometric}.
Since the sizing field's values are always positive we apply a softplus transformation to the~\gls{mpn} outputs, which greatly stabilizes training.
Similarly, we normalize the input graph features using the statistics of the replay buffer. 
Appendix~\ref{app_sec:hyperparameters} provides further details of the architecture, training procedure, and hyperparameters.

We repeat all methods for $4$ random seeds.
We evaluate on a set of unseen test geometries, and evaluate the quality of each generated mesh by comparing it to a reference expert mesh on this geometry.
For this comparison, we use the \gls{dcd}~\citep{wu2021density} over the sets of element midpoints of both meshes.
The \gls{dcd} is defined as a symmetric, exponentiated Chamfer distance with normalization factors to account for elements in one set that match multiple elements in the other.
Intuitively, it measures the closest exponential distance of every midpoint in one mesh to the other mesh, weighting this distance by the number of matches. 
It is bound in $[0,1]$, and $0$ if and only if both sets are the same. We provide a mathematical description of the \gls{dcd} in Appendix \ref{app_sec:additional_metrics}.
We also explore an alternate metric that numerically integrates the difference in element volumes between the two meshes.
\begin{figure*}[t]
\centering
    \begin{minipage}{0.86\textwidth}
    \input{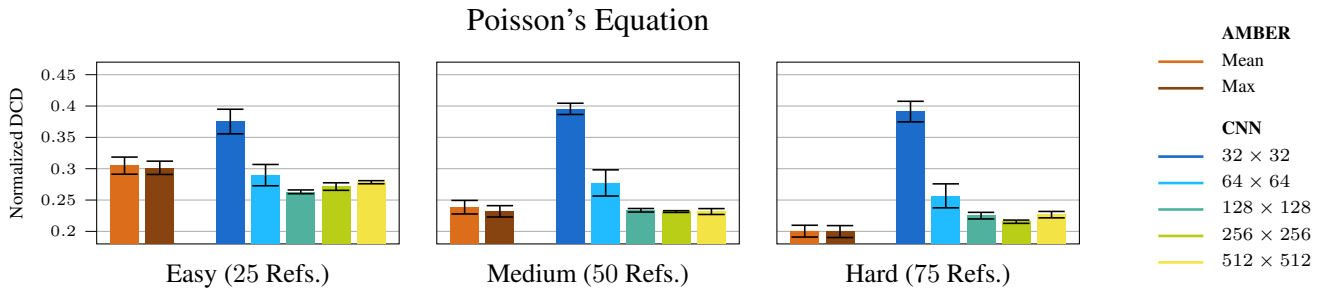}
    \end{minipage}%
    \begin{minipage}{0.03\textwidth}
    ~
    \end{minipage}%
    \begin{minipage}{0.10\textwidth}
        \begin{tikzpicture}

\tikzstyle{every node}'=[font=\scriptsize] 
\definecolor{color0}{RGB}{219,109,27}
\definecolor{color1}{RGB}{135,67,16}
\definecolor{color2}{RGB}{28,108,204}
\definecolor{color3}{RGB}{33,188,255}
\definecolor{color4}{RGB}{80,178,158}
\definecolor{color5}{RGB}{184,206,23}
\definecolor{color6}{RGB}{244,227,69}

\begin{axis}[%
hide axis,
xmin=10,
xmax=50,
ymin=0,
ymax=0.1,
legend style={
    draw=white!15!black,
    legend cell align=left,
    legend columns=1,
    legend style={
        draw=none,
        column sep=1ex,
        line width=1pt
    }
},
]

\addlegendimage{empty legend}
\addlegendentry{\textbf{\gls{amber}}}
\addlegendimage{color0}
\addlegendentry{Mean}
\addlegendimage{color1}
\addlegendentry{Max}
\addlegendimage{empty legend}
\addlegendentry{~}
\addlegendimage{empty legend}
\addlegendentry{\textbf{\gls{cnn}}}
\addlegendimage{color2}
\addlegendentry{$32\times32$}
\addlegendimage{color3}
\addlegendentry{$64\times64$}
\addlegendimage{color4}
\addlegendentry{$128\times128$}
\addlegendimage{color5}
\addlegendentry{$256\times256$}
\addlegendimage{color6}
\addlegendentry{$512\times512$}

\end{axis}
\end{tikzpicture}%
    \end{minipage}
    \vspace{-0.5\baselineskip}
    \caption{Mean and quantiles of the normalized~\gls{dcd} for~\gls{amber} and the~\gls{cnn} baseline for different input image resolutions. 
    (\textbf{Left}) Poisson Easy; the resulting meshes are comparatively coarse and can be reproduced well with the~\gls{cnn} baseline. 
    (\textbf{Middle}) Poisson Medium; the~\gls{cnn} starts to require a higher input resolution for good refinements.
    (\textbf{Right}) Poisson Hard; the expert mesh covers elements across multiple scales. Here, the~\gls{cnn} fails to provide good refinements for lower image resolutions and begins to overfit for higher resolutions. In contrast,~\gls{amber} directly acts on intermediate meshes and can thus dynamically adapt the sampling resolution of the sizing fields it predicts.}
    \label{fig:quantitative_poisson}
    \vspace{-0.2cm}
\end{figure*}

\textbf{Poisson's Equation.}
We consider problem-specific meshes using Poisson's Equation with a load function $f(\boldsymbol{x})$ and test function $v(\boldsymbol{x})$, given as
\begin{align*}
\int_\Omega \nabla u \cdot \nabla v\text{d}\boldsymbol{x} = \int_\Omega f v\text{d}\boldsymbol{x} \quad \forall v \in V\text{.}
\end{align*}
We enforce zero Dirichlet boundary conditions, i.e., \mbox{$u(\boldsymbol{x}) = 0$ on $\partial \Omega$}.
As the training data, we randomly generate L-shaped geoemtries and employ a Gaussian Mixture Model with three components for the load function. 
For this task, we evaluate the load function at each face's midpoint and use it as a node feature.
Additionally, we solve the equation for every intermediate mesh and use the mean and standard deviation of the solution on each element's vertices as additional node features. 
We generate the problem-specific expert meshes using an error-based refinement heuristic~\citep{binev2004adaptive, bangerth2012algorithms, foucart2023deep}, which marks all elements for refinement for which $\text{err}(M_i)>\theta\cdot\max_j \text{err}(M_j)$ for $0<\theta<1$.
The error is estimated heuristically based on the load function and gradient jumps on the element facets. 
We use $20$ randomly drawn systems of equations and corresponding expert meshes as the training data, and test on a disjoint, fixed set of $100$ unseen system of equations and expert meshes.
We vary the task's difficulty by adapting the number of refinement steps that the heuristic applies, considering \textit{easy}, \textit{medium}, and \textit{hard} variants with $25$, $50$, and $75$ refinement steps respectively.
Appendix \ref{app_sec:tasks_poisson} provides further details of the dataset creation and the expert heuristic.

\begin{figure*}[t]
\centering
\begin{minipage}{0.33\textwidth}
\begin{tikzpicture}

\definecolor{chocolate21910927}{RGB}{219,109,27}
\definecolor{darkgray176}{RGB}{176,176,176}
\definecolor{deepskyblue33188255}{RGB}{33,188,255}
\definecolor{indianred19152101}{RGB}{135,67,16}%
\definecolor{royalblue28108204}{RGB}{28,108,204}

\begin{axis}[
width=\textwidth,
height=4.0cm,
ytick={0.15,0.2,0.25,0.3,0.35,0.4,0.45},
xtick=\empty,
xticklabels=\empty, 
tick align=outside,
tick pos=left,
xmin=-0.64, xmax=4.64,
xminorgrids,
y grid style={darkgray176},
ylabel={Normalized DCD},
ymajorgrids,
ymin=0.13, ymax=0.47,
ytick style={color=black}
]
\draw[draw=none,fill=chocolate21910927] (axis cs:-0.4,0) rectangle (axis cs:0.4,0.195602414508215);
\draw[draw=none,fill=indianred19152101] (axis cs:0.6,0) rectangle (axis cs:1.4,0.238208319938986);
\draw[draw=none,fill=royalblue28108204] (axis cs:2.6,0) rectangle (axis cs:3.4,0.417795800583688);
\draw[draw=none,fill=deepskyblue33188255] (axis cs:3.6,0) rectangle (axis cs:4.4,0.309290379283527);
\path [draw=black, semithick]
(axis cs:0,0.191423891386997)
--(axis cs:0,0.199780937629433);

\addplot [semithick, black, mark=-, mark size=5, mark options={solid}, only marks]
table {%
0 0.191423891386997
};
\addplot [semithick, black, mark=-, mark size=5, mark options={solid}, only marks]
table {%
0 0.199780937629433
};
\path [draw=black, semithick]
(axis cs:1,0.219006328666106)
--(axis cs:1,0.257410311211866);

\addplot [semithick, black, mark=-, mark size=5, mark options={solid}, only marks]
table {%
1 0.219006328666106
};
\addplot [semithick, black, mark=-, mark size=5, mark options={solid}, only marks]
table {%
1 0.257410311211866
};
\path [draw=black, semithick]
(axis cs:3,0.400117576515817)
--(axis cs:3,0.435474024651559);

\addplot [semithick, black, mark=-, mark size=5, mark options={solid}, only marks]
table {%
3 0.400117576515817
};
\addplot [semithick, black, mark=-, mark size=5, mark options={solid}, only marks]
table {%
3 0.435474024651559
};
\path [draw=black, semithick]
(axis cs:4,0.298912243350461)
--(axis cs:4,0.319668515216594);

\addplot [semithick, black, mark=-, mark size=5, mark options={solid}, only marks]
table {%
4 0.298912243350461
};
\addplot [semithick, black, mark=-, mark size=5, mark options={solid}, only marks]
table {%
4 0.319668515216594
};
\end{axis}

\end{tikzpicture}
\end{minipage}%
\begin{minipage}{0.33\textwidth}
\vspace{0.75cm}
\begin{tikzpicture}

\definecolor{chocolate21910927}{RGB}{219,109,27}
\definecolor{darkgray176}{RGB}{176,176,176}
\definecolor{saddlebrown1356716}{RGB}{135,67,16}

\begin{axis}[
width=\textwidth,
height=4.0cm,
tick align=outside,
tick pos=left,
x grid style={darkgray176},
xlabel={Generation Step ($t$)},
xmajorgrids,
xtick={0,1,2,3,4},
xticklabels={1,2,3,4,5},
xmin=-0.2, xmax=4.2,
xtick style={color=black},
y grid style={darkgray176},
ylabel={Normalized DCD},
ymajorgrids,
ymin=0.174141703523589, ymax=0.438316493080463,
ytick style={color=black},
xlabel style={yshift=0.2cm} %
]
\path [draw=chocolate21910927, fill=chocolate21910927, opacity=0.2]
(axis cs:0,0.325897748780943)
--(axis cs:0,0.301318925994867)
--(axis cs:1,0.214400775349207)
--(axis cs:2,0.194831086989367)
--(axis cs:3,0.186149648503447)
--(axis cs:4,0.187498483350884)
--(axis cs:4,0.197563560444503)
--(axis cs:4,0.197563560444503)
--(axis cs:3,0.199688223936919)
--(axis cs:2,0.216576728307748)
--(axis cs:1,0.229420651516331)
--(axis cs:0,0.325897748780943)
--cycle;

\path [draw=saddlebrown1356716, fill=saddlebrown1356716, opacity=0.2]
(axis cs:0,0.426308548100605)
--(axis cs:0,0.404541261371501)
--(axis cs:1,0.256984663089314)
--(axis cs:2,0.229701576501596)
--(axis cs:3,0.216512322385496)
--(axis cs:4,0.212839931095232)
--(axis cs:4,0.233410450266216)
--(axis cs:4,0.233410450266216)
--(axis cs:3,0.237682021145895)
--(axis cs:2,0.250355631524335)
--(axis cs:1,0.277260374834712)
--(axis cs:0,0.426308548100605)
--cycle;

\addplot [very thick, chocolate21910927]
table {%
0 0.313608337387905
1 0.221910713432769
2 0.205703907648557
3 0.192918936220183
4 0.192531021897694
};
\addplot [very thick, saddlebrown1356716]
table {%
0 0.415424904736053
1 0.267122518962013
2 0.240028604012965
3 0.227097171765695
4 0.223125190680724
};
\end{axis}

\end{tikzpicture}
\end{minipage}%
\begin{minipage}{0.15\textwidth}
\begin{tikzpicture}

\tikzstyle{every node}'=[font=\scriptsize] 
\definecolor{color0}{RGB}{219,109,27}
\definecolor{color1}{RGB}{135,67,16}%
\definecolor{color2}{RGB}{28,108,204}
\definecolor{color3}{RGB}{33,188,255}

\begin{axis}[%
hide axis,
xmin=10,
xmax=50,
ymin=0,
ymax=0.1,
legend style={
    draw=white!15!black,
    legend cell align=left,
    legend columns=1,
    legend style={
        draw=none,
        column sep=1ex,
        line width=1pt
    }
},
]
\addlegendimage{empty legend}
\addlegendentry{\textbf{\gls{amber}}}
\addlegendimage{color0}
\addlegendentry{Mean}
\addlegendimage{color1}
\addlegendentry{Max}
\addlegendimage{empty legend}
\addlegendentry{\textbf{CNN}}
\addlegendimage{color2}
\addlegendentry{$32\times32\times32$}
\addlegendimage{color3}
\addlegendentry{$64\times64\times64$}

\end{axis}
\end{tikzpicture}%
\end{minipage}
\vspace{-0.5\baselineskip}
    \caption{
    Mean and quantiles of normalized~\gls{dcd} for~\gls{amber} and the~\gls{cnn} baseline for the Console task. 
    (\textbf{Left})~\gls{amber} naturally generalizes to the more complex $3$D domains of the console task, generating meshes that accurately match that of the human expert on unseen domains.
    In contrast, the~\gls{cnn} baseline does not scale well to $3$ dimensions, as evaluating the resulting $3$D image quickly becomes prohibitively expensive with an increased image resolution.
    (\textbf{Right})~For both \gls{amber} (Mean) and \gls{amber} (Max) mesh quality smoothly increases with more mesh generation steps.
    }
    \label{fig:quantitative_console}
    \vspace{-0.2cm}
\end{figure*}
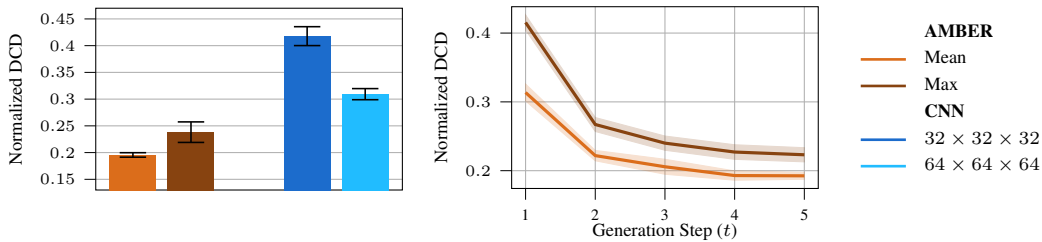

\textbf{Console Task.}
In the \textit{Console task}, we use data obtained from a real-world scenario in the automotive industry. 
We have a parameterized family of $3$D geometries representing a car's seat crossmembers.
The geometries are obtained using Onshape\footnote{\url{https://www.onshape.com/}} and feature various sharp bends as well as up to $2$ circular holes.
Tetrahedral meshes for this task are generated by human experts using ANSA\footnote{\url{https://www.beta-cae.com/ansa.htm}}.
The experts are initially presented with a coarse mesh, on which they iteratively select regions to refine, specifying the target element size of each selected region. 
The resulting meshes are optimized for downstream strength and durability analyses, but our experiments are conducted solely on the meshes and their underlying geometry.
The middle of Figure~\ref{fig:quantitative_console} provides an example close-up of an expert mesh.
The final dataset contains $22$ meshes and corresponding geometries. 
The meshes range from $6,284$ to $41,856$ elements and are randomly split into fixed sets of $15$ training, $2$ evaluation, and $5$ test meshes.

\textbf{Baselines and Ablations.}
We compare \gls{amber} to a \gls{cnn}-baseline that takes a binary image mask of the geometry and outputs a pixel-wise sizing field on the geometry.
This baseline is conceptually similar to~\citep{huang2021machine}, but with slight adaptations that improve the training stability and performance.
Instead of smoothing the expert sizing field and evaluating each pixel, we create an auxiliary mesh with roughly one element per pixel, calculate the target sizing field on this mesh as done in Section~\ref{ssec:training}, and map the resulting labels to each pixel.
Since the~\gls{cnn} does no iterative generation, we use the mean label projection method for each pixel.
Further, we train the approach on the logarithm of the sizing field, i.e., set the loss to
\begin{equation}
    \label{eq:cnn_loss}
    \mathcal{L}_{\text{CNN}}=\frac{1}{N}\sum_{i=1}^N\left(\text{CNN}(x_i)-\log(y_i)\right)^2\text{,}
\end{equation}
and subsequentially exponentiate the \gls{cnn}'s output to recover the predicted sizing field.
This change prevents larger mesh elements from dominating the~\gls{cnn}'s loss.
The~\gls{cnn} can only predict one target element size per pixel and we thus train separate~\glspl{cnn} on different image resolutions.
We also provide the same input features and normalization as for \gls{amber}.
Since the training sets are relatively small, we use all images for every training step, performing a total of $25,600$ training steps or equivalently epochs of training.

\section{Results}

\begin{figure}[t]
\centering
    \begin{tikzpicture}

\tikzstyle{every node}'=[font=\scriptsize] 

\definecolor{color10}{RGB}{219,109,27}
\definecolor{color11}{RGB}{135,67,16}
\definecolor{color12}{RGB}{191,52,101}
\definecolor{color13}{RGB}{115,22,131}
\definecolor{color14}{RGB}{28,108,204}
\definecolor{color15}{RGB}{33,188,255}
\definecolor{color16}{RGB}{135,67,16}
\definecolor{color17}{RGB}{80,178,158}
\definecolor{color18}{RGB}{184,206,23}

\definecolor{color2}{RGB}{28,108,204}
\definecolor{color3}{RGB}{33,188,255}
\definecolor{color4}{RGB}{80,178,158}
\definecolor{color5}{RGB}{184,206,23}
\definecolor{color6}{RGB}{244,227,69}

\begin{axis}[%
hide axis,
xmin=10,
xmax=50,
ymin=0,
ymax=0.1,
legend style={
    draw=white!15!black,
    legend cell align=left,
    legend columns=2,
    legend style={
        draw=none,
        column sep=1ex,
        line width=1pt
    }
},
]
\addlegendimage{empty legend}
\addlegendentry{\textbf{\gls{amber}}}
\addlegendimage{empty legend}
\addlegendentry{\textbf{CNN}}

\addlegendimage{color10}
\addlegendentry{Mean}
\addlegendimage{color5}
\addlegendentry{$256\times256$}

\addlegendimage{color11}
\addlegendentry{Max}
\addlegendimage{color4}
\addlegendentry{MSE Training, Softplus}

\addlegendimage{color12}
\addlegendentry{Mean, \gls{cnn} Loss}
\addlegendimage{color3}
\addlegendentry{MSE~\citep{huang2021machine}}

\addlegendimage{color13}
\addlegendentry{Max, \gls{cnn} Loss}

\addlegendimage{color2}
\addlegendentry{Max Aggregation}

\end{axis}
\end{tikzpicture}%

    \begin{minipage}{0.4\textwidth}
    \begin{tikzpicture}

\definecolor{black18}{RGB}{18,18,18}
\definecolor{cadetblue80178158}{RGB}{80,178,158}
\definecolor{chocolate21910927}{RGB}{219,109,27}
\definecolor{darkgray176}{RGB}{176,176,176}
\definecolor{darkgray153}{RGB}{153,153,153}
\definecolor{lightgray204}{RGB}{204,204,204}
\definecolor{royalblue28108204}{RGB}{28,108,204}
\definecolor{saddlebrown1356716}{RGB}{135,67,16}
\definecolor{deepskyblue33188255}{RGB}{33,188,255}
\definecolor{yellowgreen18420623}{RGB}{184,206,23}
\definecolor{indianred19152101}{RGB}{191,52,101}
\definecolor{plum223165229}{RGB}{223,165,229}
\definecolor{gray127}{RGB}{127,127,127}
\definecolor{sandybrown24422769}{RGB}{244,227,69}
\definecolor{crimson170085}{RGB}{170,0,85}
\definecolor{crimson198057}{RGB}{198,0,57}
\definecolor{crimson227028}{RGB}{227,0,28}
\definecolor{indigo850170}{RGB}{85,0,170}
\definecolor{mediumblue280227}{RGB}{28,0,227}
\definecolor{mediumblue570198}{RGB}{57,0,198}
\definecolor{purple1130142}{RGB}{113,0,142}
\definecolor{purple1420113}{RGB}{142,0,113}
\definecolor{purple11522131}{RGB}{115,22,131}

\begin{axis}[
width=\textwidth,
height=4.0cm,
ytick={0.2,0.3,0.4,0.5,0.6},
xtick=\empty,
xticklabels=\empty, 
tick align=outside,
tick pos=left,
xmin=-0.64, xmax=8.64,
xminorgrids,
xlabel=\normalsize{Poisson's Equation ($75$ Refs.)},
y grid style={darkgray176},
ylabel={Normalized DCD},
ymajorgrids,
ymin=0.13, ymax=0.62,
ytick style={color=black}
]

\path [draw=black, semithick]
(axis cs:0,0.19096313825436)
--(axis cs:0,0.209773773603126);
\addplot [semithick, black, mark=-, mark size=5, mark options={solid}, only marks]
table {%
0 0.19096313825436
};
\addplot [semithick, black, mark=-, mark size=5, mark options={solid}, only marks]
table {%
0 0.209773773603126
};

\path [draw=black, semithick]
(axis cs:1,0.190206800025923)
--(axis cs:1,0.209222463046725);
\addplot [semithick, black, mark=-, mark size=5, mark options={solid}, only marks]
table {%
1 0.190206800025923
};
\addplot [semithick, black, mark=-, mark size=5, mark options={solid}, only marks]
table {%
1 0.209222463046725
};

\path [draw=black, semithick]
(axis cs:2,0.22356201746585)
--(axis cs:2,0.232406056938884);
\addplot [semithick, black, mark=-, mark size=5, mark options={solid}, only marks]
table {%
2 0.22356201746585
};
\addplot [semithick, black, mark=-, mark size=5, mark options={solid}, only marks]
table {%
2 0.232406056938884
};

\path [draw=black, semithick]
(axis cs:3,0.222665824135626)
--(axis cs:3,0.231054612072501);
\addplot [semithick, black, mark=-, mark size=5, mark options={solid}, only marks]
table {%
3 0.222665824135626
};
\addplot [semithick, black, mark=-, mark size=5, mark options={solid}, only marks]
table {%
3 0.231054612072501
};

\draw[draw=none,fill=chocolate21910927] (axis cs:-0.4,0) rectangle (axis cs:0.4,0.200368455928743);
\draw[draw=none,fill=saddlebrown1356716] (axis cs:0.6,0) rectangle (axis cs:1.4,0.199714631536324);
\draw[draw=none,fill=indianred19152101] (axis cs:1.6,0) rectangle (axis cs:2.4,0.227984037202367);
\draw[draw=none,fill=purple11522131] (axis cs:2.6,0) rectangle (axis cs:3.4,0.226860218104063);

\draw[draw=none,fill=yellowgreen18420623] (axis cs:4.6,0) rectangle (axis cs:5.4,0.215415352850449);
\draw[draw=none,fill=cadetblue80178158] (axis cs:5.6,0) rectangle (axis cs:6.4,0.288895297176798);
\draw[draw=none,fill=deepskyblue33188255] (axis cs:6.6,0) rectangle (axis cs:7.4,0.520542706640327);
\draw[draw=none,fill=royalblue28108204] (axis cs:7.6,0) rectangle (axis cs:8.4,0.242330134428096);

\path [draw=black, semithick]
(axis cs:5,0.212815377980913)
--(axis cs:5,0.218015327719985);
\addplot [semithick, black, mark=-, mark size=5, mark options={solid}, only marks]
table {%
5 0.212815377980913
};
\addplot [semithick, black, mark=-, mark size=5, mark options={solid}, only marks]
table {%
5 0.218015327719985
};

\path [draw=black, semithick]
(axis cs:6,0.282263995753954)
--(axis cs:6,0.295526598599642);
\addplot [semithick, black, mark=-, mark size=5, mark options={solid}, only marks]
table {%
6 0.282263995753954
};
\addplot [semithick, black, mark=-, mark size=5, mark options={solid}, only marks]
table {%
6 0.295526598599642
};

\path [draw=black, semithick]
(axis cs:7,0.43784373970665)
--(axis cs:7,0.603241673574003);
\addplot [semithick, black, mark=-, mark size=5, mark options={solid}, only marks]
table {%
7 0.43784373970665
};
\addplot [semithick, black, mark=-, mark size=5, mark options={solid}, only marks]
table {%
7 0.603241673574003
};

\path [draw=black, semithick]
(axis cs:8,0.237994030392508)
--(axis cs:8,0.246666238463684);
\addplot [semithick, black, mark=-, mark size=5, mark options={solid}, only marks]
table {%
8 0.237994030392508
};
\addplot [semithick, black, mark=-, mark size=5, mark options={solid}, only marks]
table {%
8 0.246666238463684
};

\end{axis}

\end{tikzpicture}
    \end{minipage}%
    \vspace{-0.5\baselineskip}
    \caption{Mean and quantiles of normalized~\gls{dcd} for~\gls{amber} and the~\gls{cnn} baseline with a $256\times256$ resolution for different loss functions and sizing field interpolation types. 
    \gls{amber} performs worse when using the~\gls{cnn} loss of Equation~\ref{eq:cnn_loss}.
    In contrast, the~\gls{cnn} baseline yields worse meshes when trained on~\gls{amber}'s \gls{mse} loss.
    Omitting the softplus transformation of the predicted outputs results in a method similar to~\citep{huang2021machine}, but leads to a significantly less robust algorithm.
    Finally, using the maximum expert sizing field for the labels of the~\gls{cnn} baseline yields worse results as the target sizing file is too coarse. 
    }
    \label{app_fig:quantitative_ablation}
    \vspace{-0.2cm}
\end{figure}
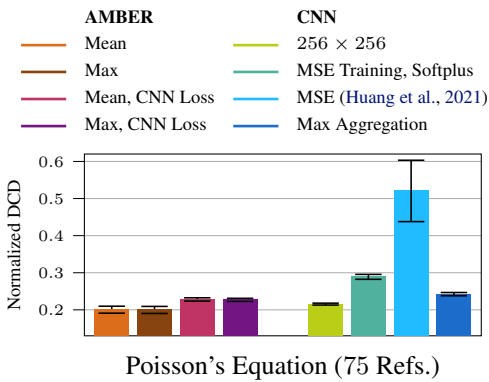

\textbf{Qualitative Results.}
Figure~\ref{fig:qualitative_amber_mean_max} visualizes exemplary~\gls{amber} refinements trained on the Poisson task with $50$ oracle heuristic refinement steps.
On the top row,~\gls{amber} (Mean) converges to a relatively accurate mesh after only $2$ prediction steps. 
However, the approach tends to locally over-refine some elements, causing potentially unnecessary refinements in intermediate meshes.
\gls{amber} (Max.) is shown on the bottom row. 
It produces fewer elements per step and thus requires more iterations until it converges to the expert mesh. 
Figure~\ref{fig:quantitative_amber_poisson_25_50_75} shows~\gls{amber} (Mean) refinements for the same test example when trained on expert meshes that use $25$, $50$ and $75$ heuristic refinement steps.
Our approach closely matches the expert heuristic, producing meshes of varying granularity depending on the expert meshes that it sees during training.
Figure~\ref{fig:qualitative_amber_expert_cnn} compares~\gls{amber} (Max) and the~\gls{cnn} baseline to an expert mesh on an exemplary test geometry on the Console task, finding that~\gls{amber} matches the expert's local element resolution well, while the~\gls{cnn} produces more uniform meshes and does not provide accurate sizing field predictions for the different geometric features.
Appendix~\ref{app_sec:extended_results} provides visualizations for~\gls{amber} and the~\gls{cnn} baseline for all tasks. 

\textbf{Quantitative Results.}
We measure mesh quality via the similarity to the expert mesh using the normalized \glsreset{dcd}\gls{dcd}~\cite{wu2021density} for~\gls{amber} for both mean and max sizing field interpolations and the~\gls{cnn} baseline for different image resolutions.
Figure~\ref{fig:quantitative_poisson} evaluates Poisson's Equation for $25$, $50$ and $75$ heuristic refinement steps of the expert mesh. 
While the~\gls{cnn} baseline works well for $25$ and even $50$ steps, its fixed image resolution eventually causes issues as the expert meshes become more complex.
In contrast,~\gls{amber} directly acts on intermediate meshes of arbitrary resolution.
Here, \gls{amber} works well on all $3$ task difficulties, as the method learns to adapt to the complexity of the expert's meshes.
This trend is more pronounced on the $3$D console task, as shown on the left side of Figure~\ref{fig:quantitative_console}.
\gls{amber} naturally generalizes to the more complex $3$D setting as it directly acts on meshes.
Thus, each~\gls{amber} forward pass is linear in the number of mesh elements.
In contrast, the~\gls{cnn} baseline quickly becomes computationally expensive due to cubic scaling with the number of voxels per dimension.
Finally, Figure~\ref{fig:quantitative_console} shows the benefits of~\gls{amber}'s iterative mesh generation process, finding that both~\gls{amber} (Mean) and~\gls{amber} (Max) benefit from up to $5$ intermediate mesh generation steps.
Appendix~\ref{app_sec:vdm} evaluates all tasks on an alternate metric that measures the difference in element volumes between the two meshes, yielding consistent results.

\textbf{Further Experiments.}
Figure~\ref{app_fig:quantitative_ablation} shows that \gls{amber} performs slightly worse when trained on the~\gls{cnn} loss (c.f. Equation~\ref{eq:cnn_loss}), likely because~\gls{amber} already implicitly weights small elements more strongly as there are more of them.
In contrast, the~\gls{cnn} baseline performs worse on~\gls{amber}'s~\gls{mse} loss, presumably because the same relative error makes a larger difference in mesh generation for pixels responsible for smaller elements when using this loss. 
Omitting the softplus transformation of the output on this loss, which results in a method similar to that of~\citep{huang2021machine}, further decreases stability and performance.
Finally, using a maximum operator to project the sizing fields for the~\gls{cnn} baseline leads to more conservative target estimation and thus too few elements, yielding worse results.
We explore additional variants and ablations of~\gls{amber} in Appendix~\ref{app_sec:extended_results} and find that~\gls{amber} benefits from additional data but provides strong generalization performance to unseen scenarios from as few as $5$ expert meshes on Poisson's Equation. 
Appendix~\ref{app_sec:extended_results} validates the effectiveness of normalizing the input features, using a softplus transformation to predict the strictly positive sizing field, and maintaining an equal number of meshes for each intermediate generation step instead of randomly adding samples.

\section{Conclusion}

We introduce~\glsreset{amber}\gls{amber}, an iterative adaptive mesh generation algorithm based on imitation learning.
\gls{amber} combines a Message Passing Graph Neural Network and a replay buffer with automatically labeled intermediate meshes to learn how to imitate meshes produced by experts. 
During mesh generation, \gls{amber} takes an arbitrary problem geometry and generates a series of increasingly accurate intermediate meshes by predicting a sizing field on each mesh and feeding this sizing field to an out-of-the-box mesh generator.
Rather than optimizing any particular metric, \gls{amber} learns to imitate an expert's meshing behavior from examples, enabling it to create suitable meshes for various engineering applications.

Experimental results on heuristic $2$D meshes on Poisson problems, and human expert meshes on real-world $3$D geometries demonstrate the strong performance of our approach.
\gls{amber} is data efficient and generates highly accurate meshes, significantly outperforming a strong CNN baseline when evaluated on more complex task setups.

\textbf{Limitations and Future Work.}
\gls{amber} requires a dataset with consistent expert behavior and always produces meshes that align with this single behavior.
We plan to extend \gls{amber} to allow conditioning the mesh generation process on a desired expert behavior, e.g. producing varying numbers of final elements with the same model~\cite{yang2023multi}. 
Additionally, we aim to replace the current piece-wise constant sizing field over the elements with a piece-wise linear sizing field over the mesh's vertices.
This change reduces the size of the mesh's graph representation while increasing the information provided in each sizing field.

\clearpage

\section*{Acknowledgements}
NF was supported by the BMBF project Davis (Datengetriebene Vernetzung für die ingenieurtechnische Simulation).
This work is also part of the DFG AI Resarch Unit 5339 regarding the combination of physics-based simulation with AI-based methodologies for the fast maturation of manufacturing processes. The financial support by German Research Foundation (DFG, Deutsche Forschungsgemeinschaft) is gratefully acknowledged. The authors acknowledge support by the state of Baden-Württemberg through bwHPC, as well as the HoreKa supercomputer funded by the Ministry of Science, Research and the Arts Baden-Württemberg and by the German Federal Ministry of Education and Research.

\bibliography{05_bibliography/icml2024}
\bibliographystyle{05_bibliography/icml2024}

\newpage
\appendix
\onecolumn
\section{Message Passing Networks}
\label{app_sec:mpn}

\glspl{mpn} iteratively update latent features for the graph nodes and edges over $L$ message-passing steps. 
Using linear embeddings $\mathbf{x}_v^0=\mathbf{x}_v$ and $\mathbf{x}_e^0=\mathbf{x}_e$ of the initial node and edge features, each step $l$ builds on the previous one, computing features
$$
\mathbf{x}^{l+1}_{e} = f^{l}_{\mathcal{E}}(\mathbf{x}^{l}_v, \mathbf{x}^{l}_u, \mathbf{x}^{l}_{e}), \textrm{ with } e = (u, v)\text{,}
$$
$$
\mathbf{x}^{l+1}_{v} = f^{l}_{\mathcal{V}}(\mathbf{x}^{l}_{v}, \bigoplus_{e=(v,u)\in \mathcal{E}} \mathbf{x}^{l+1}_{e})\text{.}
$$
The operator $\bigoplus$ is a permutation-invariant aggregation, such as sum, mean, or maximum operator. 
Each $f^l_\cdot$ is parameterized as a learned~\gls{mlp}. 
The final output of the~\gls{mpn} is a learned representation $\mathbf{x}^L_v$ for each node $v \in \mathcal{V}$.
We feed this representation into a decoder~\gls{mlp} to yield a scalar value per node that we train to approximate the target sizing field per mesh element.
Each $f^l_\cdot$ is a learned function, usually parameterized as a simple \gls{mlp}. The final output of the~\gls{mpn} is a learned representation $\mathbf{x}^L_v$ for each node $v \in \mathcal{V}$.

\section{AMBER Training details}
\label{app_sec:training_details}

\subsection{Replay Buffer}
To prevent memory issues during training, we only add new meshes to the buffer if their total number of elements does not exceed a threshold of $1.2$ times the largest expert mesh.
Additionally, we count the number of generation steps each intermediate mesh has undergone, setting a maximum depth in the buffer to prevent the replay buffer from eventually filling up with too-fine meshes.
We use this depth parameter to ensure an even distribution over mesh depths in the buffer by adding new meshes with a stratified mesh generation procedure.
Here, we first select a target depth uniformly at random, and then choose a random mesh from this depth to select the next mesh to refine and add to the buffer.

\subsection{Train Batch Construction}
As different meshes may have very different sizes, we do not train on a fixed number of meshes in each batch.
Instead, we define our batch size over the maximum number of nodes in a batch of graphs and fill up batches by iteratively adding a graph without replacement to this batch until the maximum number of nodes is reached.
To prevent smaller graphs from being sampled disproportionally often, we count the number of training steps each graph has been used for, and prefer the least used graphs in the batch construction if possible.

\section{Metrics} \label{app_sec:additional_metrics}
\subsection{Density-Aware Chamfer Metric} \label{app_sec:dcd}
This section describes our primary evaluation metric, the normalized \glsreset{dcd}\gls{dcd}, in more detail. 
The \gls{dcd} is defined as a symmetric, exponentiated Chamfer distance with normalization factors to account for elements in one set that match multiple elements in the other.

Mathematically, we first define the closest point function
\begin{equation*}
\hat{z}(z, S) = \arg \min_{w \in S} \|z - w\|_2\text{.}
\end{equation*}
Using this function, we define $n_{\hat{z}}$ as the number of matches that $z$ has in the complementary set $\overline{S}$:
\begin{equation*}
n_{\hat{z}} = |\{w \in \overline{S} \mid \hat{w}(w, S) = z\}|\text{.}
\end{equation*}
Given a point $z\in S$, let
\begin{equation*}
C(S, z, \overline{S}) = 1 - \frac{1}{n_{\hat{z}}} e^{-\|z - \hat{z}(z, \overline{S})\|_2},
\end{equation*}
be the contribution function.
The full metric is then defined as
\begin{equation*}
d_{DCD}(S_1, S_2)=\frac{1}{2} \left( \frac{1}{|S_1|} \sum_{x \in S_1} C(S_1, x, S_2) + \frac{1}{|S_2|} \sum_{y \in S_2} C(S_2, y, S_1) \right)\text{.}
\end{equation*}

To accommodate the inaccuracies of the mesh generator, we report a normalized version of this metric as
\begin{equation*}
d_{NDCD}(M^t, M^*)=\frac{d_{DCD}(\{p_{M^t_i}\}, \{p_{M^*_j}\}) - d_{DCD}(\{p_{M^0_i}\}, \{p_{M^*_j}\})}{d_{DCD}(\{p_{\hat{M}_i}\}, \{p_{M^*_j}\}) - d_{DCD}(\{p_{M^0_i}\}, \{p_{M^*_j}\})}\text{,}
\end{equation*}
where $\hat{M}$ is the reconstruction of $M^*$ that we obtain by directly feeding the sizing field of $M^*$ into the mesh generator.

\subsection{Volume Difference Metric}
\label{app_sec:vdm}
\begin{figure*}
\centering
    \begin{tikzpicture}

\tikzstyle{every node}'=[font=\scriptsize] 
\definecolor{color0}{RGB}{219,109,27}
\definecolor{color1}{RGB}{135,67,16}
\definecolor{color2}{RGB}{28,108,204}
\definecolor{color3}{RGB}{33,188,255}
\definecolor{color4}{RGB}{80,178,158}
\definecolor{color5}{RGB}{184,206,23}
\definecolor{color6}{RGB}{244,227,69}

\begin{axis}[%
hide axis,
xmin=10,
xmax=50,
ymin=0,
ymax=0.1,
legend style={
    draw=white!15!black,
    legend cell align=left,
    legend columns=9,
    legend style={
        draw=none,
        column sep=1ex,
        line width=1pt
    }
},
]
\addlegendimage{empty legend}
\addlegendentry{\textbf{\gls{amber}}$\hphantom{x^d}$}
\addlegendimage{color0}
\addlegendentry{Mean$\hphantom{x^d}$}
\addlegendimage{color1}
\addlegendentry{Max$\hphantom{x^d}$}
\addlegendimage{empty legend}
\addlegendentry{\textbf{CNN}$\hphantom{x^d}$}
\addlegendimage{color2}
\addlegendentry{$32^d$}
\addlegendimage{color3}
\addlegendentry{$64^d$}
\addlegendimage{color4}
\addlegendentry{$128^d$}
\addlegendimage{color5}
\addlegendentry{$256^d$}
\addlegendimage{color6}
\addlegendentry{$512^d$}

\end{axis}
\end{tikzpicture}%
    \input{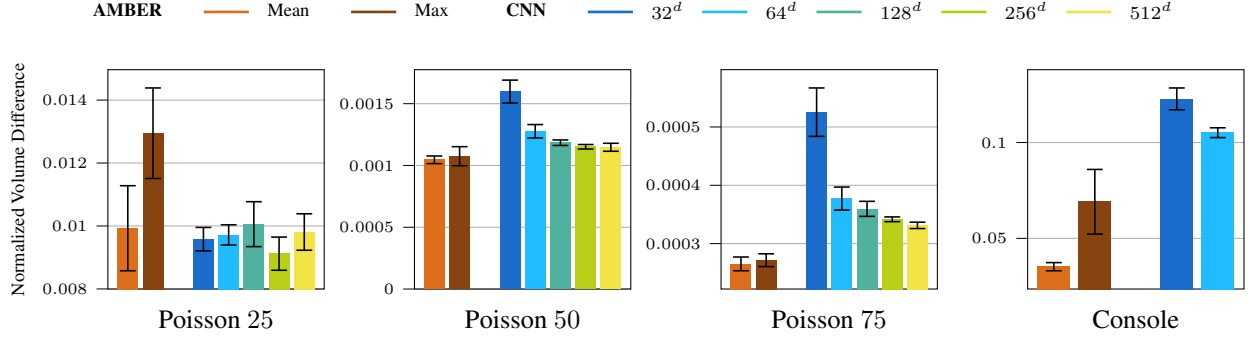}
    \caption{
    Normalized Volume Difference for~\gls{amber} and the~\gls{cnn} baseline for all considered tasks.
    The performance on this metric is largely consistent with that on the~\gls{dcd} across methods and tasks.
    }
    \label{app_fig:vol_difference}
\end{figure*}

In addition to the normalized~\gls{dcd} of Section~\ref{app_sec:dcd}, we evaluate the difference in element volumes between the generated mesh $M^T$ and the expert $M^*$.
Let $V(M^T_i)$ be the volume of element $M^T_i$, and $p_{M^T_i}$ be its midpoint. For an index $i$, we define $j^*(i)$ as the index of the element $M^*_{j^*(i)}$ which contains the midpoint $p_{M^T_i}$. Analogously, we set $j^T(i)$ as the index of the element $M^T_{j^T(i)}$ which contains the midpoint $p_{M^*_i}$.
We then calculate the volume difference between two meshes as
\begin{equation}
    \label{app_eq:vol_dif_metric}
    d_{\text{VD}} = \sum_{\{i \mid M_i^T \subseteq M^T\}} | V(M^T_i) - V(M^*_{j^*(i)}) | + \sum_{\{i \mid M^*_i \subseteq M^*\}} | V(M^*_i) - V(M^T_{j^T(i)}) | 
\end{equation}
Intuitively, this metric numerically approximates the integrated absolute difference in element volumes across the generated and the expert mesh using the elements of both meshes as integration points.

Figure~\ref{app_fig:vol_difference} evaluates the volume difference metric of Equation~\ref{app_eq:vol_dif_metric} for all tasks.
While the results for the volume difference metric have more variance than the normalized~\gls{dcd}, they are consistent with that of Figure~\ref{fig:quantitative_amber_poisson_25_50_75} and Figure~\ref{fig:quantitative_console}. 

\section{Tasks and Data Collection}
\label{app_sec:tasks_poisson}

For Poisson's equation, we generate random L-shaped geometries, defined as $(0,1)^2\backslash (p_0\times(1,1))$, where the lower left corner $p_0$ is sampled from $U(0.2, 0.95)$ in $x$ and $y$ direction. 
For the load function, we employ a Gaussian Mixture Model with three components. 
The mean of each component is drawn from $U(0.1, 0.9)^2$ and re-sampled until it is within the geometry. 
We then independently draw diagonal covariances from a log-uniform distribution $\exp(U(\log(0.0001, 0.001)))$ and randomly rotate them to yield a full covariance matrix. 
Component weights are generated from $\exp(N(0,1))+1$ and subsequently normalized.
For this task, we evaluate the load function at each face's midpoint and use it as a node feature.
Additionally, we solve the equation for every intermediate mesh, and use the mean and standard deviation of the solution on each element's vertices as additional node features.

We generate the expert meshes using an error-based heuristic on a refinement threshold $\theta$~\citep{binev2004adaptive, bangerth2012algorithms, foucart2023deep}, which marks all elements for refinement for which $\text{err}(M_i)>\theta\cdot\max_j \text{err}(M_j)$.
The error is estimated by a heuristic 
$$\text{err}(\Omega^t_i) = h^2 || f ||^2 + h || \left[\left[ \nabla u \cdot \mathbf{n} \right]\right] ||^2\text{,}$$
which assumes high errors in areas with high values of the load function $f$ and with large gradients jumps on the element edges or facets, respectively.
Once elements with a high error are marked, the mesh is processed by a remesher, which compute a conforming refined mesh using the red-green-blue refinement method~\citep{carstensen2004adaptive}.
We apply Laplacian smoothing after each refinement step.

\section{Network Architectures and Hyperparameters}
\label{app_sec:hyperparameters}

All methods are trained on an NVIDIA A100 GPU, with each method given a computational budget of up to $48$ hours.
For mesh generation, we clip all predicted sizing field to the minimum sizing field of any expert mesh in the training data.

The~\gls{mpn} of~\gls{amber} consists of $10$ separate message passing steps.
Each message passing step uses separate $2$-layer \glspl{mlp} and LeakyReLU activations for its node and edge updates.
We independently apply residual connections~\citep{he2016deep} and layer normalization~\citep{ba2016layer} to the node and edge updates. 
The final node features are fed into a $2$-layer \gls{mlp} decoder.
All~\glspl{mlp} use a latent dimension of $64$.
We provide an overview of~\gls{amber} hyperparameters in Table~\ref{tab:hyperparams}.

For the~\gls{cnn} baseline, we use a~\gls{unet}~\citep{ronneberger2015u} architecture with $32$ initial channels and $4$ down- and up-convolution blocks. 
Each convolution block consists of $2$ convolutions with a kernel size of $3$, followed by batch normalization and a ReLU activation function.
After each down-convolution, we use max-pooling with a kernel size and stride of $2$ to halve the image resolution, and double the number of channels.
This process is reversed for the up-convolutions, and we add skip connections between the respective depths.
We use $2$D and $3$D convolutions, batch normalization and pooling operations for the $2$D and $3$D tasks, respectively.

\begin{table}[ht]
\centering
\caption{AMBER hyperparameters and experiment configurations}
\begin{tabularx}{.5\textwidth}{l l}
\hline
Parameter & Value \\
\hline
Optimizer & ADAM \\
Learning rate & $3.0e$-$4$ \\
Latent dimension & $64$ \\
\gls{mpn} aggregation function & mean \\
\gls{mpn} steps & $10$ \\
\gls{mpn} activation function & Leaky ReLU \\
\gls{mpn} edge dropout & $0.1$ \\
\gls{mlp} layers & $2$ \\
Maximum replay buffer depth & $5$ \\
Maximum buffer size & $1\,000$ meshes \\
Training steps & $128\,000$ \\
Buffer addition frequency & every $8$ training steps \\
\gls{amber} batch size & $400\,000$ graph nodes \\
\gls{amber} inference steps & $5$ \\
\hline
\end{tabularx}
\label{tab:hyperparams}
\end{table}

\clearpage
\section{Extended Results}
\label{app_sec:extended_results}

\begin{figure*}[ht!]
\centering
    \begin{minipage}{0.60\textwidth}
    \begin{tikzpicture}

\definecolor{cadetblue80178158}{RGB}{80,178,158}
\definecolor{chocolate21910927}{RGB}{219,109,27}
\definecolor{darkgray176}{RGB}{176,176,176}
\definecolor{deepskyblue33188255}{RGB}{33,188,255}
\definecolor{indianred19152101}{RGB}{191,52,101}
\definecolor{purple11522131}{RGB}{115,22,131}
\definecolor{royalblue28108204}{RGB}{28,108,204}
\definecolor{saddlebrown1356716}{RGB}{135,67,16}
\definecolor{yellowgreen18420623}{RGB}{184,206,23}

\begin{axis}[
width=\textwidth,
height=4.5cm,
ytick={0.15,0.2,0.25,0.3,0.35,0.4,0.45},
xtick=\empty,
xticklabels=\empty, 
ytick style={draw=none}, 
tick align=outside,
title=\large{Poisson's Equation ($75$ Refs.)},
tick pos=left,
xmin=-0.64, xmax=4.64,
xminorgrids,
y grid style={darkgray176},
xlabel=\normalsize{\#Train Meshes},
ylabel={Normalized DCD},
ymajorgrids,
ymin=0.13, ymax=0.27,
ytick style={color=black}
]
\draw[draw=none,fill=royalblue28108204] (axis cs:-0.4,0) rectangle (axis cs:0.4,0.220502589619411);
\draw[draw=none,fill=deepskyblue33188255] (axis cs:0.6,0) rectangle (axis cs:1.4,0.213371808320843);
\draw[draw=none,fill=saddlebrown1356716] (axis cs:1.6,0) rectangle (axis cs:2.4,0.199714631536324);
\draw[draw=none,fill=cadetblue80178158] (axis cs:2.6,0) rectangle (axis cs:3.4,0.17538381266362);
\draw[draw=none,fill=yellowgreen18420623] (axis cs:3.6,0) rectangle (axis cs:4.4,0.168525640287167);
\path [draw=black, semithick]
(axis cs:0,0.214308455265938)
--(axis cs:0,0.226696723972884);

\addplot [semithick, black, mark=-, mark size=5, mark options={solid}, only marks]
table {%
0 0.214308455265938
};
\addplot [semithick, black, mark=-, mark size=5, mark options={solid}, only marks]
table {%
0 0.226696723972884
};
\path [draw=black, semithick]
(axis cs:1,0.201262009967182)
--(axis cs:1,0.225481606674504);

\addplot [semithick, black, mark=-, mark size=5, mark options={solid}, only marks]
table {%
1 0.201262009967182
};
\addplot [semithick, black, mark=-, mark size=5, mark options={solid}, only marks]
table {%
1 0.225481606674504
};
\path [draw=black, semithick]
(axis cs:2,0.190206800025923)
--(axis cs:2,0.209222463046725);

\addplot [semithick, black, mark=-, mark size=5, mark options={solid}, only marks]
table {%
2 0.190206800025923
};
\addplot [semithick, black, mark=-, mark size=5, mark options={solid}, only marks]
table {%
2 0.209222463046725
};
\path [draw=black, semithick]
(axis cs:3,0.168200041911114)
--(axis cs:3,0.182567583416126);

\addplot [semithick, black, mark=-, mark size=5, mark options={solid}, only marks]
table {%
3 0.168200041911114
};
\addplot [semithick, black, mark=-, mark size=5, mark options={solid}, only marks]
table {%
3 0.182567583416126
};
\path [draw=black, semithick]
(axis cs:4,0.16405253495103)
--(axis cs:4,0.172998745623304);

\addplot [semithick, black, mark=-, mark size=5, mark options={solid}, only marks]
table {%
4 0.16405253495103
};
\addplot [semithick, black, mark=-, mark size=5, mark options={solid}, only marks]
table {%
4 0.172998745623304
};
\end{axis}

\end{tikzpicture}
    \end{minipage}%
    \begin{minipage}{0.03\textwidth}
    ~
    \end{minipage}%
    \begin{minipage}{0.10\textwidth}
        \begin{tikzpicture}

\tikzstyle{every node}'=[font=\scriptsize] 
\definecolor{color0}{RGB}{219,109,27}
\definecolor{color1}{RGB}{135,67,16}
\definecolor{color2}{RGB}{191,52,101}
\definecolor{color3}{RGB}{115,22,131}
\definecolor{color4}{RGB}{28,108,204}
\definecolor{color5}{RGB}{33,188,255}
\definecolor{color6}{RGB}{135,67,16}
\definecolor{color7}{RGB}{80,178,158}
\definecolor{color8}{RGB}{184,206,23}

\begin{axis}[%
hide axis,
xmin=10,
xmax=50,
ymin=0,
ymax=0.1,
legend style={
    draw=white!15!black,
    legend cell align=left,
    legend columns=1,
    legend style={
        draw=none,
        column sep=1ex,
        line width=1pt
    }
},
]
\addlegendimage{empty legend}
\addlegendentry{\textbf{\#Meshes}}
\addlegendimage{color4}
\addlegendentry{$5$}
\addlegendimage{color5}
\addlegendentry{$10$}
\addlegendimage{color6}
\addlegendentry{$20$ (Default)}
\addlegendimage{color7}
\addlegendentry{$50$}
\addlegendimage{color8}
\addlegendentry{$100$}

\end{axis}
\end{tikzpicture}%
    \end{minipage}
    \caption{Normalized~\gls{dcd} for~\gls{amber} (Max) on Poisson's Equation with $75$ refinement steps for different numbers of training meshes. 
    A larger number of training meshes consistently improves performance, yet even a training set size of $5$ yields strong generalization performance to unseen meshes.
    }
    \label{fig:quantitative_num_train_meshes}
\end{figure*}
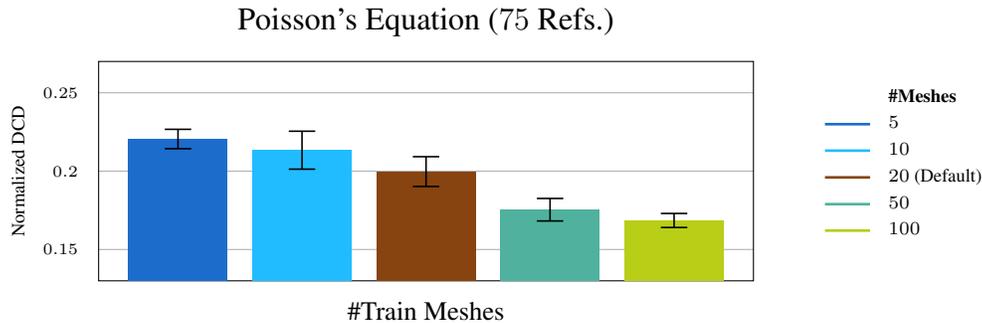

\begin{figure*}[ht!]
\centering
    \begin{minipage}{0.57\textwidth}
    \begin{tikzpicture}

\definecolor{darkgray176}{RGB}{176,176,176}
\definecolor{indianred19152101}{RGB}{191,52,101}
\definecolor{plum223165229}{RGB}{223,165,229}
\definecolor{purple11522131}{RGB}{115,22,131}
\definecolor{saddlebrown1356716}{RGB}{135,67,16}

\begin{axis}[
width=\textwidth,
height=4.0cm,
ytick={0.15,0.2,0.25},
xtick=\empty,
xticklabels=\empty, 
tick align=outside,
title=\large{Poisson's Equation ($75$ Refs.)},
tick pos=left,
xmin=0.36, xmax=4.64,
xminorgrids,
y grid style={darkgray176},
xlabel=\normalsize{Design Choices},
ylabel={Normalized DCD},
ymajorgrids,
ymin=0.13, ymax=0.27,
ytick style={color=black}
]
\draw[draw=none,fill=saddlebrown1356716] (axis cs:0.6,0) rectangle (axis cs:1.4,0.199714631536324);
\draw[draw=none,fill=plum223165229] (axis cs:1.6,0) rectangle (axis cs:2.4,0.233282092927681);
\draw[draw=none,fill=indianred19152101] (axis cs:2.6,0) rectangle (axis cs:3.4,0.214612559112516);
\draw[draw=none,fill=purple11522131] (axis cs:3.6,0) rectangle (axis cs:4.4,0.226315569200811);
\path [draw=black, semithick]
(axis cs:1,0.190206800025923)
--(axis cs:1,0.209222463046725);

\addplot [semithick, black, mark=-, mark size=5, mark options={solid}, only marks]
table {%
1 0.190206800025923
};
\addplot [semithick, black, mark=-, mark size=5, mark options={solid}, only marks]
table {%
1 0.209222463046725
};
\path [draw=black, semithick]
(axis cs:2,0.202926402106909)
--(axis cs:2,0.263637783748453);

\addplot [semithick, black, mark=-, mark size=5, mark options={solid}, only marks]
table {%
2 0.202926402106909
};
\addplot [semithick, black, mark=-, mark size=5, mark options={solid}, only marks]
table {%
2 0.263637783748453
};
\path [draw=black, semithick]
(axis cs:3,0.204469983363259)
--(axis cs:3,0.224755134861774);

\addplot [semithick, black, mark=-, mark size=5, mark options={solid}, only marks]
table {%
3 0.204469983363259
};
\addplot [semithick, black, mark=-, mark size=5, mark options={solid}, only marks]
table {%
3 0.224755134861774
};
\path [draw=black, semithick]
(axis cs:4,0.22204701496041)
--(axis cs:4,0.230584123441212);

\addplot [semithick, black, mark=-, mark size=5, mark options={solid}, only marks]
table {%
4 0.22204701496041
};
\addplot [semithick, black, mark=-, mark size=5, mark options={solid}, only marks]
table {%
4 0.230584123441212
};
\end{axis}

\end{tikzpicture}
    \end{minipage}%
    \begin{minipage}{0.03\textwidth}
    ~
    \end{minipage}%
    \begin{minipage}{0.10\textwidth}
        \begin{tikzpicture}

\tikzstyle{every node}'=[font=\scriptsize] 
\definecolor{color0}{RGB}{135,67,16}
\definecolor{color1}{RGB}{223,165,229}
\definecolor{color2}{RGB}{191,52,101}
\definecolor{color3}{RGB}{115,22,131}

\begin{axis}[%
hide axis,
xmin=10,
xmax=50,
ymin=0,
ymax=0.1,
legend style={
    draw=white!15!black,
    legend cell align=left,
    legend columns=1,
    legend style={
        draw=none,
        column sep=1ex,
        line width=1pt
    }
},
]
\addlegendimage{empty legend}
\addlegendentry{\textbf{\gls{amber} (Max)}}
\addlegendimage{color0}
\addlegendentry{Ours}
\addlegendimage{color1}
\addlegendentry{No Pred. Transform}
\addlegendimage{color2}
\addlegendentry{Random Buffer Sampling}
\addlegendimage{color3}
\addlegendentry{No Observation Norm.}

\end{axis}
\end{tikzpicture}%
    \end{minipage}
    \caption{
    Normalized~\gls{dcd} for~\gls{amber} (Max) on Poisson's Equation with $75$ refinement steps for different algorithmic design choices.
    Either Omitting the softplus output transformation for the sizing field prediction or not normalizing the input features yields worse results.
    Randomly adding intermediate training meshes to the replay buffer instead of adding the same amount of intermediate meshes for each mesh generation step decreases performance.
    }
    \label{app_fig:quantitative_amber_buffer_transform}
\end{figure*}
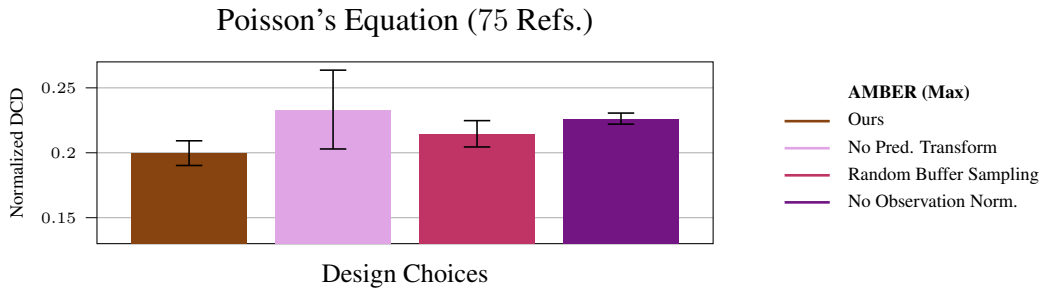

We explore the data efficiency of~\gls{amber} in Figure~\ref{fig:quantitative_num_train_meshes}.
While the approach continuously improves its predictions when provided with more training data, as few as $5$ training meshes and corresponding geometries are sufficient for accurate mesh generation.
This generalization capabilities are likely a results of the local, per-node~\gls{mse} loss and the~\gls{mpn} network architecture.
Figure~\ref{app_fig:quantitative_amber_buffer_transform} explores design choices of~\gls{amber}'s training scheme.
Both input normalization and transforming the output predictions improve performance. 
The stratified sampling of new training data for~\gls{amber}'s replay buffer improves performance, presumably because this ensures an even distribution of training data across mesh generation iterations, whereas randomly adding new meshes eventually under-represents the early meshes in the training data.

\subsection{Visualizations}
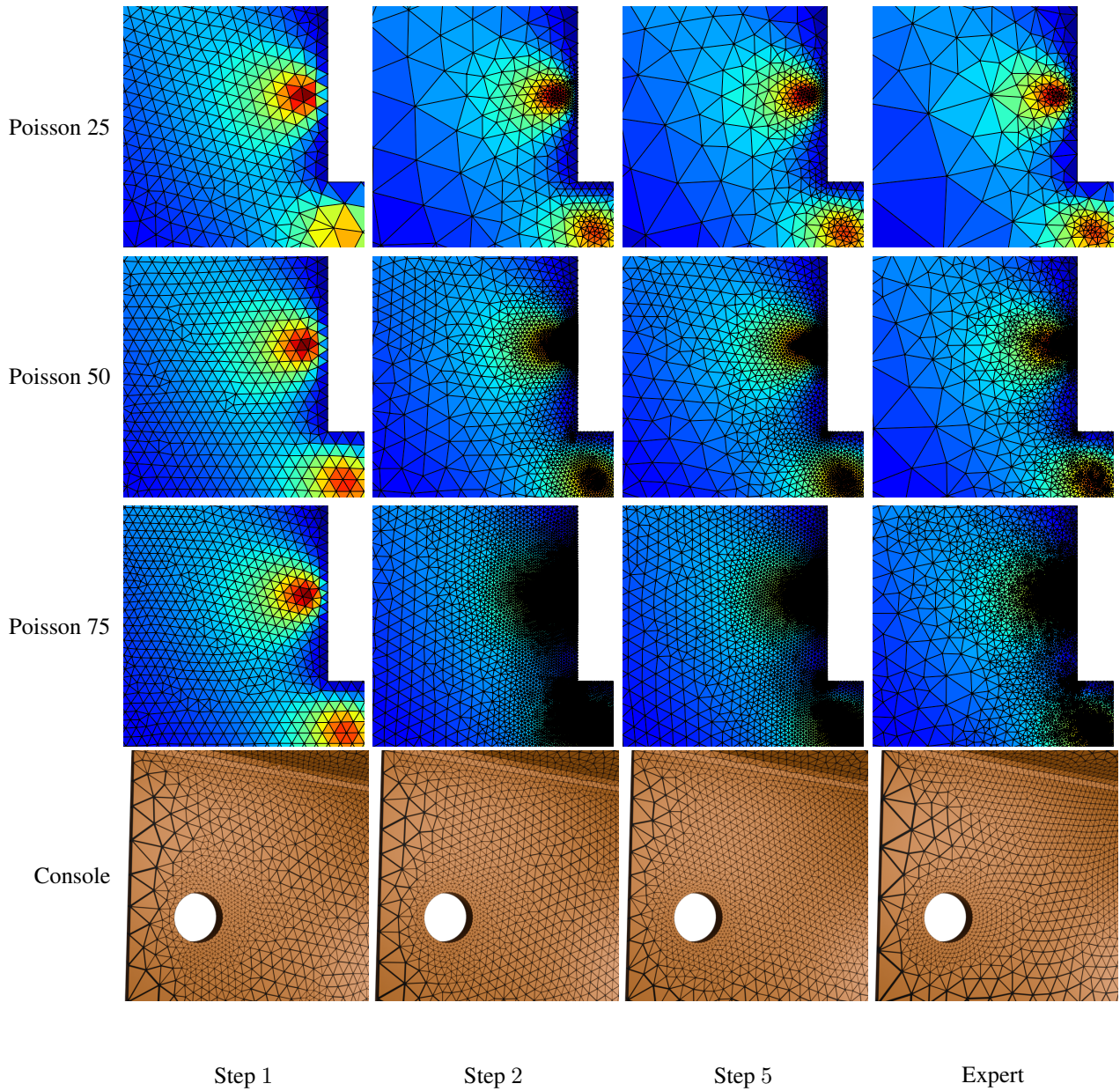
\begin{figure}
    \centering
    \begin{tikzpicture}
    \def\imagewidth{0.22\textwidth}
    \def\imageheight{0.22\textwidth}

    \node[anchor=east] at (0.5*\imagewidth, -0.5*\imageheight + 0.5*\imageheight) {Poisson 25};
    \node[anchor=east] at (0.5*\imagewidth, -1.5*\imageheight + 0.5*\imageheight) {Poisson 50};
    \node[anchor=east] at (0.5*\imagewidth, -2.5*\imageheight + 0.5*\imageheight) {Poisson 75};
    \node[anchor=east] at (0.5*\imagewidth, -3.5*\imageheight + 0.5*\imageheight) {Console};

    \node[anchor=north] at (1*\imagewidth, -4*\imageheight + 1cm) {Step $1$};
    \node[anchor=north] at (2*\imagewidth, -4*\imageheight + 1cm) {Step $2$};
    \node[anchor=north] at (3*\imagewidth, -4*\imageheight + 1cm) {Step $5$};
    \node[anchor=north] at (4*\imagewidth, -4*\imageheight + 1cm) {Expert};

    \node at (1*\imagewidth, 0*\imageheight) {\includegraphics[width=\imagewidth, height=\imageheight]{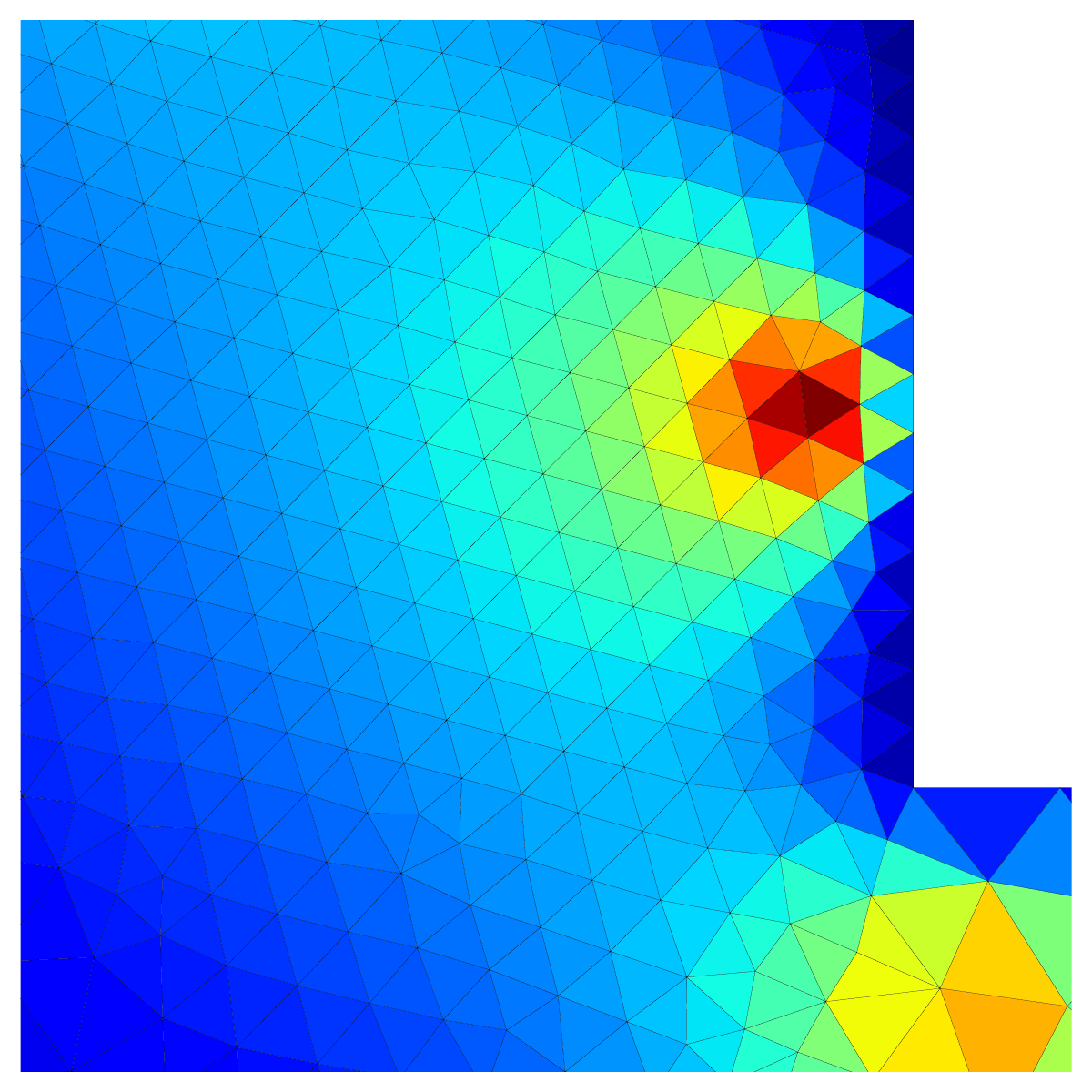}};
    \node at (2*\imagewidth, 0*\imageheight) {\includegraphics[width=\imagewidth, height=\imageheight]{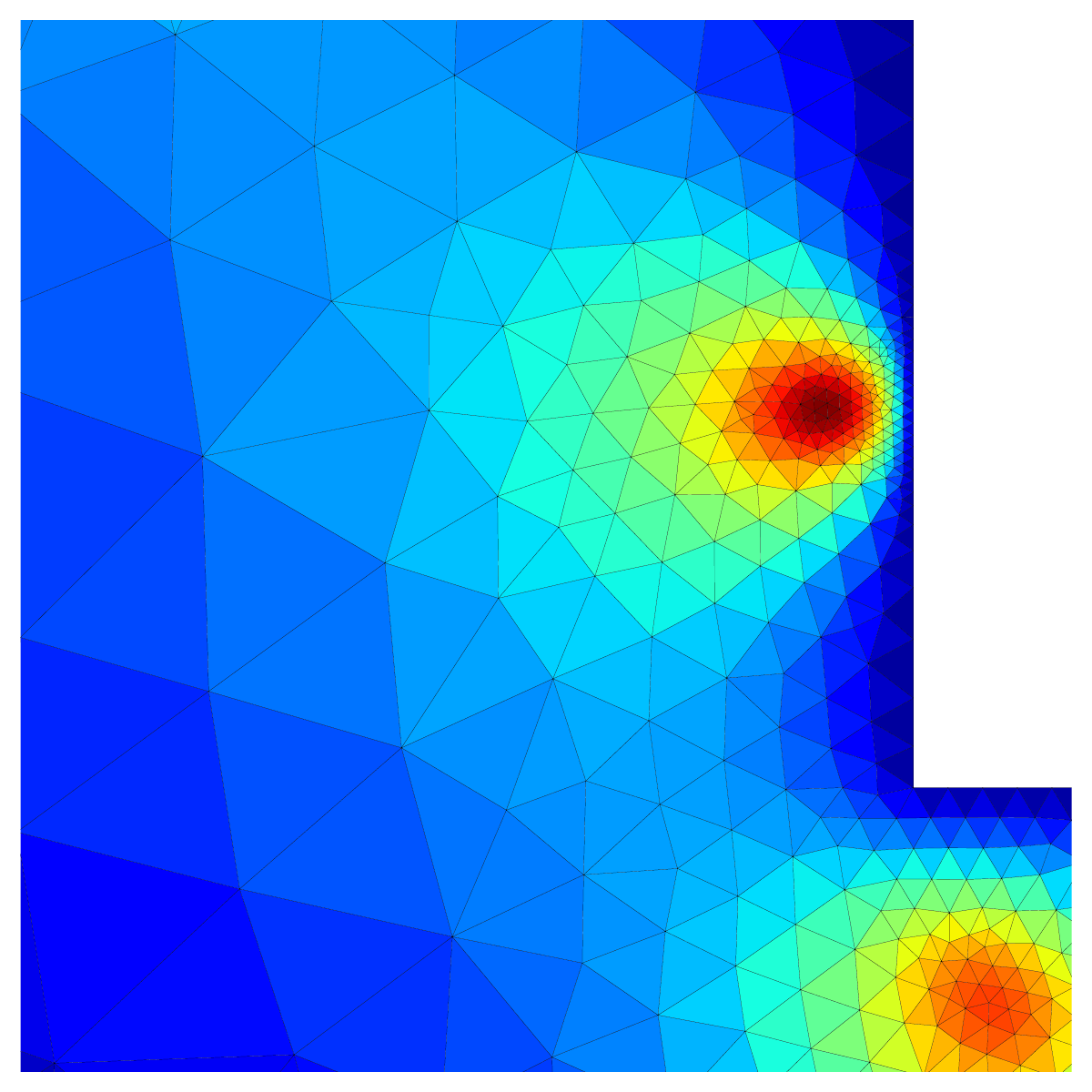}};
    \node at (3*\imagewidth, 0*\imageheight) {\includegraphics[width=\imagewidth, height=\imageheight]{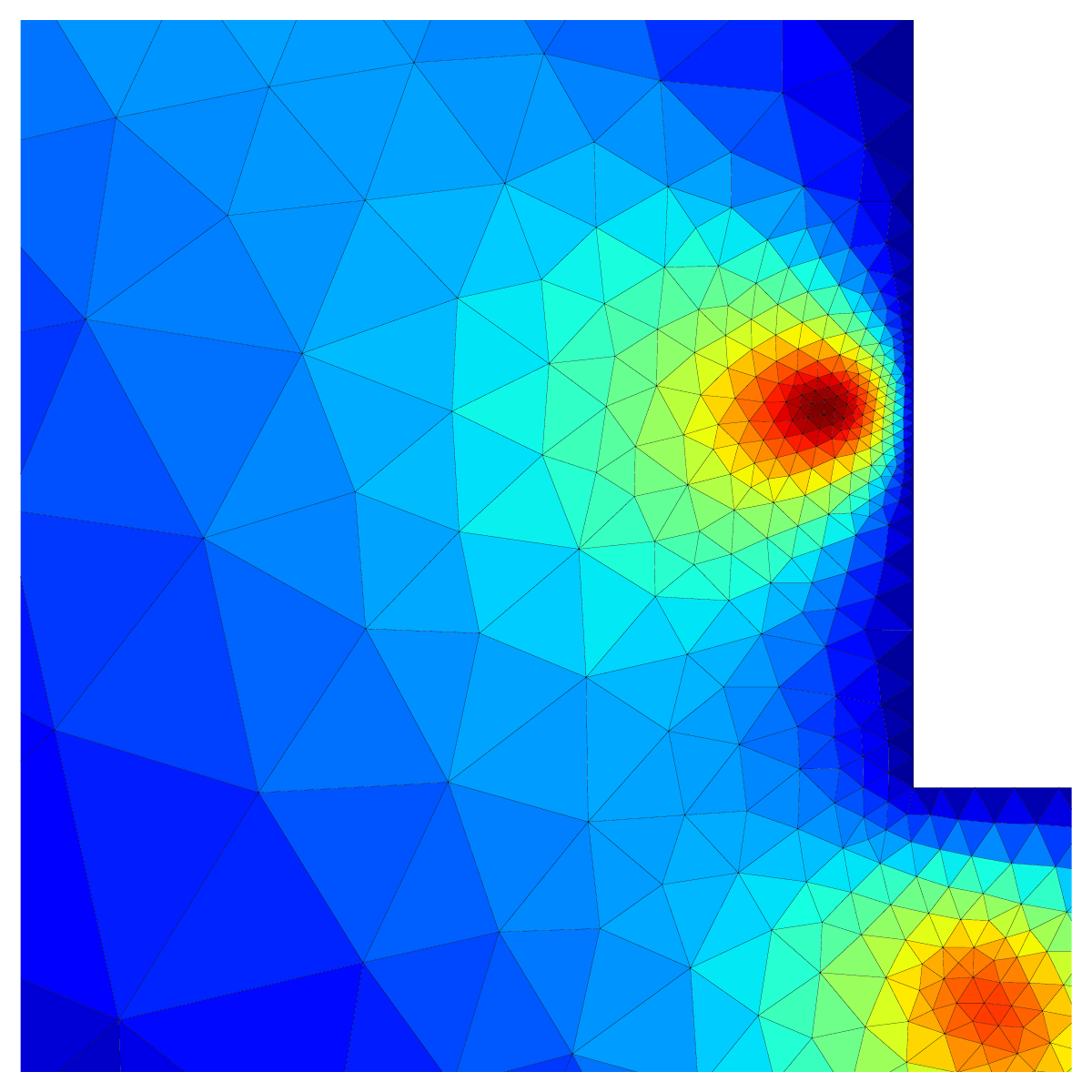}};
    \node at (4*\imagewidth, 0*\imageheight) {\includegraphics[width=\imagewidth, height=\imageheight]{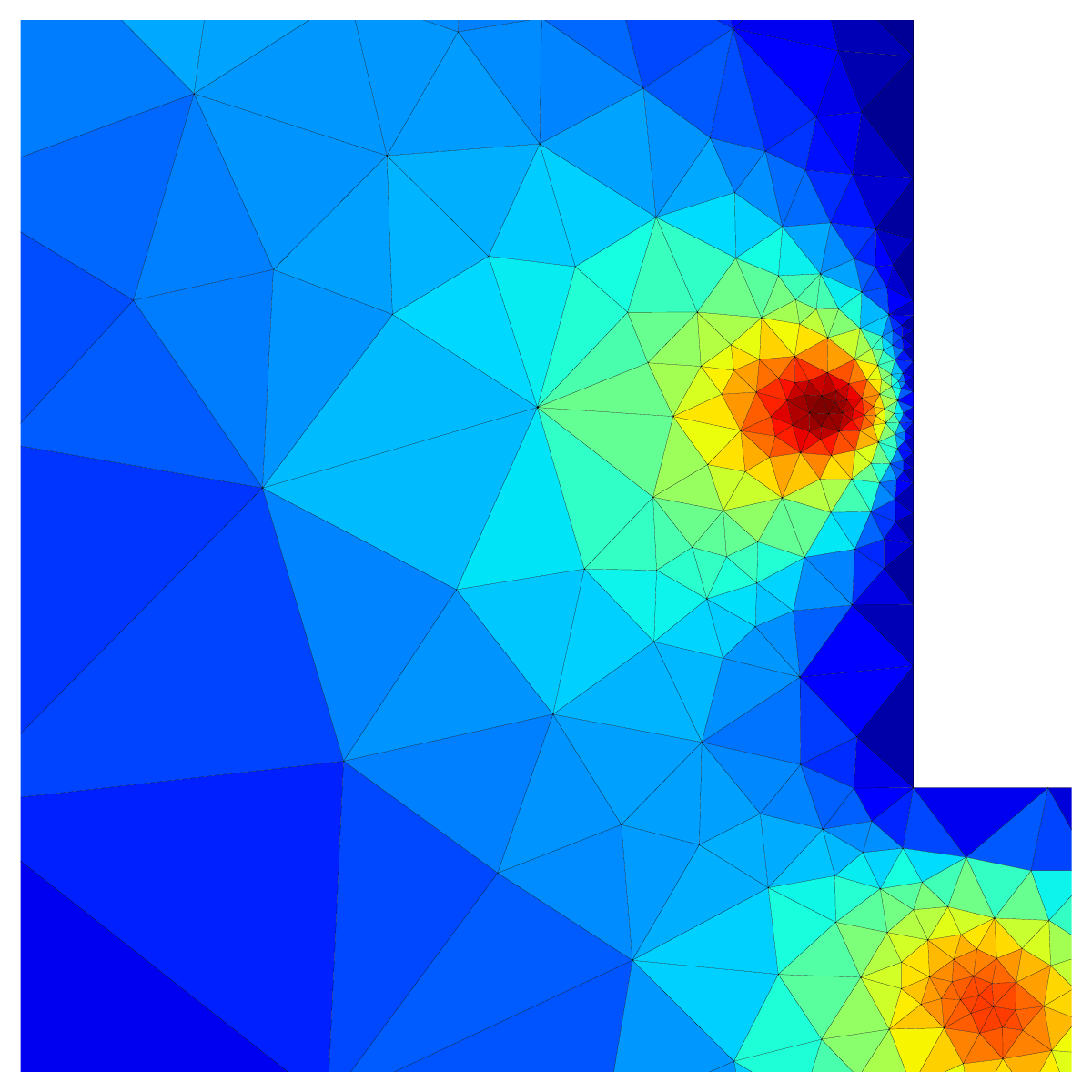}};
    
    \node at (1*\imagewidth, -1*\imageheight) {\includegraphics[width=\imagewidth, height=\imageheight]{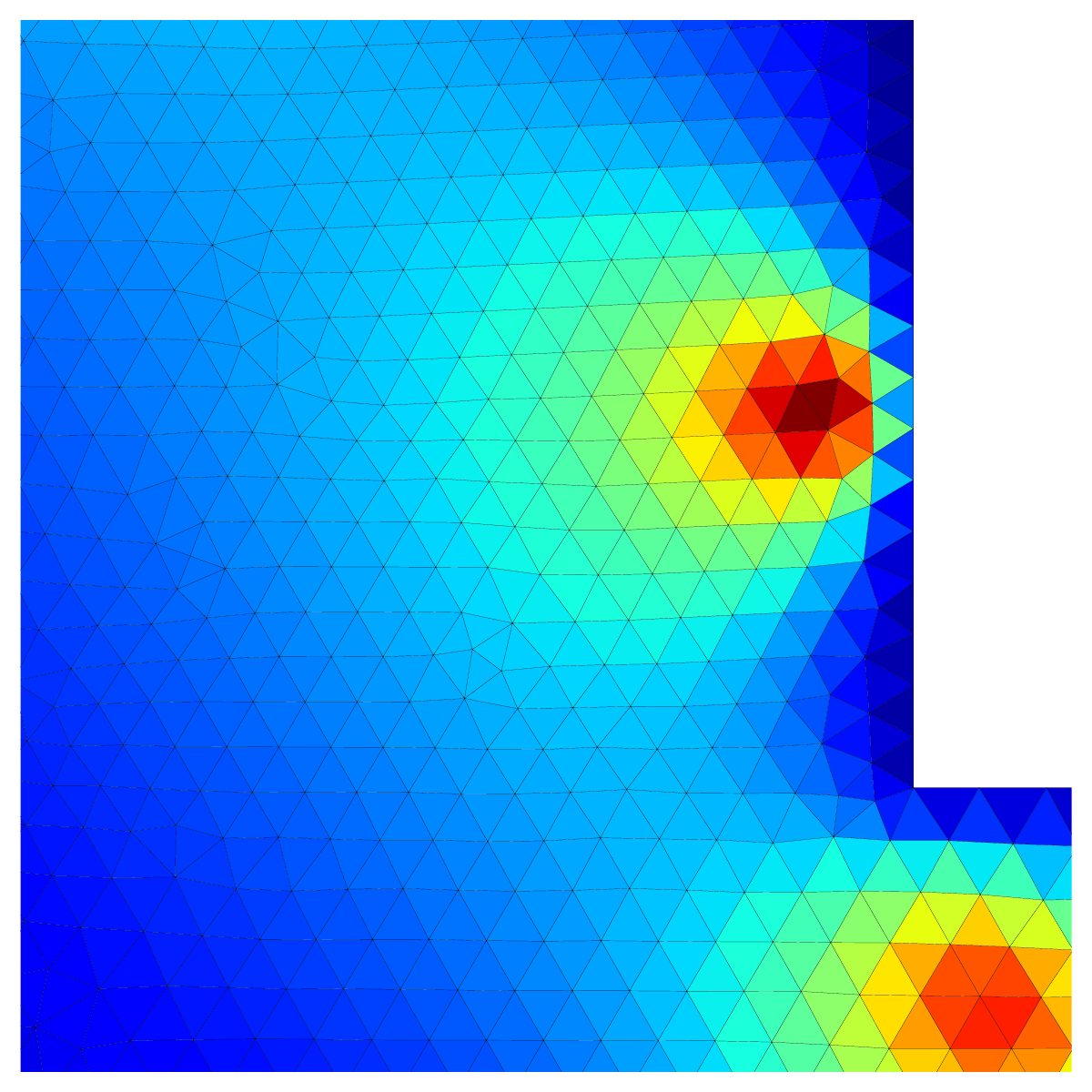}};
    \node at (2*\imagewidth, -1*\imageheight) {\includegraphics[width=\imagewidth, height=\imageheight]{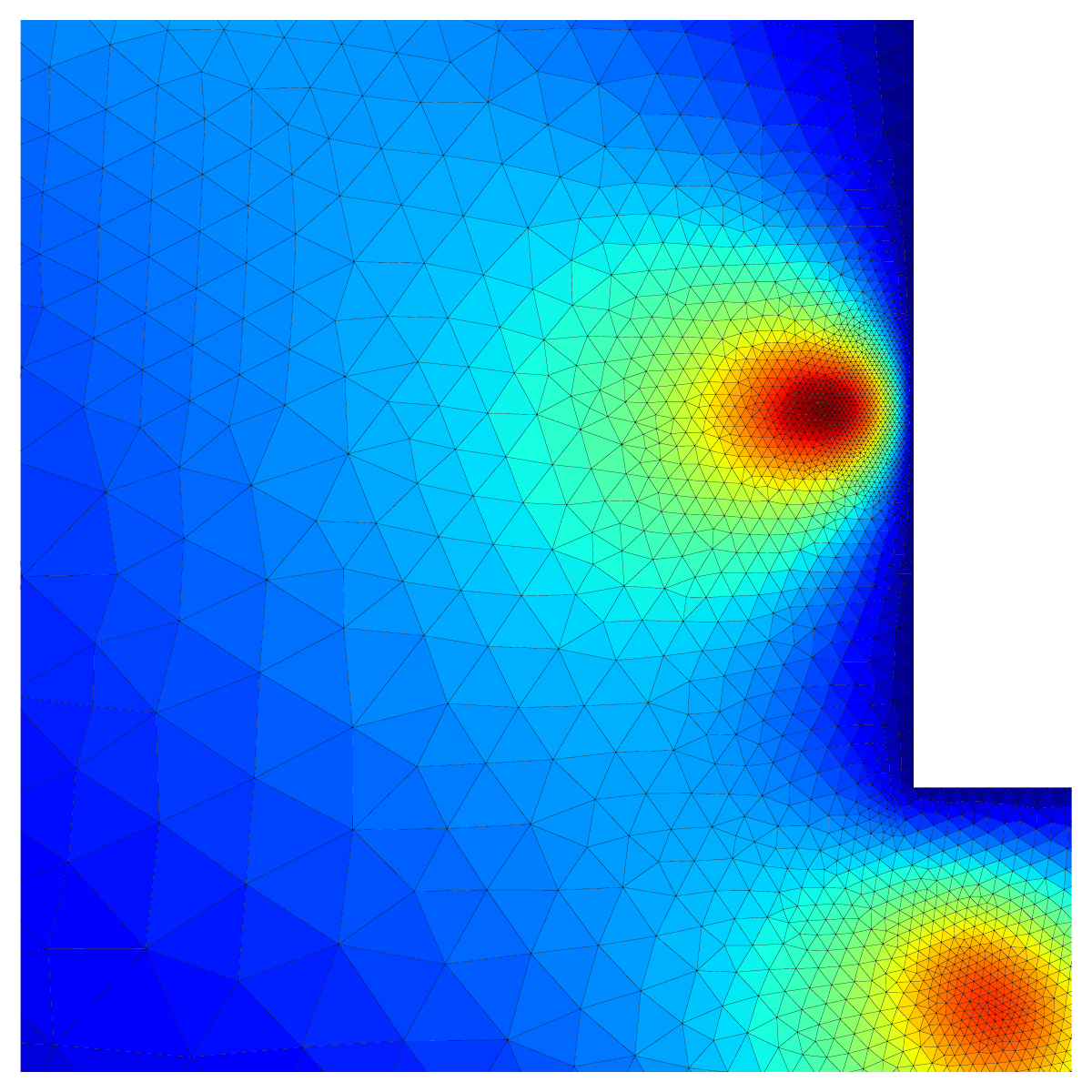}};
    \node at (3*\imagewidth, -1*\imageheight) {\includegraphics[width=\imagewidth, height=\imageheight]{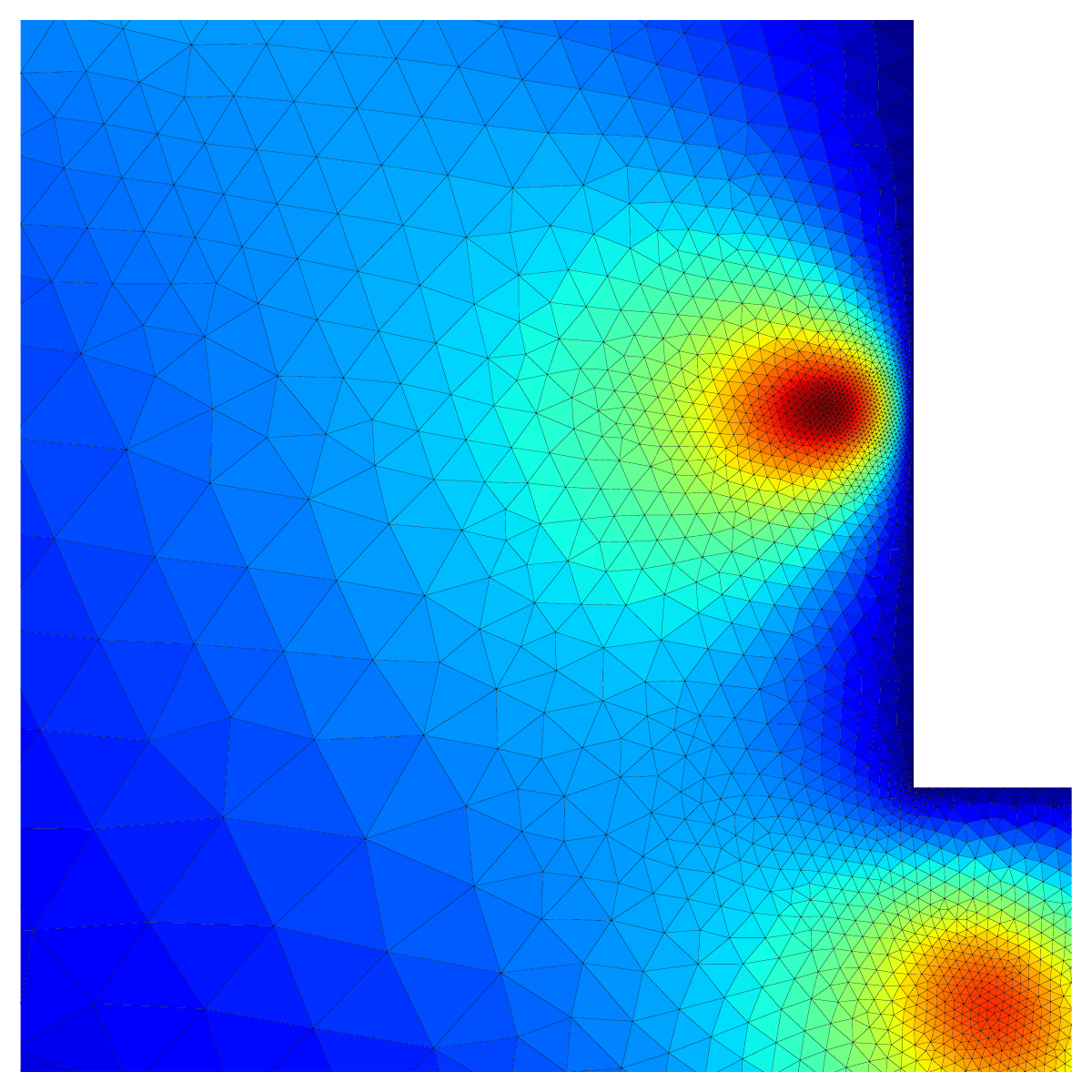}};
    \node at (4*\imagewidth, -1*\imageheight) {\includegraphics[width=\imagewidth, height=\imageheight]{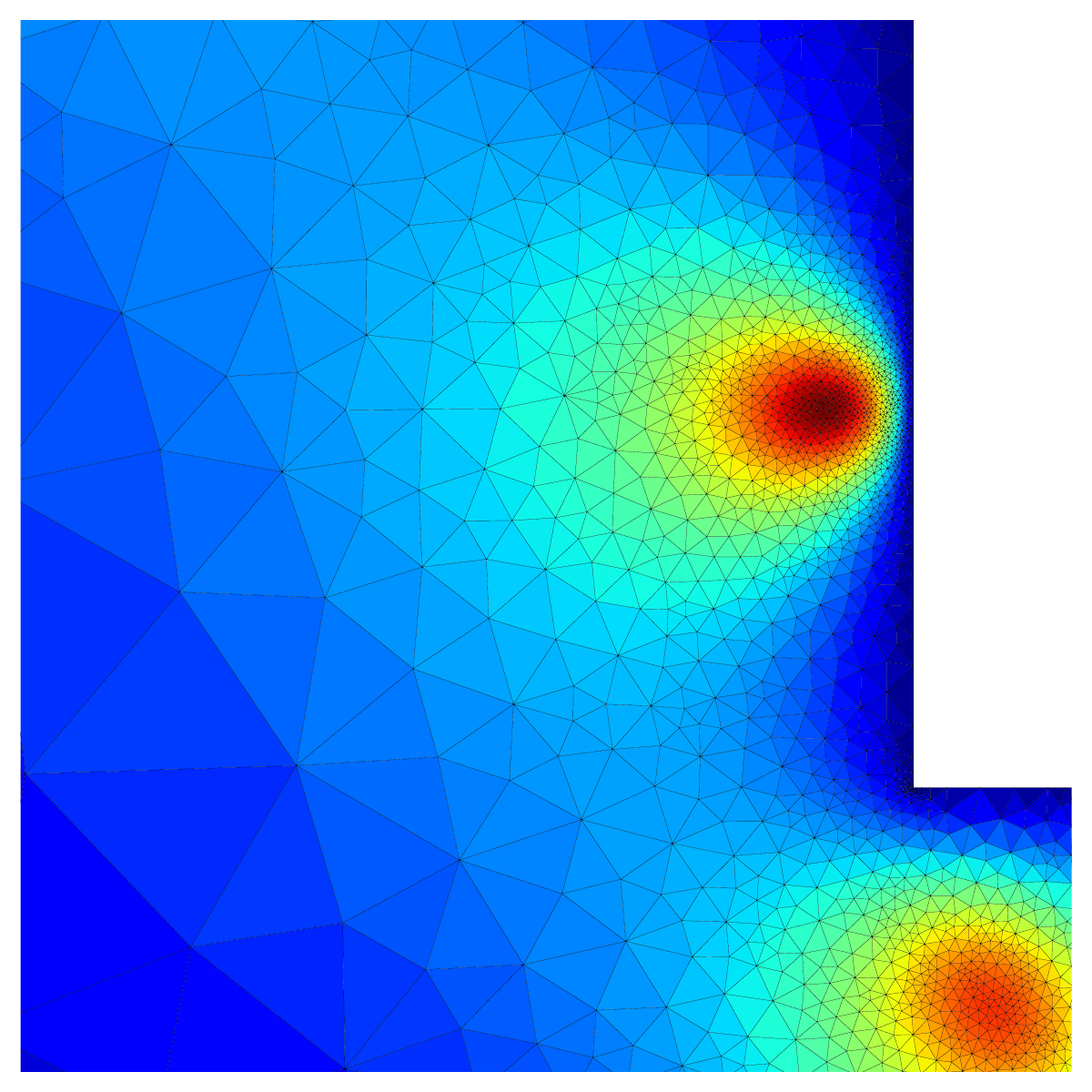}};
    
    \node at (1*\imagewidth, -2*\imageheight) {\includegraphics[width=\imagewidth, height=\imageheight]{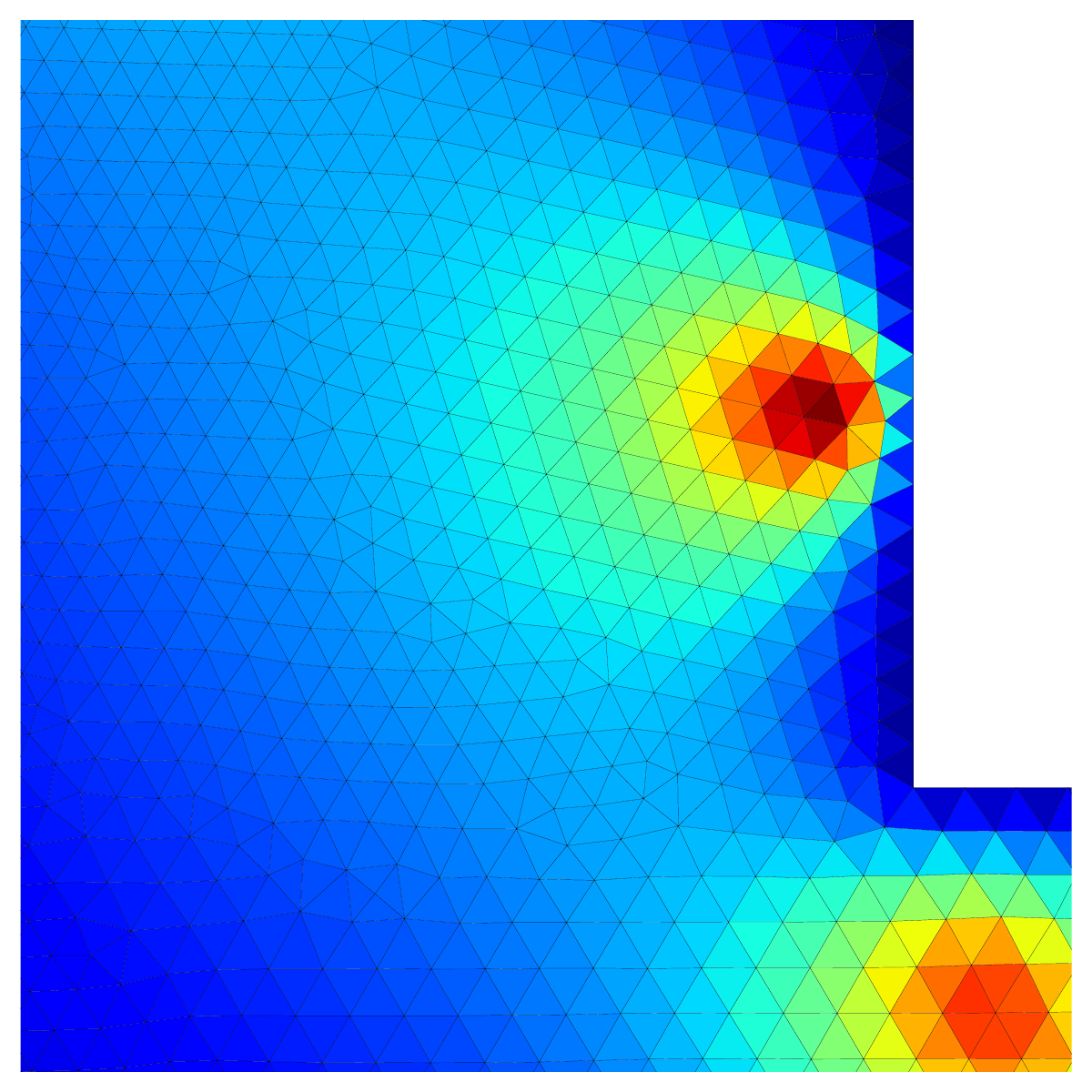}};
    \node at (2*\imagewidth, -2*\imageheight) {\includegraphics[width=\imagewidth, height=\imageheight]{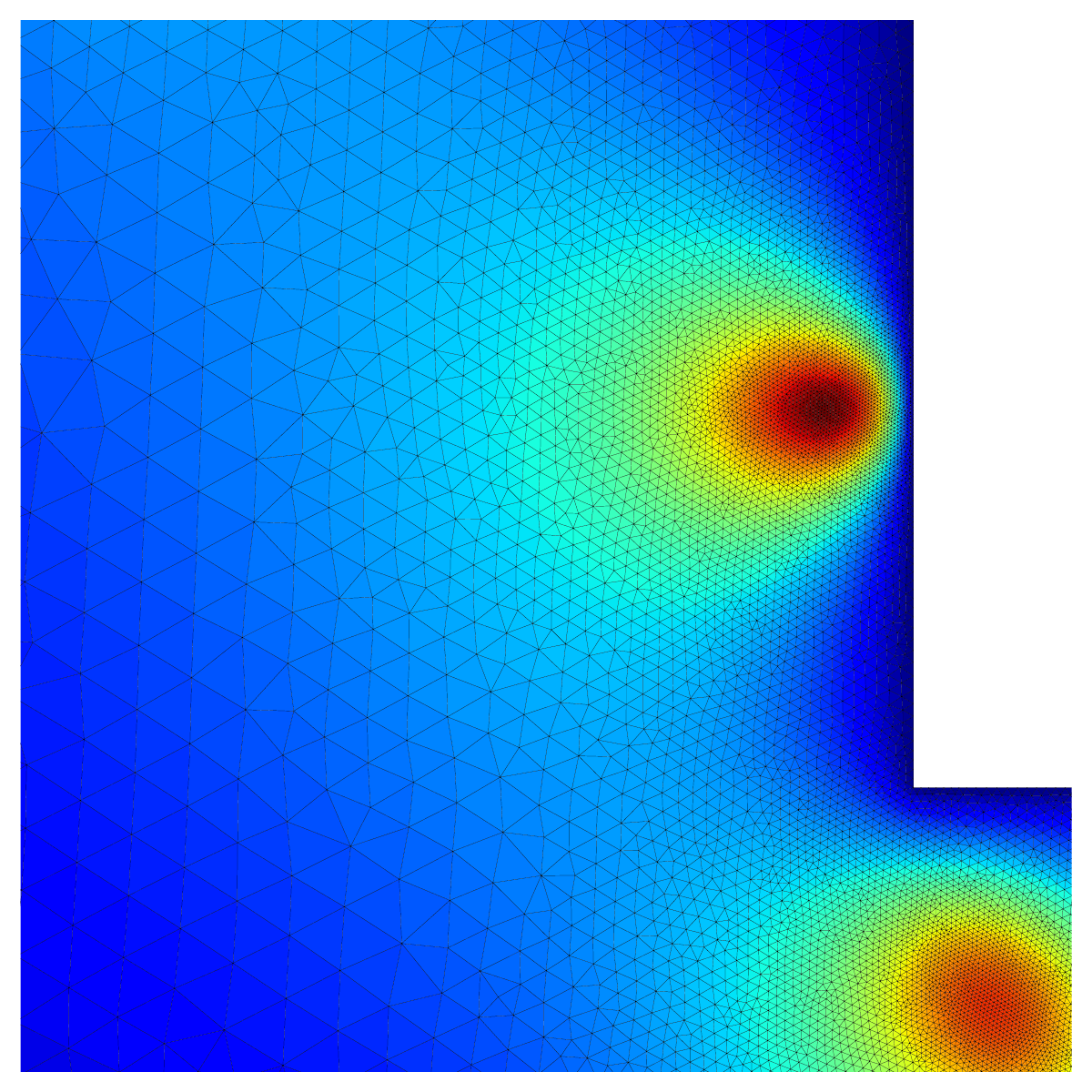}};
    \node at (3*\imagewidth, -2*\imageheight) {\includegraphics[width=\imagewidth, height=\imageheight]{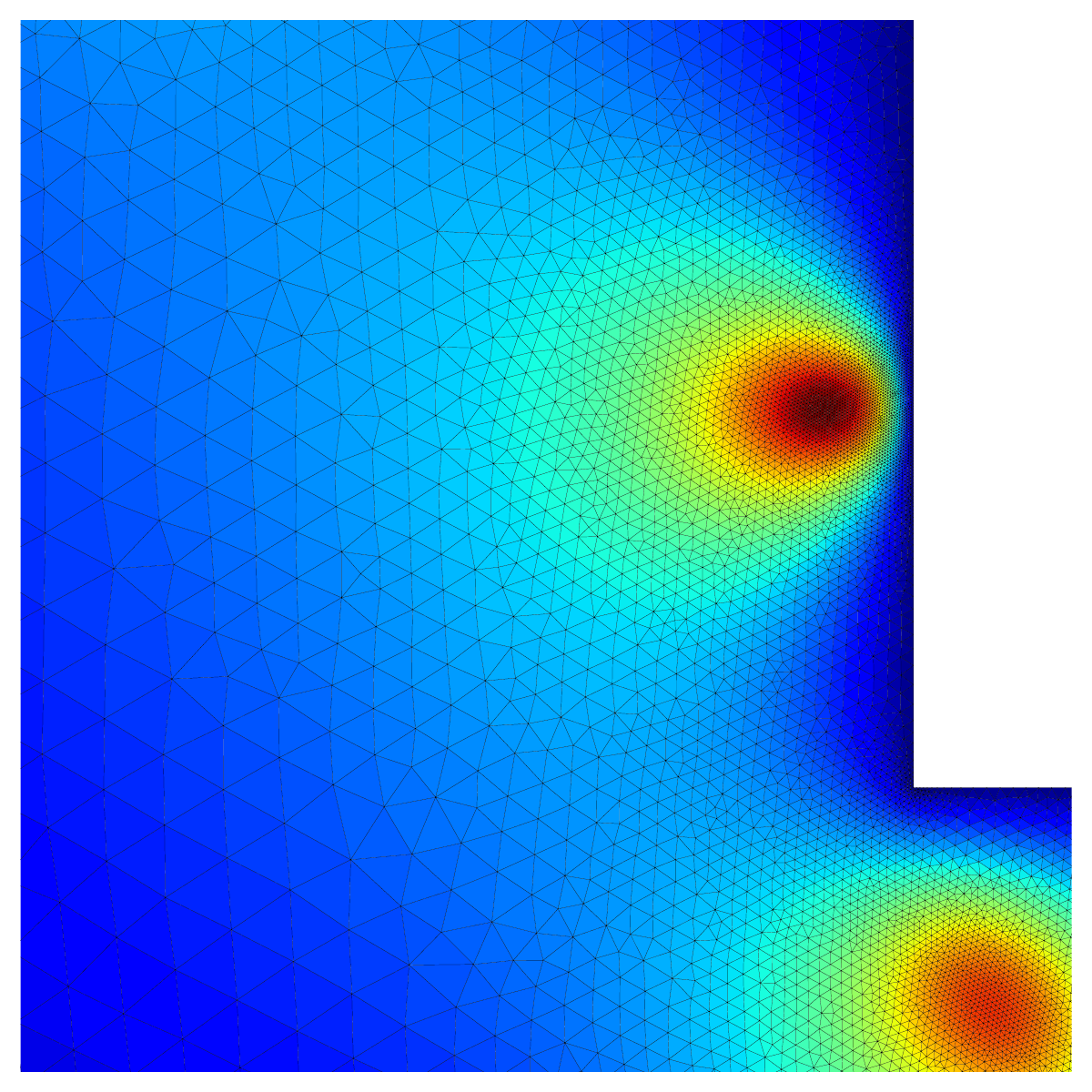}};
    \node at (4*\imagewidth, -2*\imageheight) {\includegraphics[width=\imagewidth, height=\imageheight]{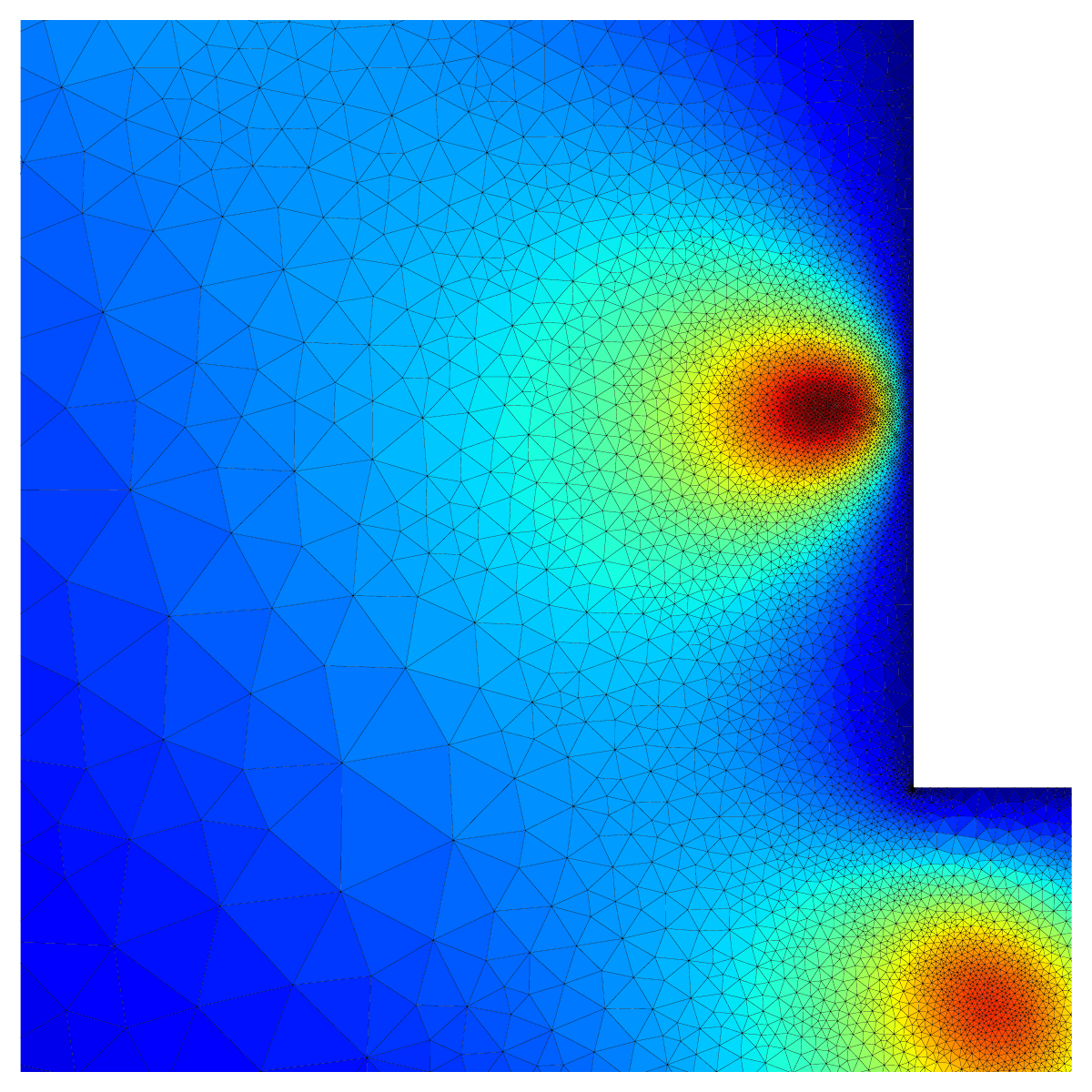}};
    
    \node at (1*\imagewidth, -3*\imageheight) {\includegraphics[width=\imagewidth, height=\imageheight]{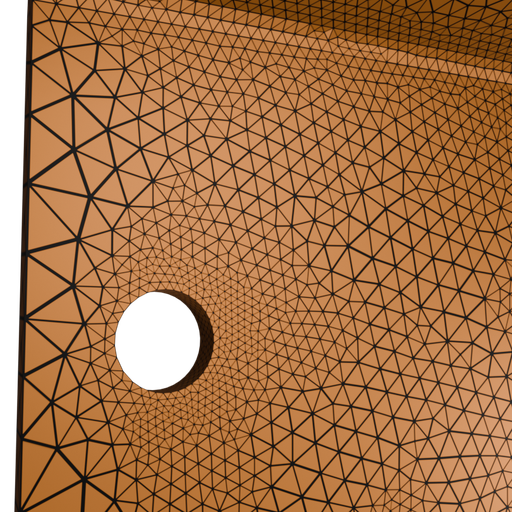}};
    \node at (2*\imagewidth, -3*\imageheight) {\includegraphics[width=\imagewidth, height=\imageheight]{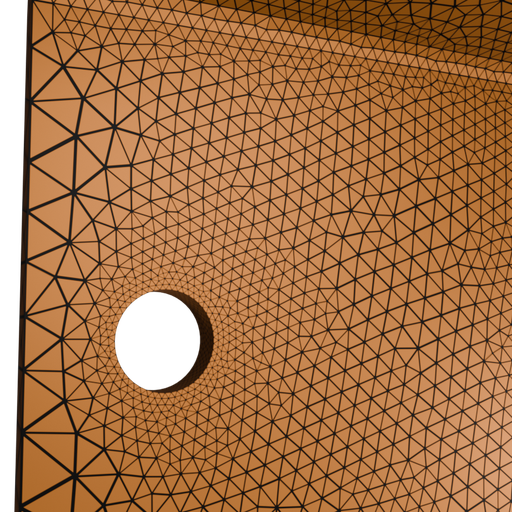}};
    \node at (3*\imagewidth, -3*\imageheight) {\includegraphics[width=\imagewidth, height=\imageheight]{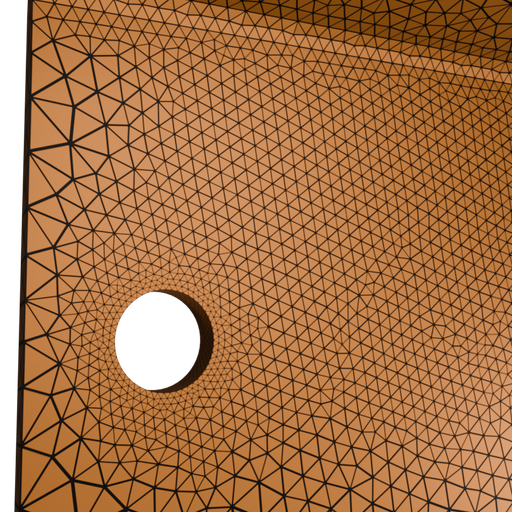}};
    \node at (4*\imagewidth, -3*\imageheight) {\includegraphics[width=\imagewidth, height=\imageheight]{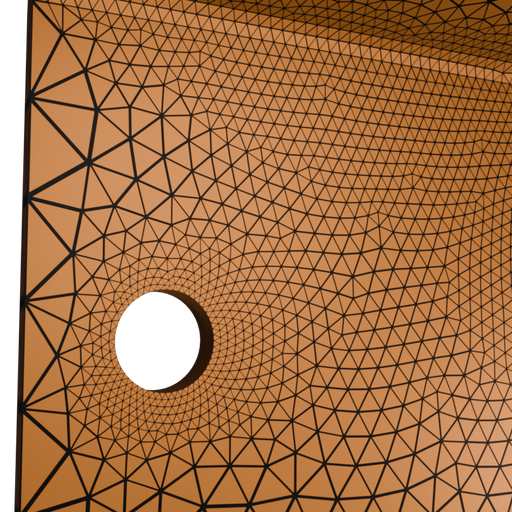}};
\end{tikzpicture}
\caption{
    Mesh generation steps of \gls{amber} (Mean) for all tasks. 
    \textbf{Row 1 to 3:} Poisson's Equation with $25$/$50$/$75$ expert refinement steps. \textbf{Row 4:} Console task. 
    \gls{amber} (Mean) converges quickly to meshes that match the expert (displayed in the last column).
    } 
    \label{app_fig:matrix_amber_mean}
\end{figure}

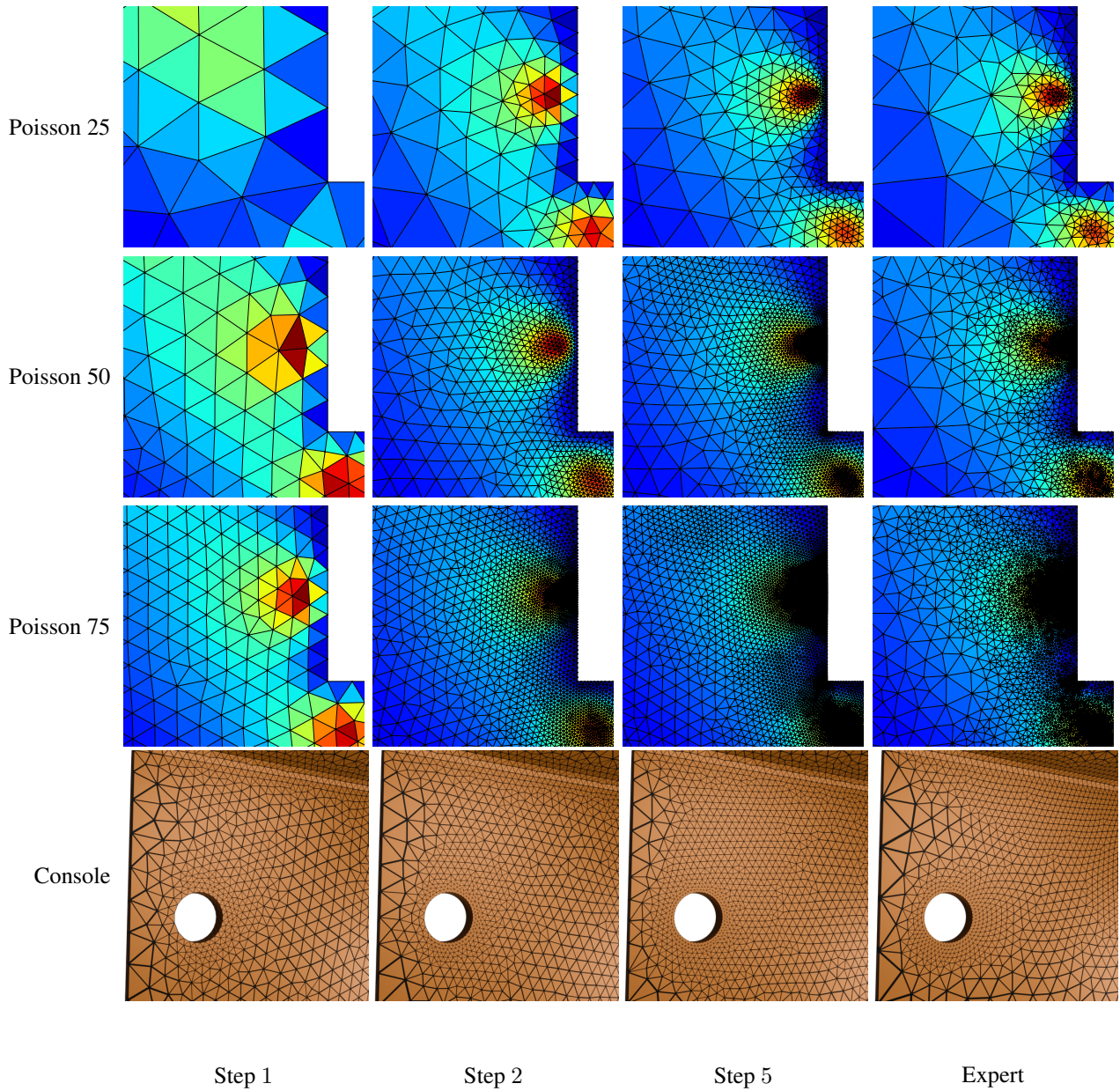
\begin{figure}
    \centering
    \begin{tikzpicture}
    \def\imagewidth{0.22\textwidth}
    \def\imageheight{0.22\textwidth}

    \node[anchor=east] at (0.5*\imagewidth, -0.5*\imageheight + 0.5*\imageheight) {Poisson 25};
    \node[anchor=east] at (0.5*\imagewidth, -1.5*\imageheight + 0.5*\imageheight) {Poisson 50};
    \node[anchor=east] at (0.5*\imagewidth, -2.5*\imageheight + 0.5*\imageheight) {Poisson 75};
    \node[anchor=east] at (0.5*\imagewidth, -3.5*\imageheight + 0.5*\imageheight) {Console};

    \node[anchor=north] at (1*\imagewidth, -4*\imageheight + 1cm) {Step $1$};
    \node[anchor=north] at (2*\imagewidth, -4*\imageheight + 1cm) {Step $2$};
    \node[anchor=north] at (3*\imagewidth, -4*\imageheight + 1cm) {Step $5$};
    \node[anchor=north] at (4*\imagewidth, -4*\imageheight + 1cm) {Expert};

    \node at (1*\imagewidth, 0*\imageheight) {\includegraphics[width=\imagewidth, height=\imageheight]{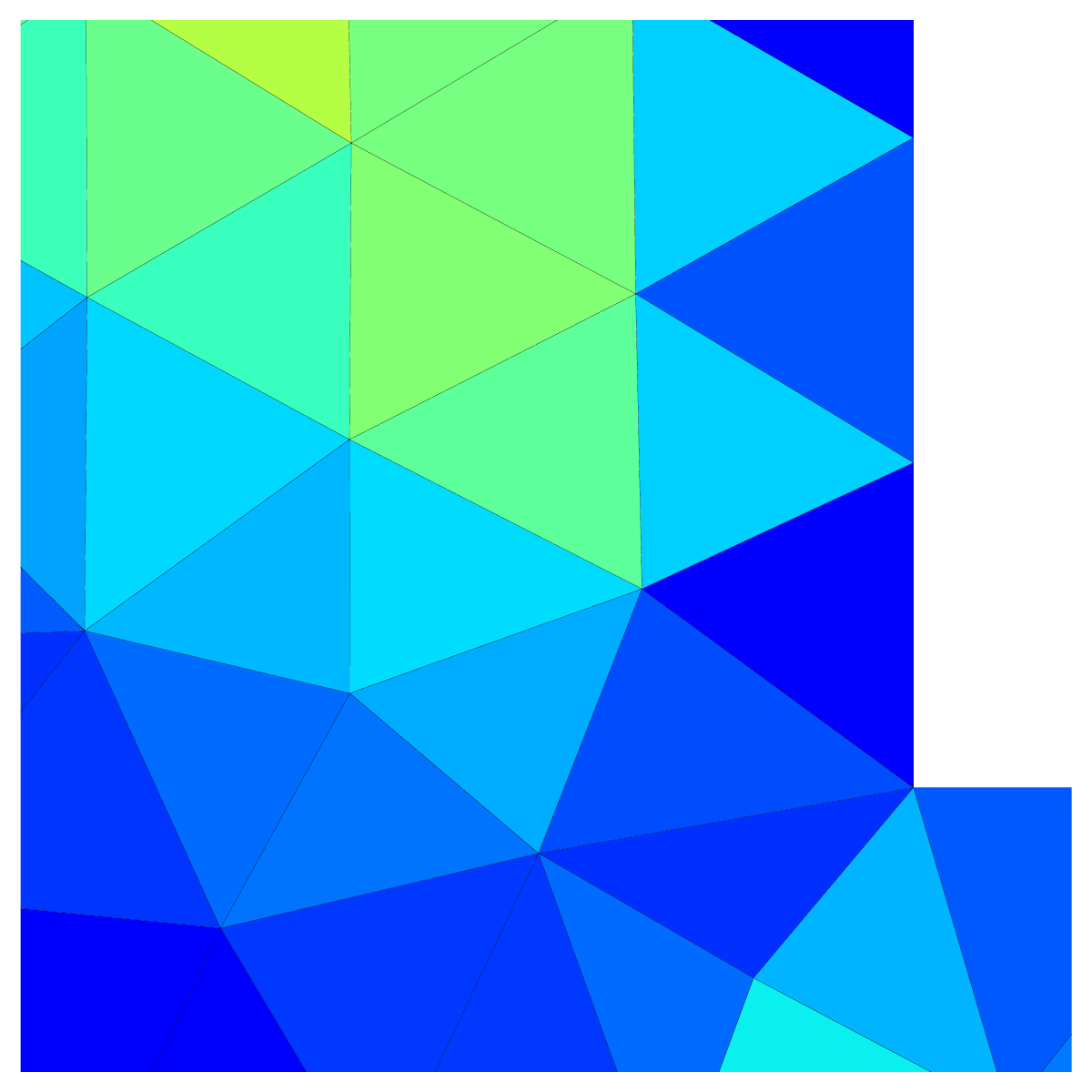}};
    \node at (2*\imagewidth, 0*\imageheight) {\includegraphics[width=\imagewidth, height=\imageheight]{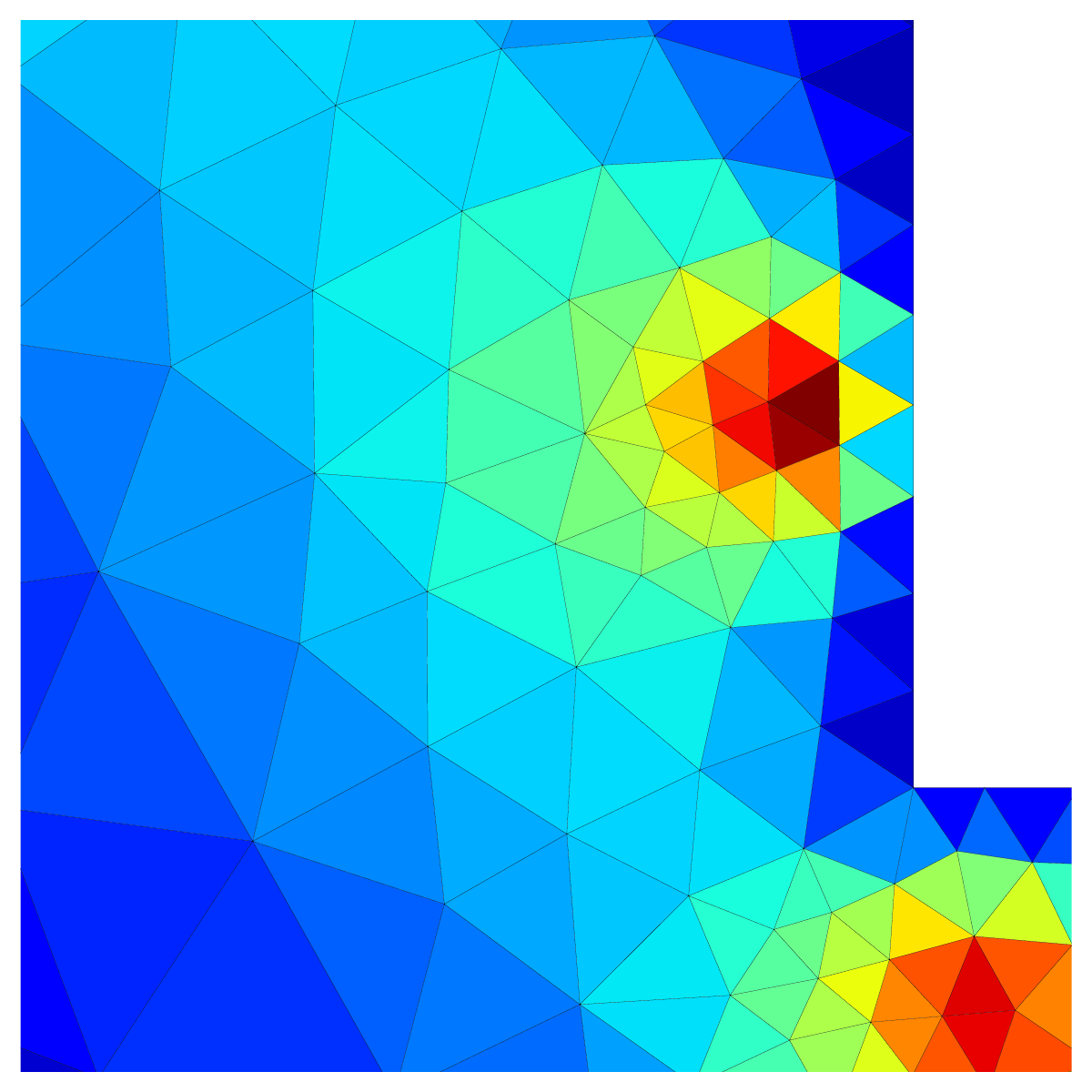}};
    \node at (3*\imagewidth, 0*\imageheight) {\includegraphics[width=\imagewidth, height=\imageheight]{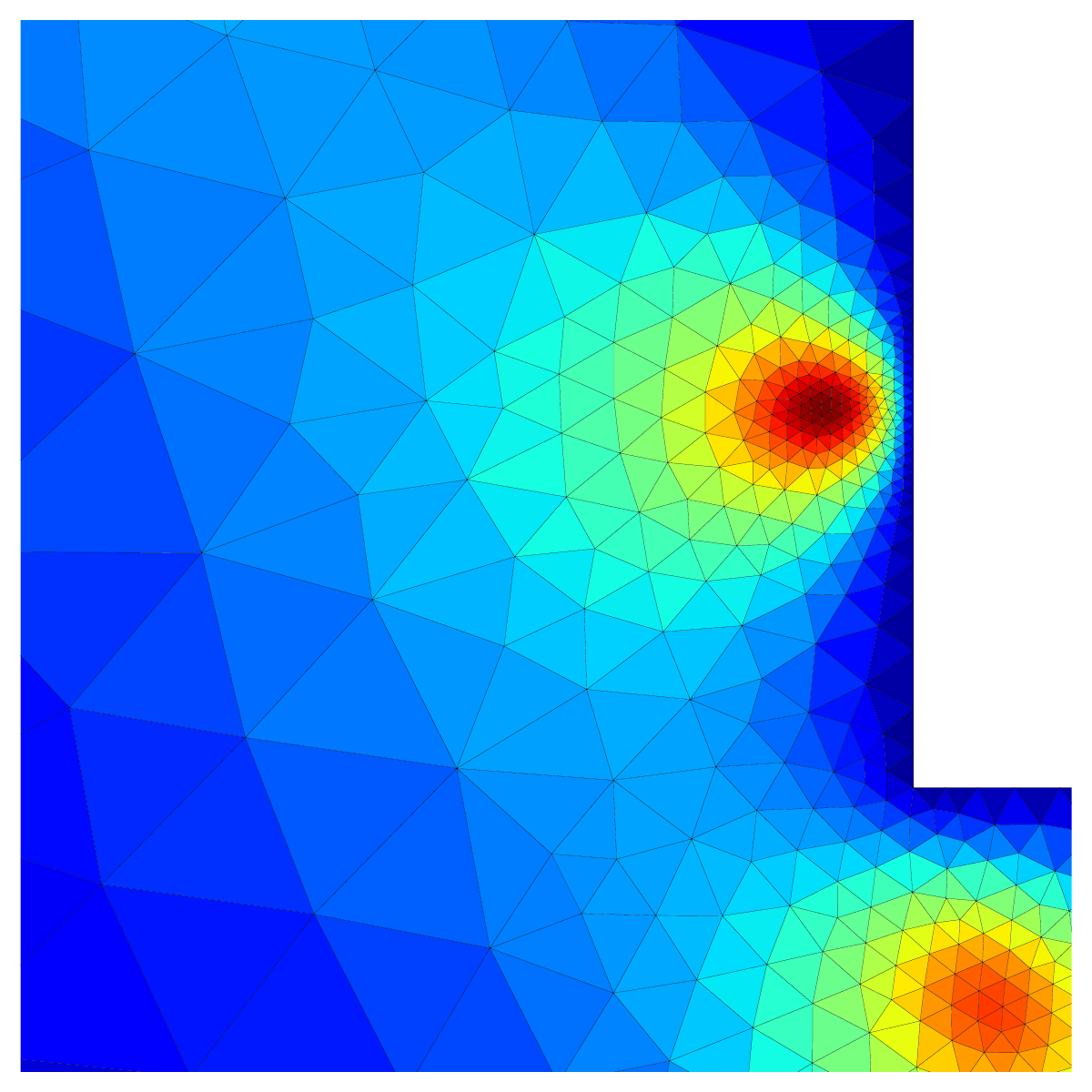}};
    \node at (4*\imagewidth, 0*\imageheight) {\includegraphics[width=\imagewidth, height=\imageheight]{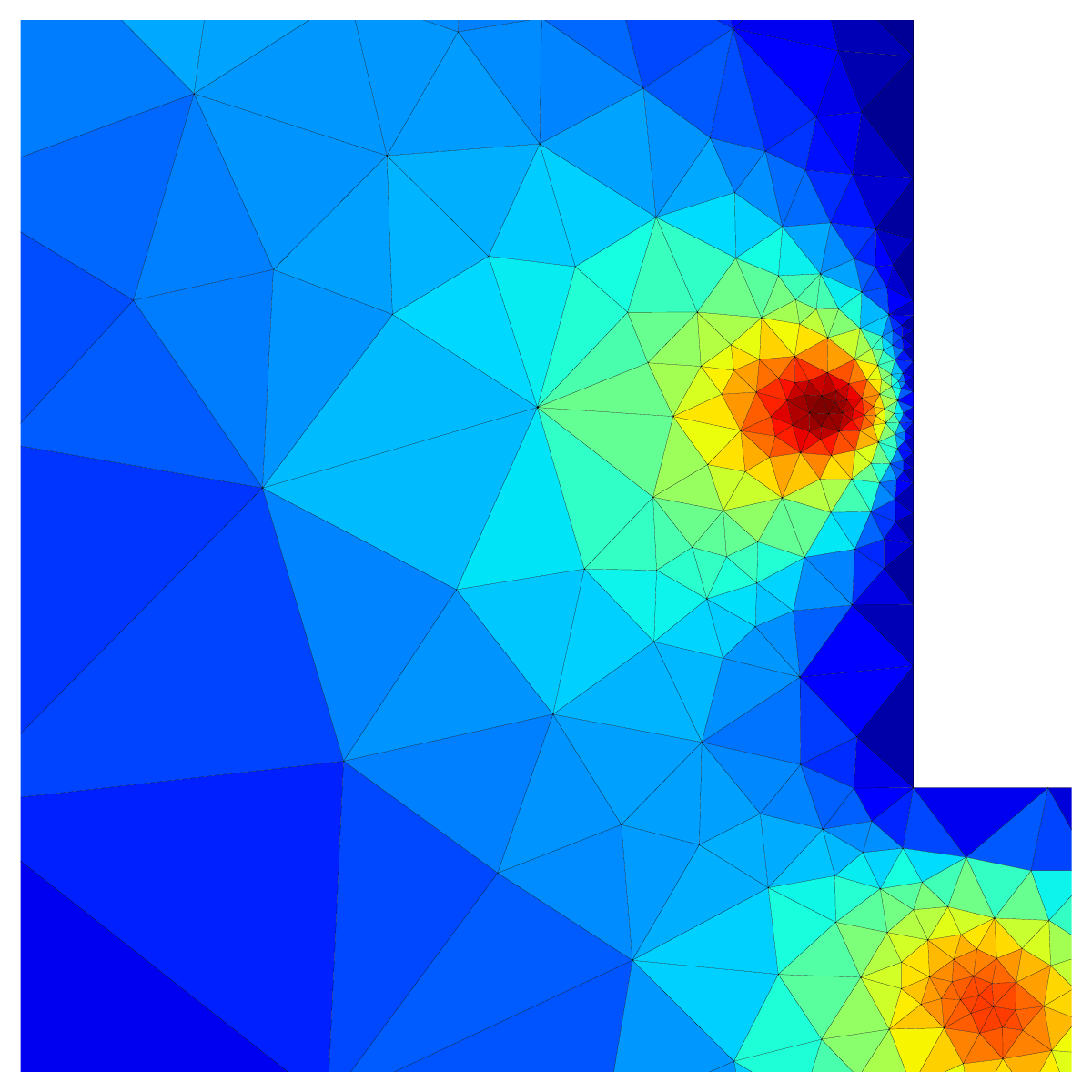}};
    
    \node at (1*\imagewidth, -1*\imageheight) {\includegraphics[width=\imagewidth, height=\imageheight]{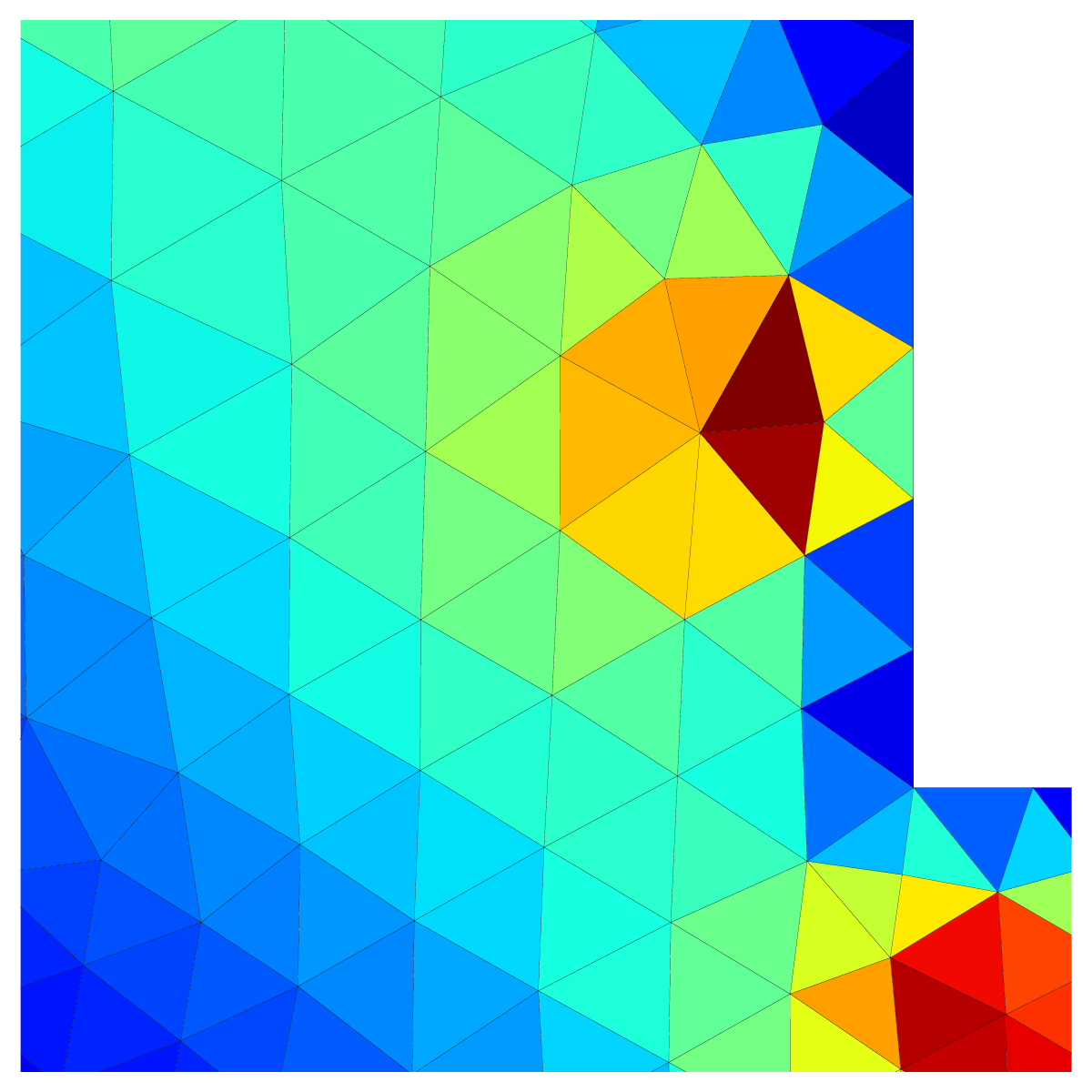}};
    \node at (2*\imagewidth, -1*\imageheight) {\includegraphics[width=\imagewidth, height=\imageheight]{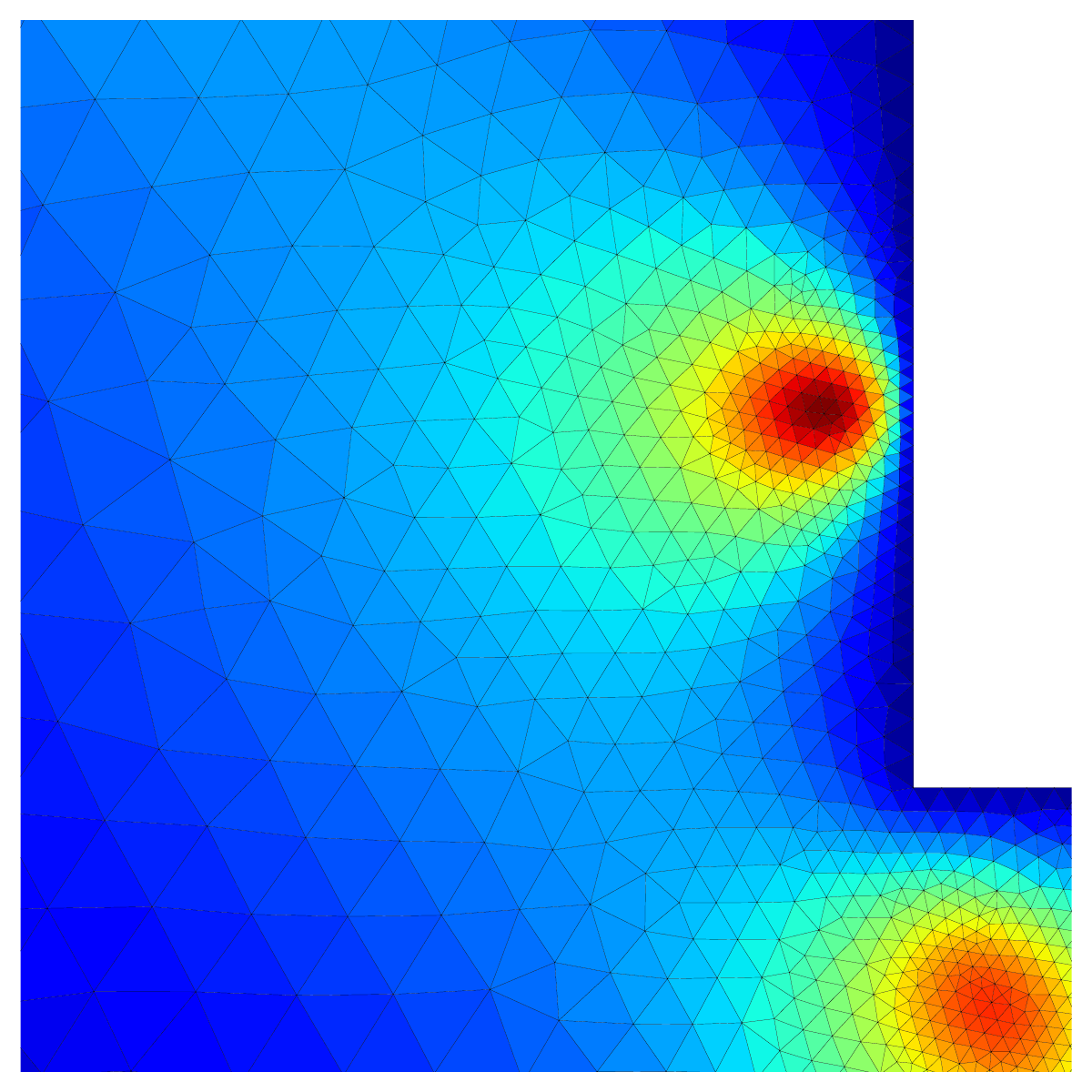}};
    \node at (3*\imagewidth, -1*\imageheight) {\includegraphics[width=\imagewidth, height=\imageheight]{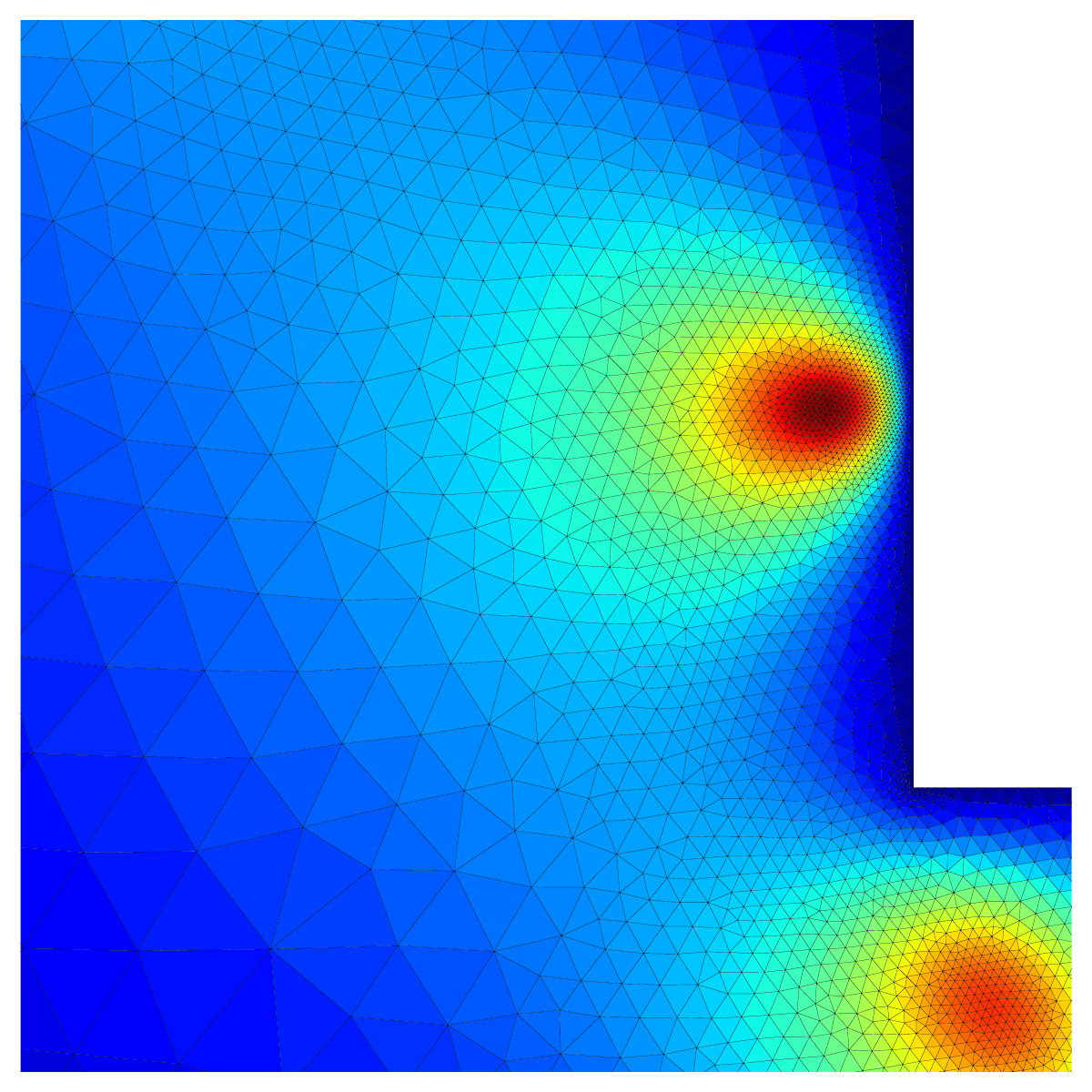}};
    \node at (4*\imagewidth, -1*\imageheight) {\includegraphics[width=\imagewidth, height=\imageheight]{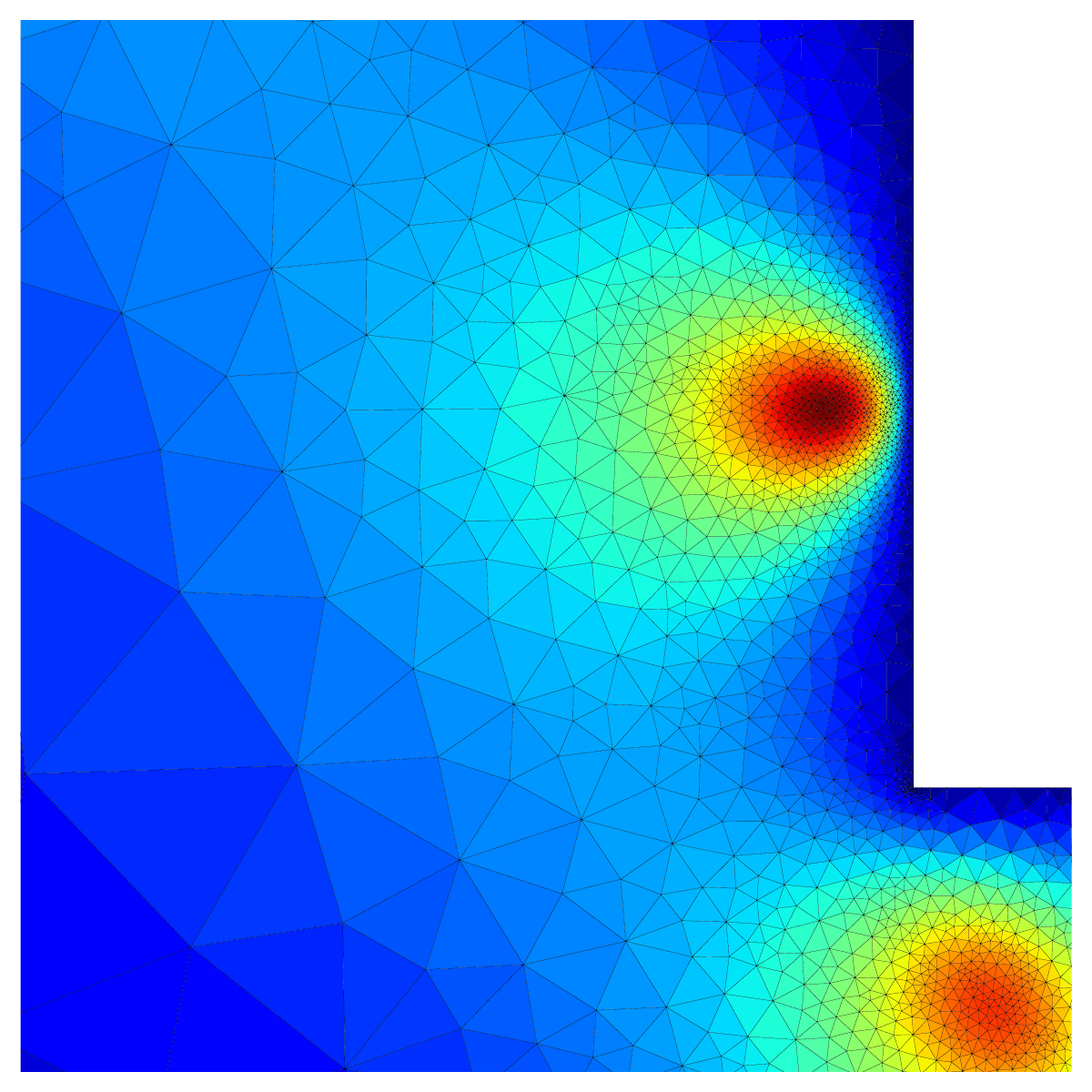}};

    \node at (1*\imagewidth, -2*\imageheight) {\includegraphics[width=\imagewidth, height=\imageheight]{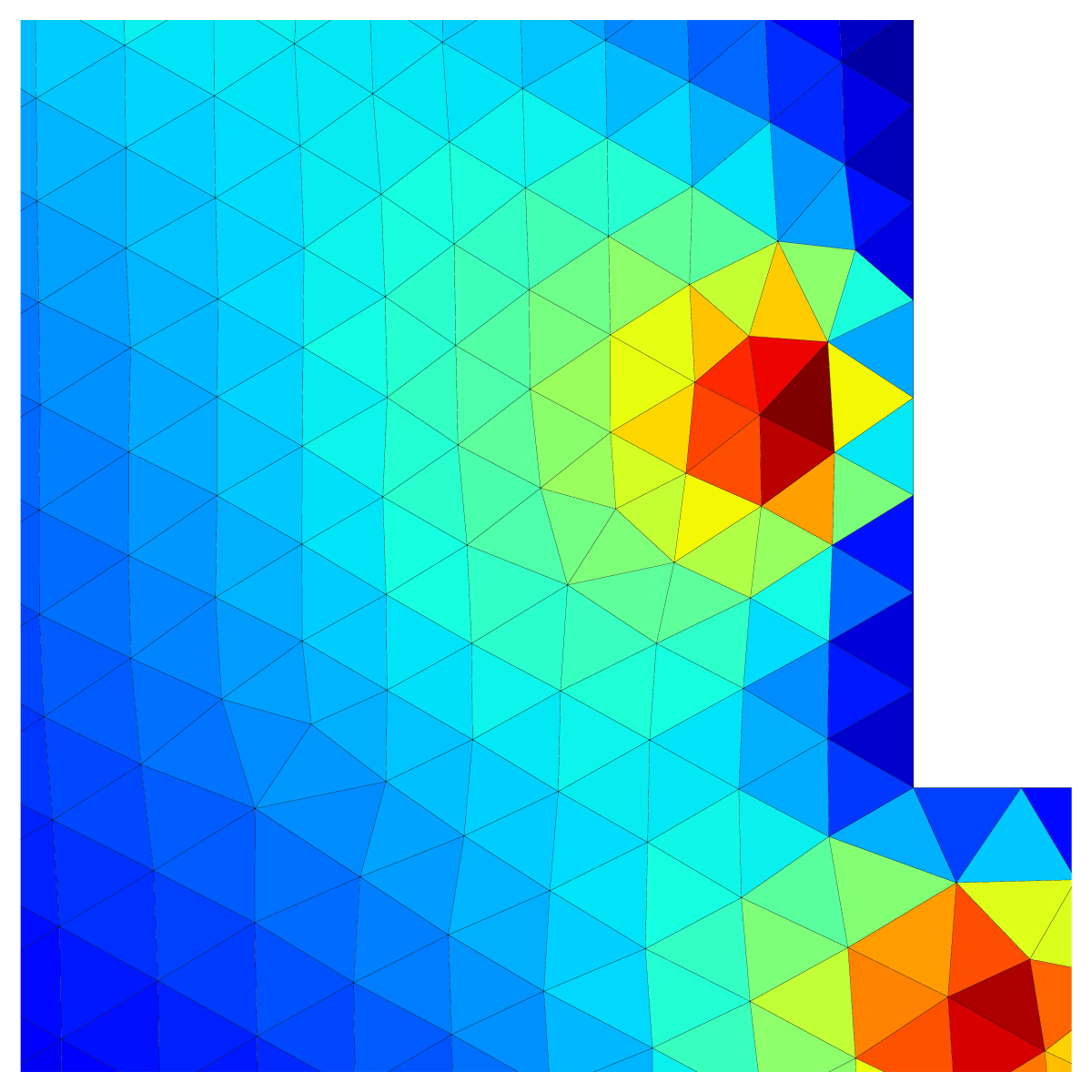}};
    \node at (2*\imagewidth, -2*\imageheight) {\includegraphics[width=\imagewidth, height=\imageheight]{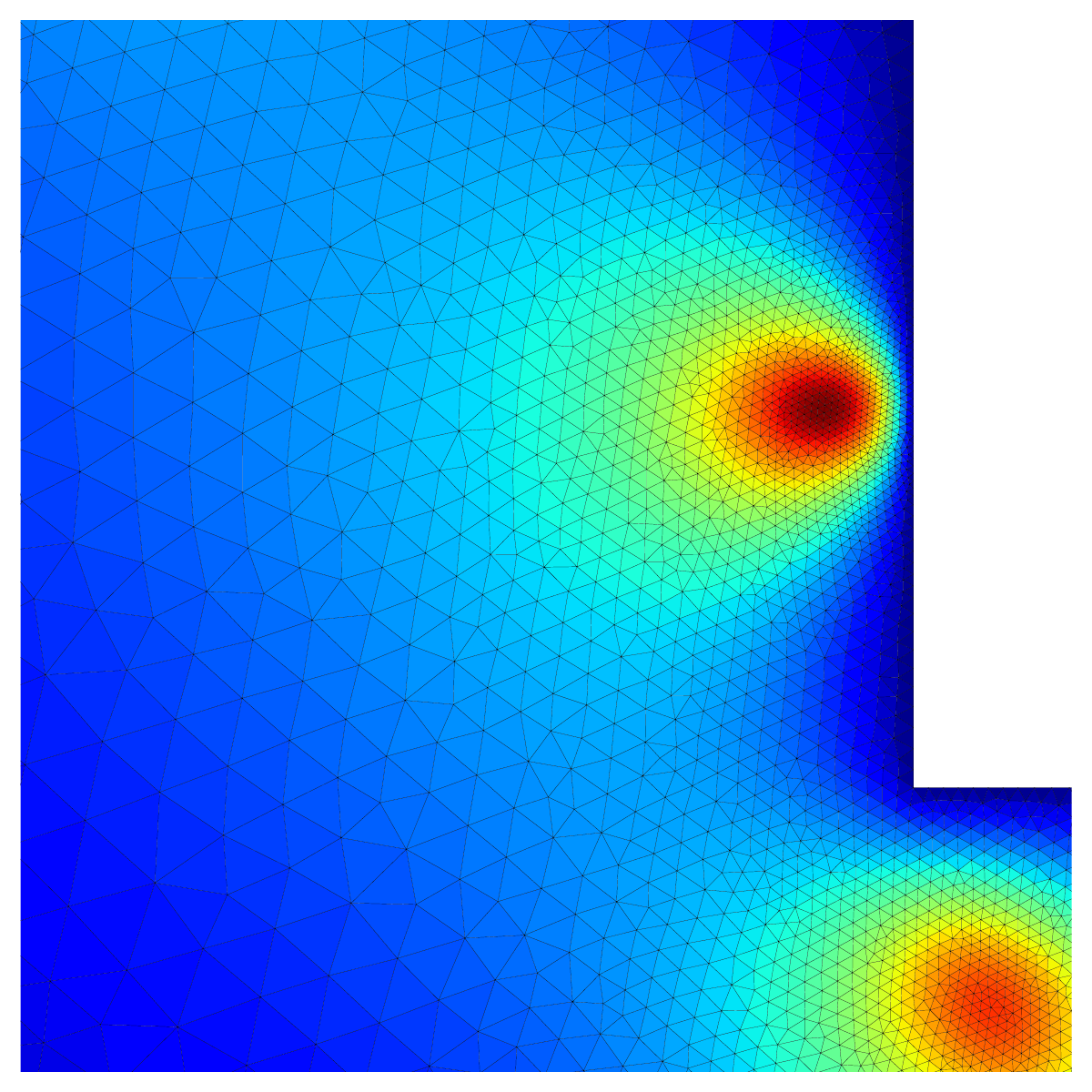}};
    \node at (3*\imagewidth, -2*\imageheight) {\includegraphics[width=\imagewidth, height=\imageheight]{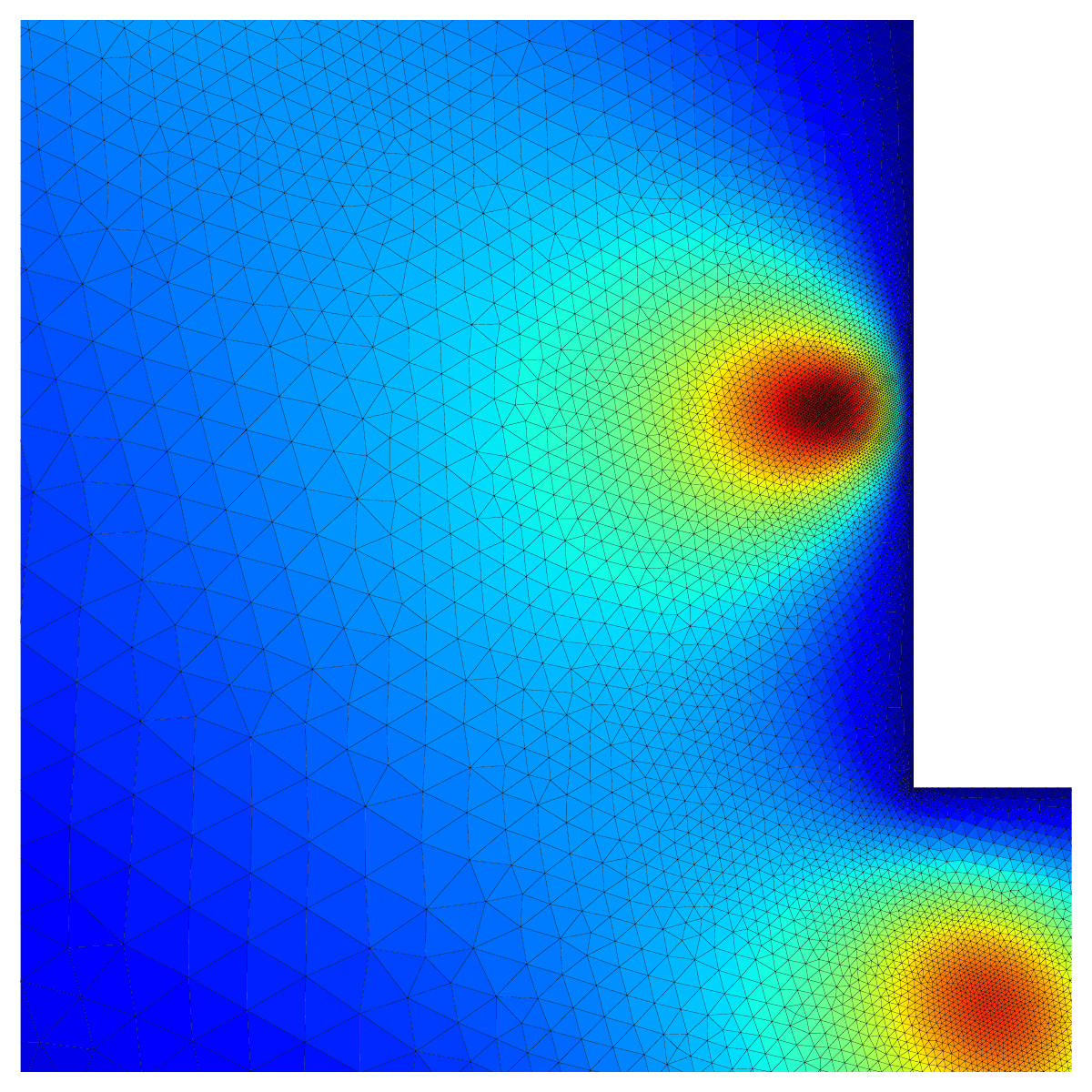}};
    \node at (4*\imagewidth, -2*\imageheight) {\includegraphics[width=\imagewidth, height=\imageheight]{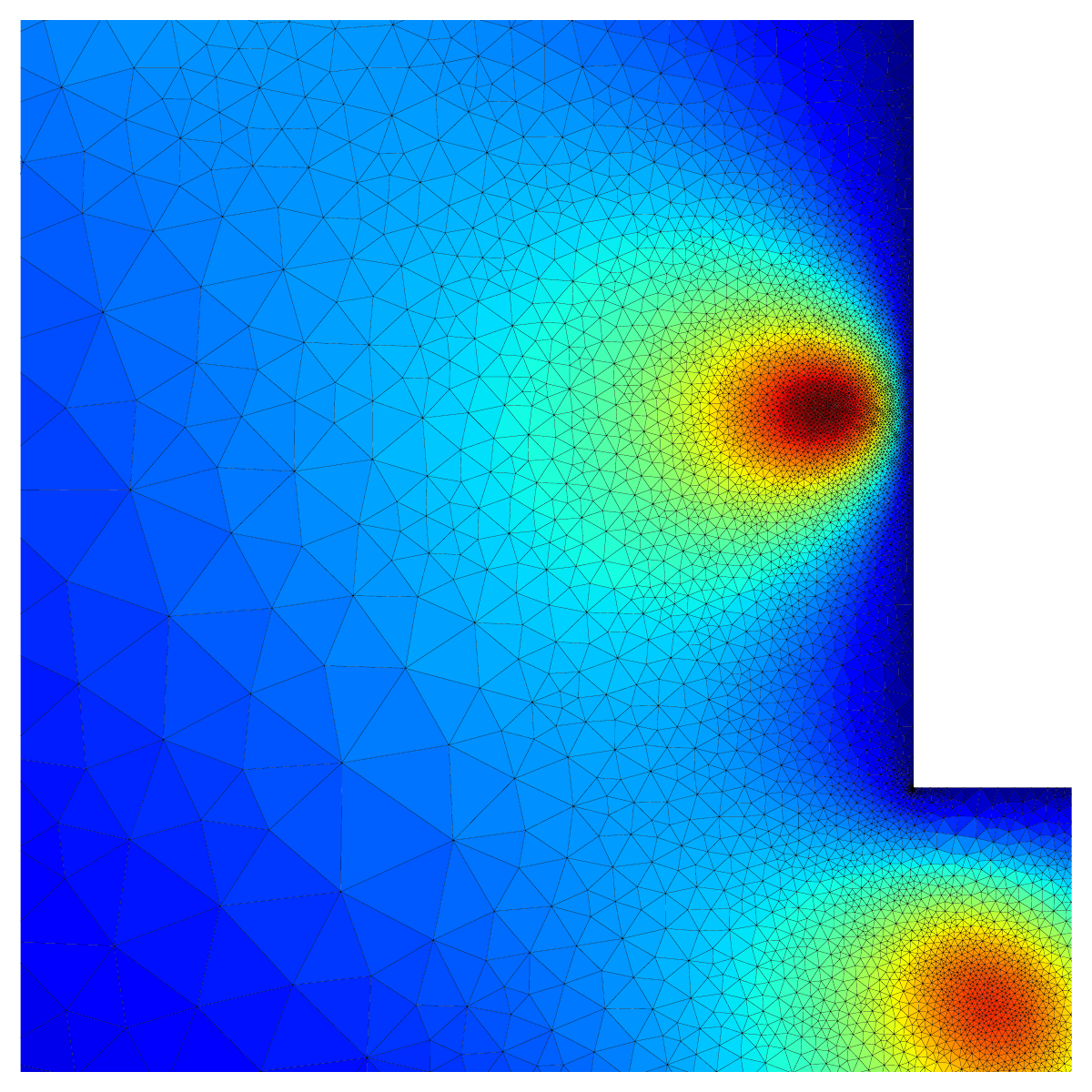}};
    
    \node at (1*\imagewidth, -3*\imageheight) {\includegraphics[width=\imagewidth, height=\imageheight]{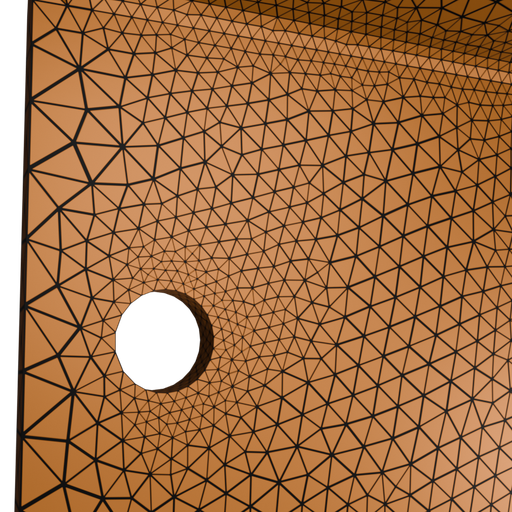}};
    \node at (2*\imagewidth, -3*\imageheight) {\includegraphics[width=\imagewidth, height=\imageheight]{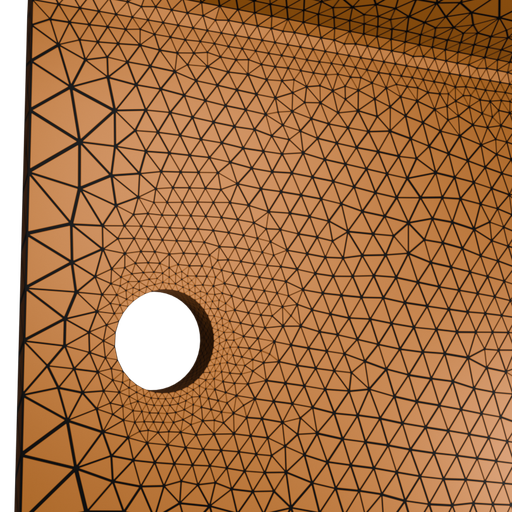}};
    \node at (3*\imagewidth, -3*\imageheight) {\includegraphics[width=\imagewidth, height=\imageheight]{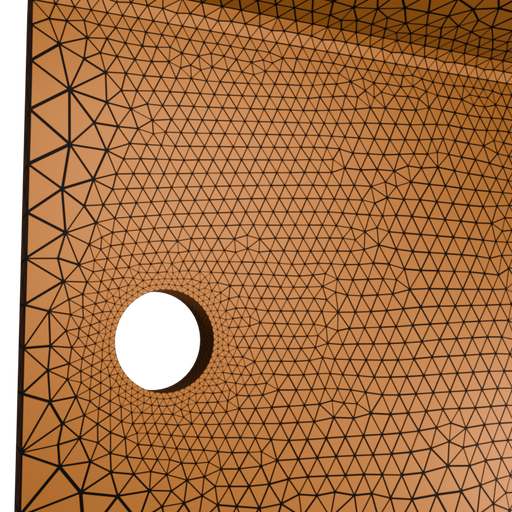}};
    \node at (4*\imagewidth, -3*\imageheight) {\includegraphics[width=\imagewidth, height=\imageheight]{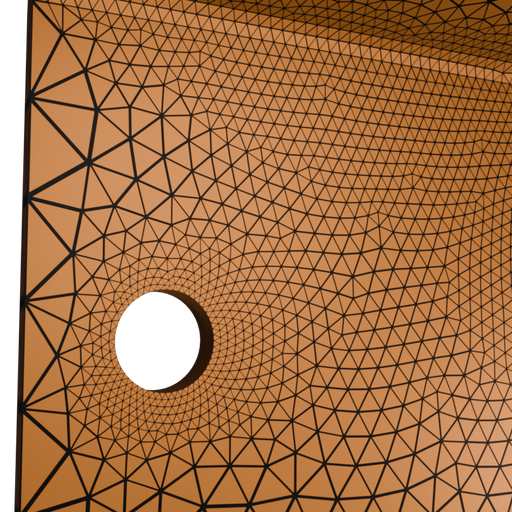}};
\end{tikzpicture}
\caption{
    Mesh generation steps of \gls{amber} (Max) for all tasks. 
    \textbf{Row 1 to 3:} Poisson's Equation with $25$/$50$/$75$ expert refinement steps. \textbf{Row 4:} Console task. 
    \gls{amber} (Max) creates generates more conservative meshes, yet still closely matches the expert after $5$ refinement steps.} 
    \label{app_fig:matrix_amber_max}
\end{figure}

\begin{figure}
    \centering
    \begin{tikzpicture}
        \def\imagewidth{0.22\textwidth}
        \def\imageheight{0.22\textwidth}
    
        \node[anchor=east] at (-0.5*\imagewidth, -0.5*\imageheight + 0.5*\imageheight) {Poisson 25};
        \node[anchor=east] at (-0.5*\imagewidth, -1.5*\imageheight + 0.5*\imageheight) {Poisson 50};
        \node[anchor=east] at (-0.5*\imagewidth, -2.5*\imageheight + 0.5*\imageheight) {Poisson 75};
        \node[anchor=east] at (-0.5*\imagewidth, -3.5*\imageheight + 0.5*\imageheight) {Console};
    
        \node[anchor=north] at (0*\imagewidth, -4*\imageheight + 1cm) {$32^d$};
        \node[anchor=north] at (1*\imagewidth, -4*\imageheight + 1cm) {$64^d$};
        \node[anchor=north] at (2*\imagewidth, -4*\imageheight + 1cm) {$256^d$};
        \node[anchor=north] at (3*\imagewidth, -4*\imageheight + 1cm) {Expert};
    
        \node at (0*\imagewidth, 0*\imageheight) {\includegraphics[width=\imagewidth, height=\imageheight]{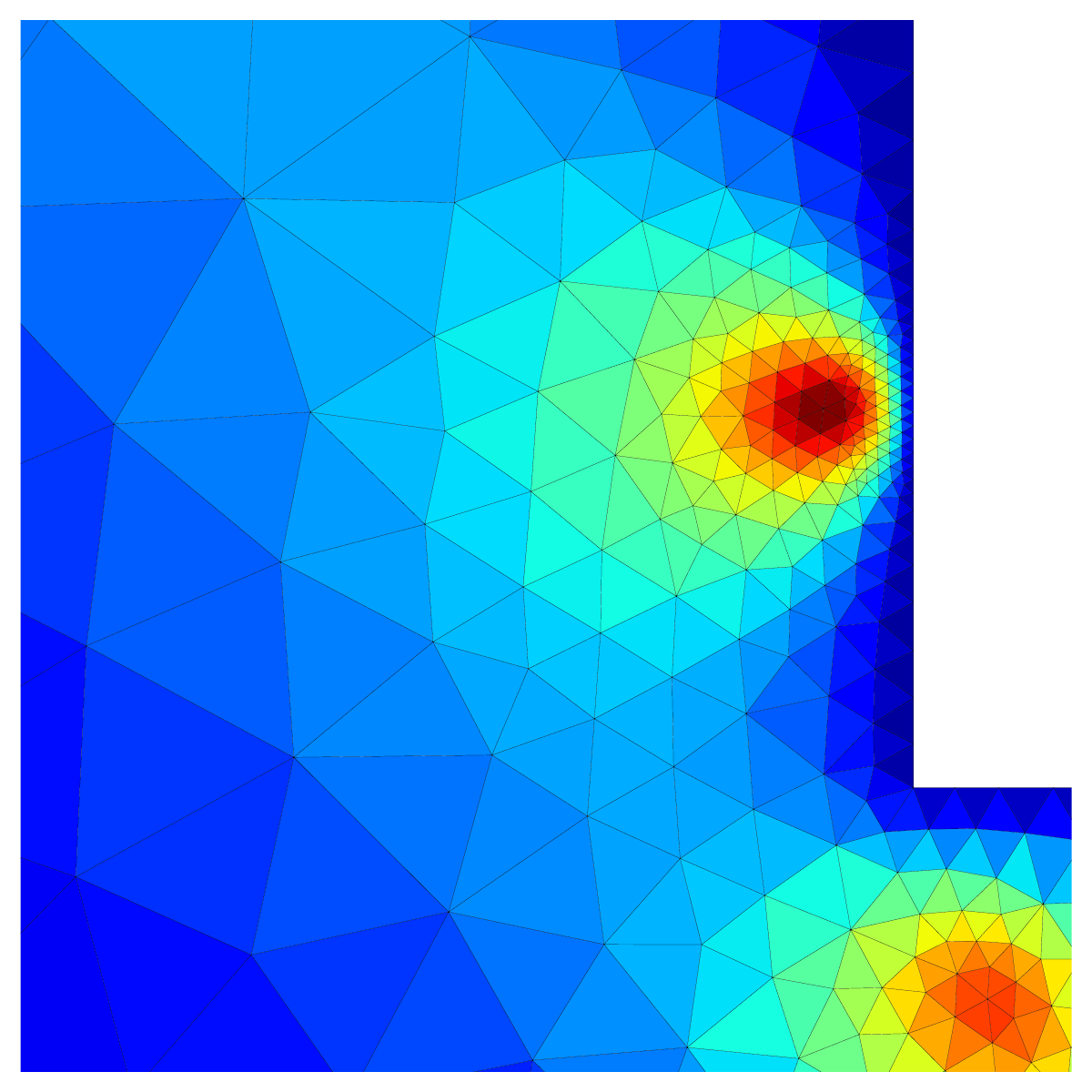}};
        \node at (1*\imagewidth, 0*\imageheight) {\includegraphics[width=\imagewidth, height=\imageheight]{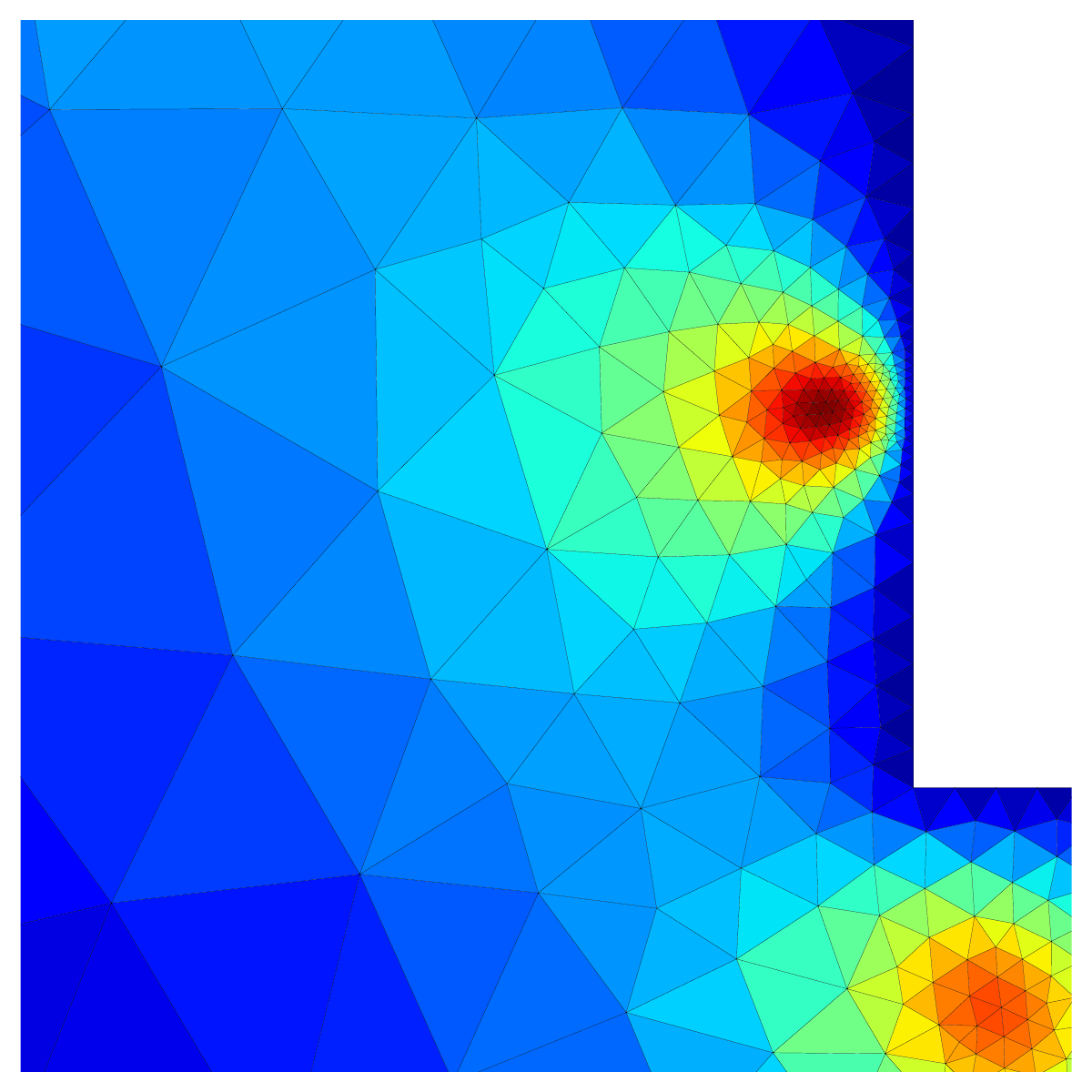}};
        \node at (2*\imagewidth, 0*\imageheight) {\includegraphics[width=\imagewidth, height=\imageheight]{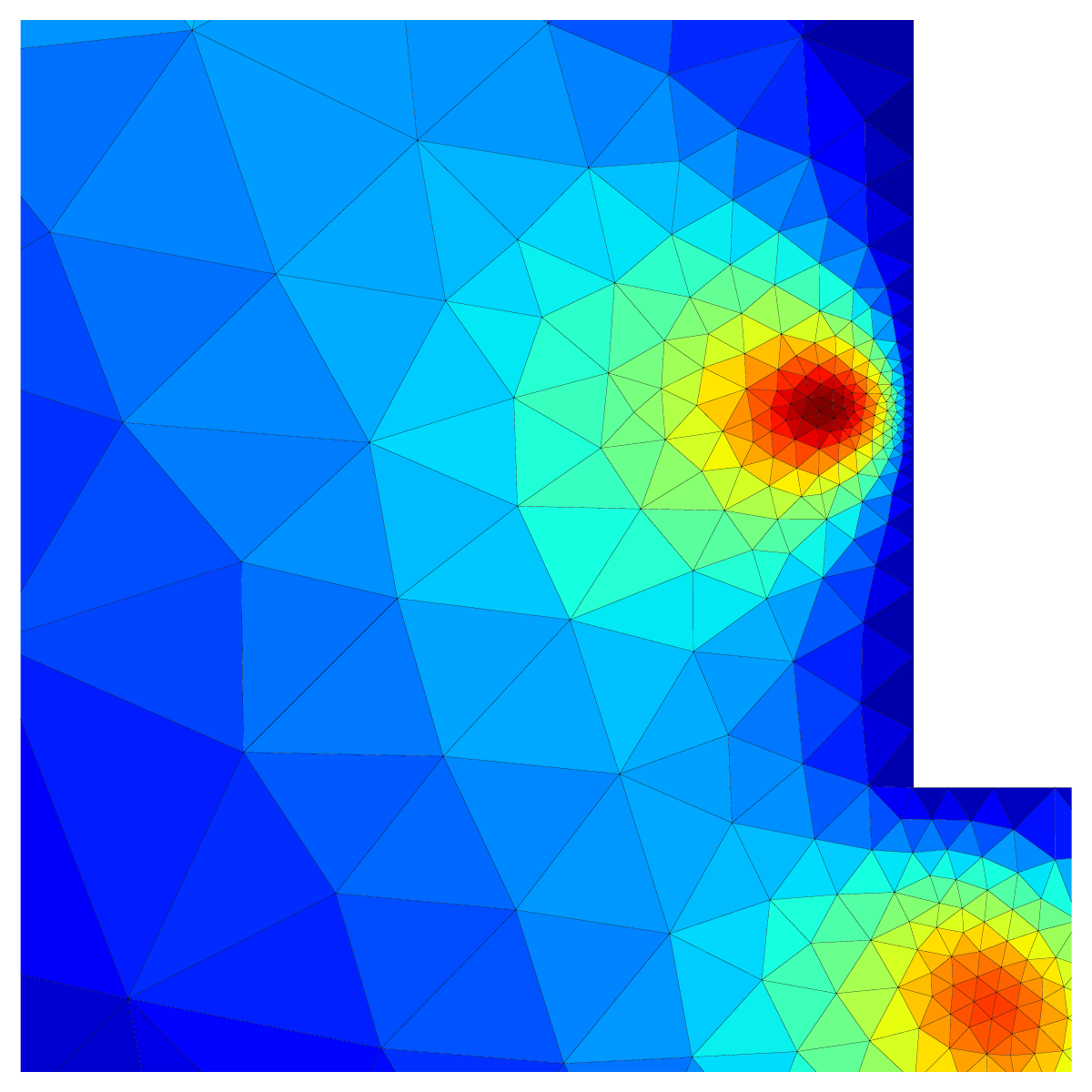}};
        \node at (3*\imagewidth, 0*\imageheight) {\includegraphics[width=\imagewidth, height=\imageheight]{03_appendix/02_figures/matrix_grids/1002/poisson_25/solution_expert.pdf}};
        
        \node at (0*\imagewidth, -1*\imageheight) {\includegraphics[width=\imagewidth, height=\imageheight]{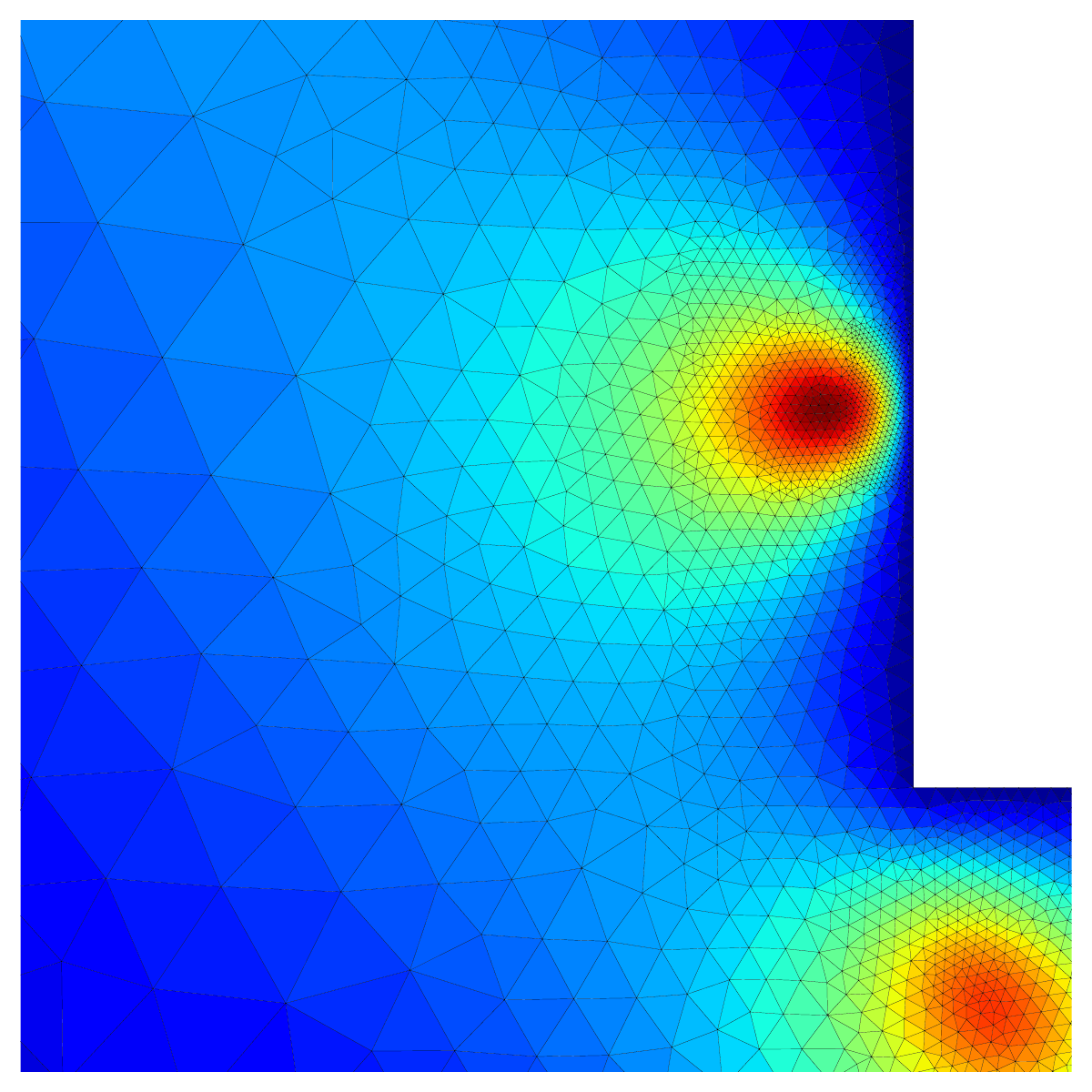}};
        \node at (1*\imagewidth, -1*\imageheight) {\includegraphics[width=\imagewidth, height=\imageheight]{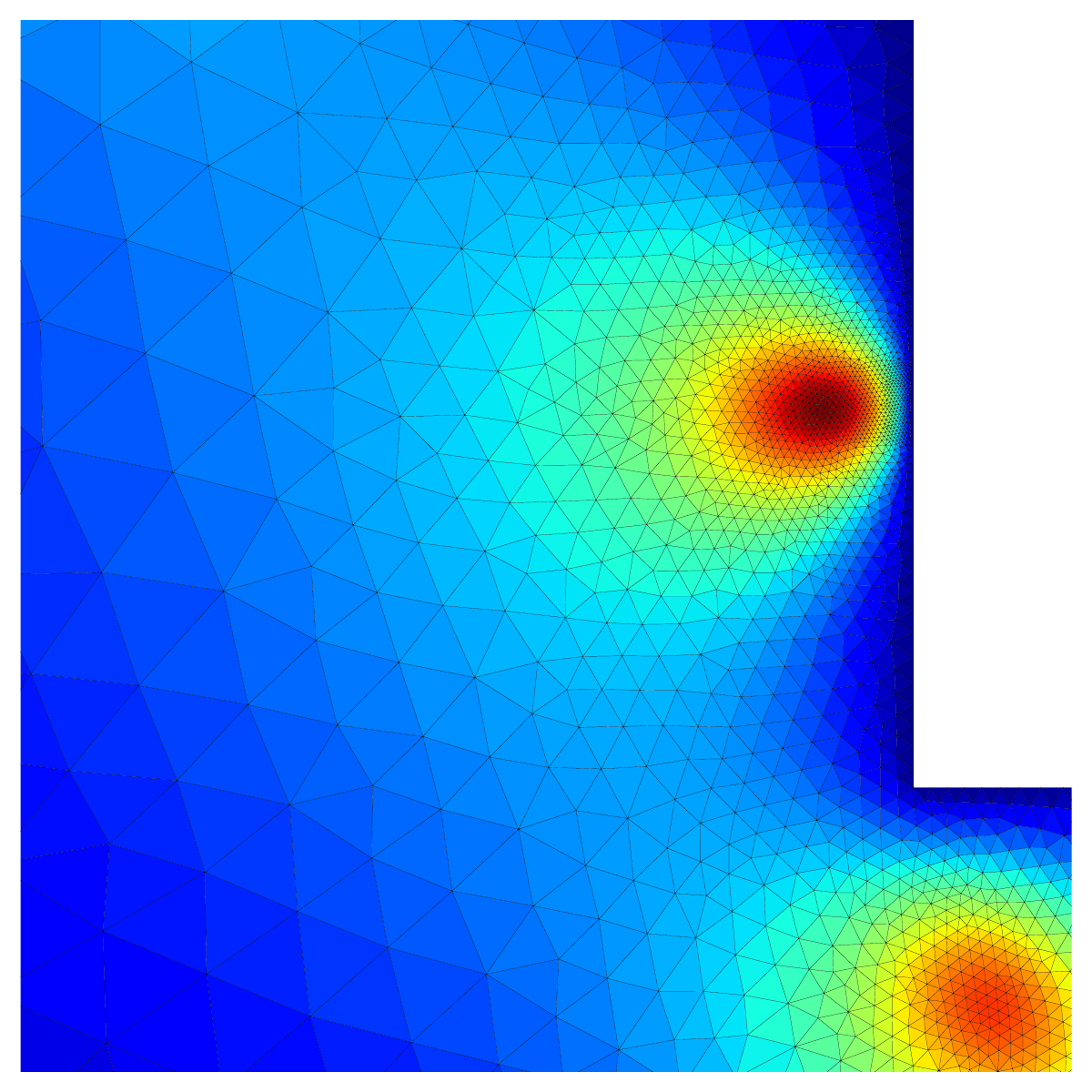}};
        \node at (2*\imagewidth, -1*\imageheight) {\includegraphics[width=\imagewidth, height=\imageheight]{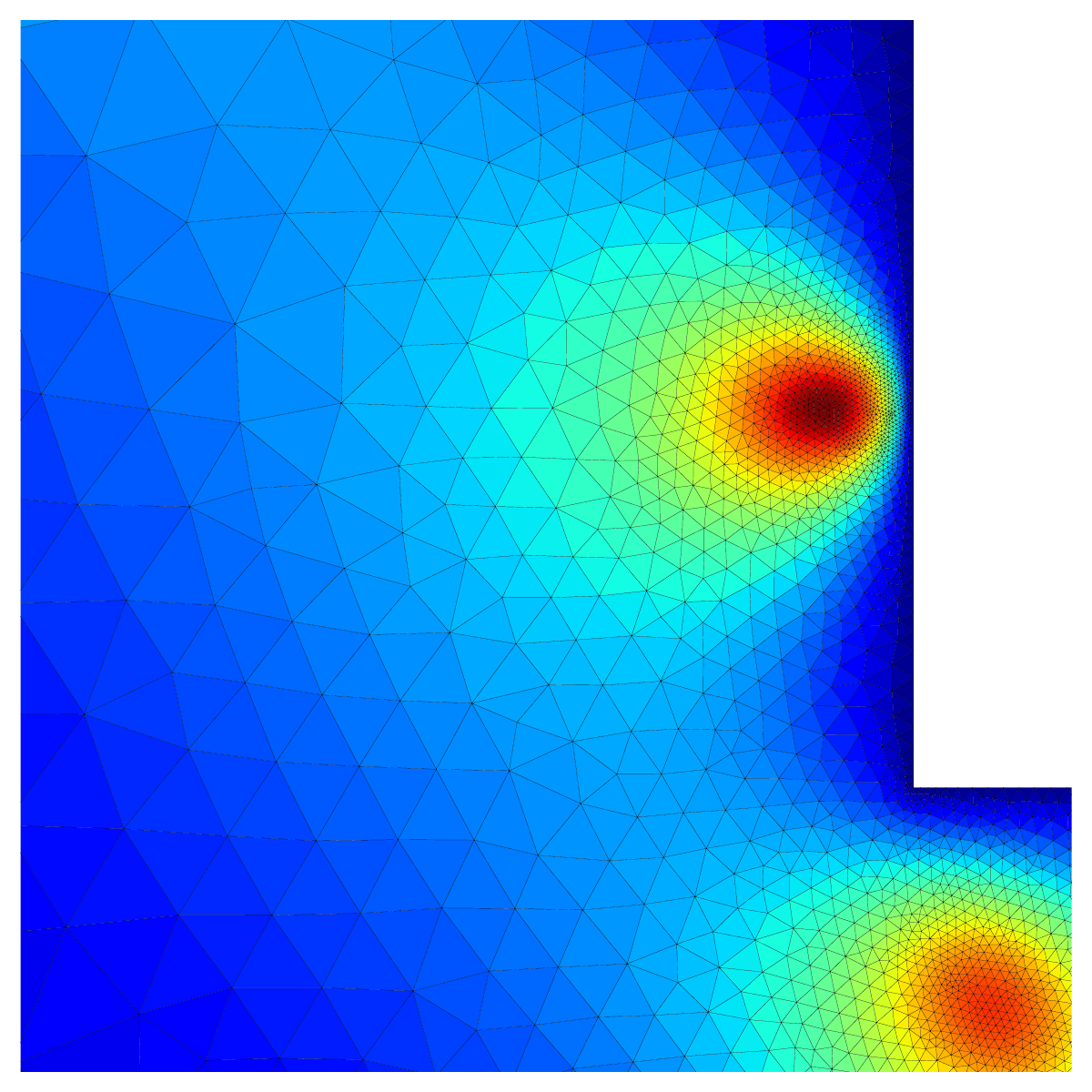}};
        \node at (3*\imagewidth, -1*\imageheight) {\includegraphics[width=\imagewidth, height=\imageheight]{03_appendix/02_figures/matrix_grids/1002/poisson_50/solution_expert.pdf}};
        
        \node at (0*\imagewidth, -2*\imageheight) {\includegraphics[width=\imagewidth, height=\imageheight]{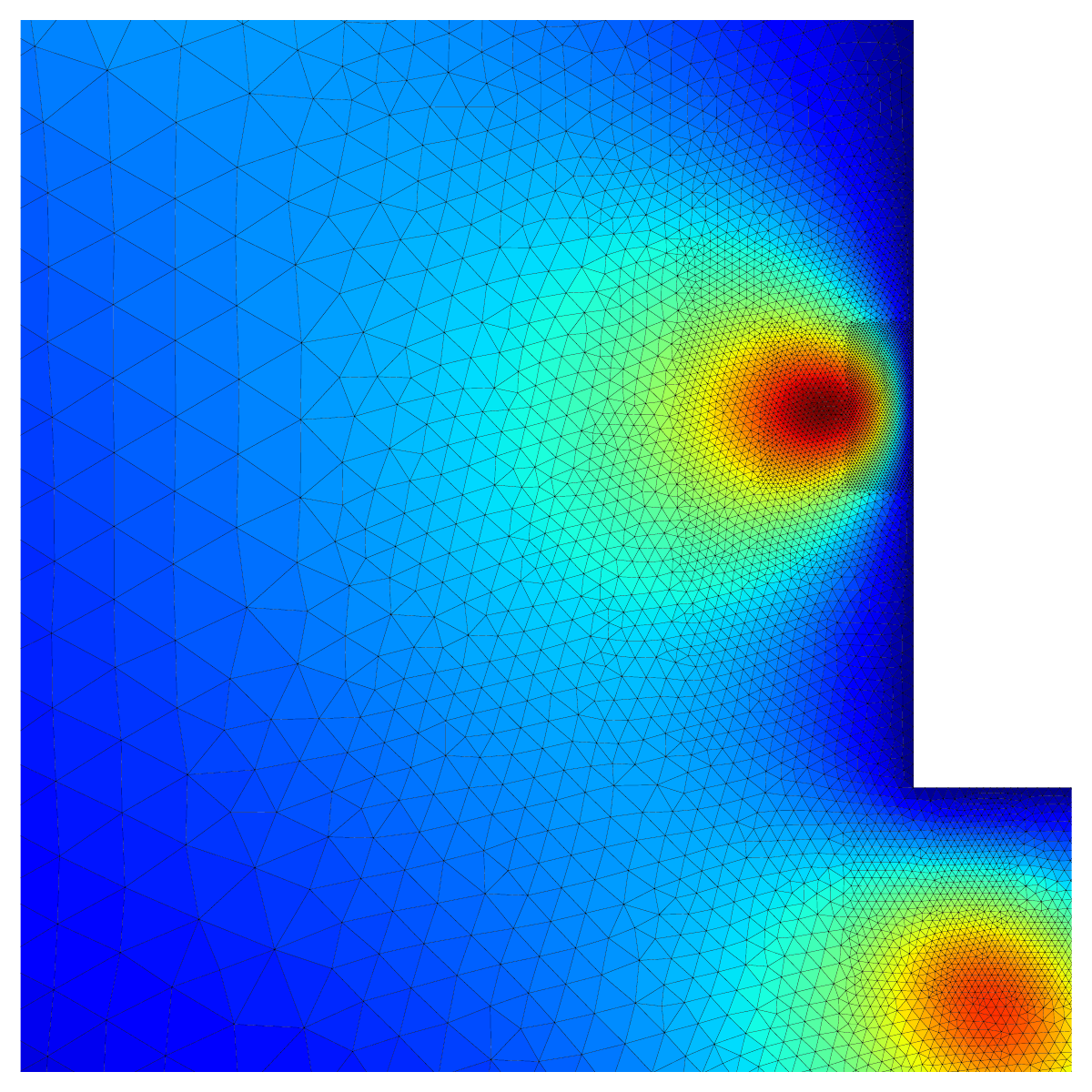}};
        \node at (1*\imagewidth, -2*\imageheight) {\includegraphics[width=\imagewidth, height=\imageheight]{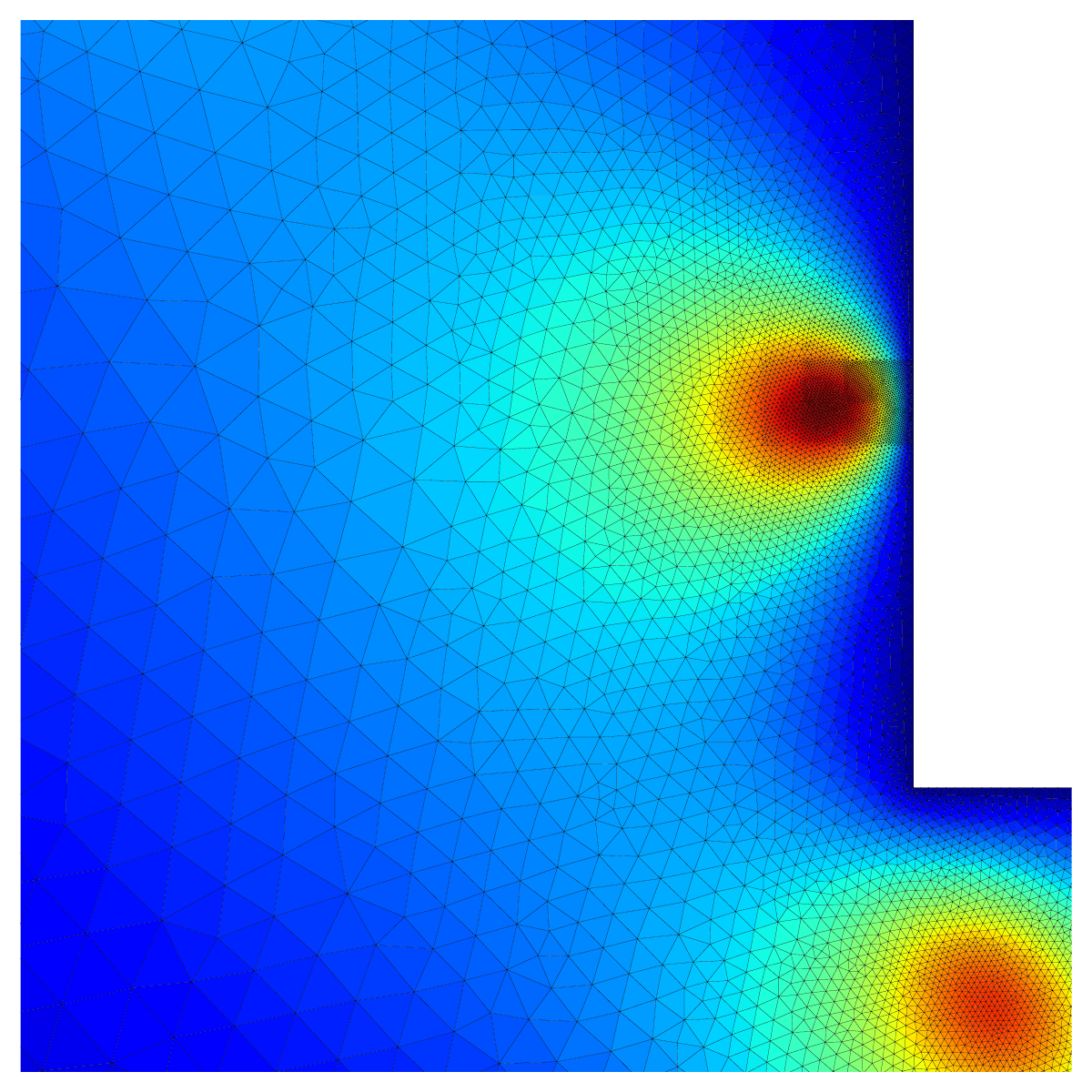}};
        \node at (2*\imagewidth, -2*\imageheight) {\includegraphics[width=\imagewidth, height=\imageheight]{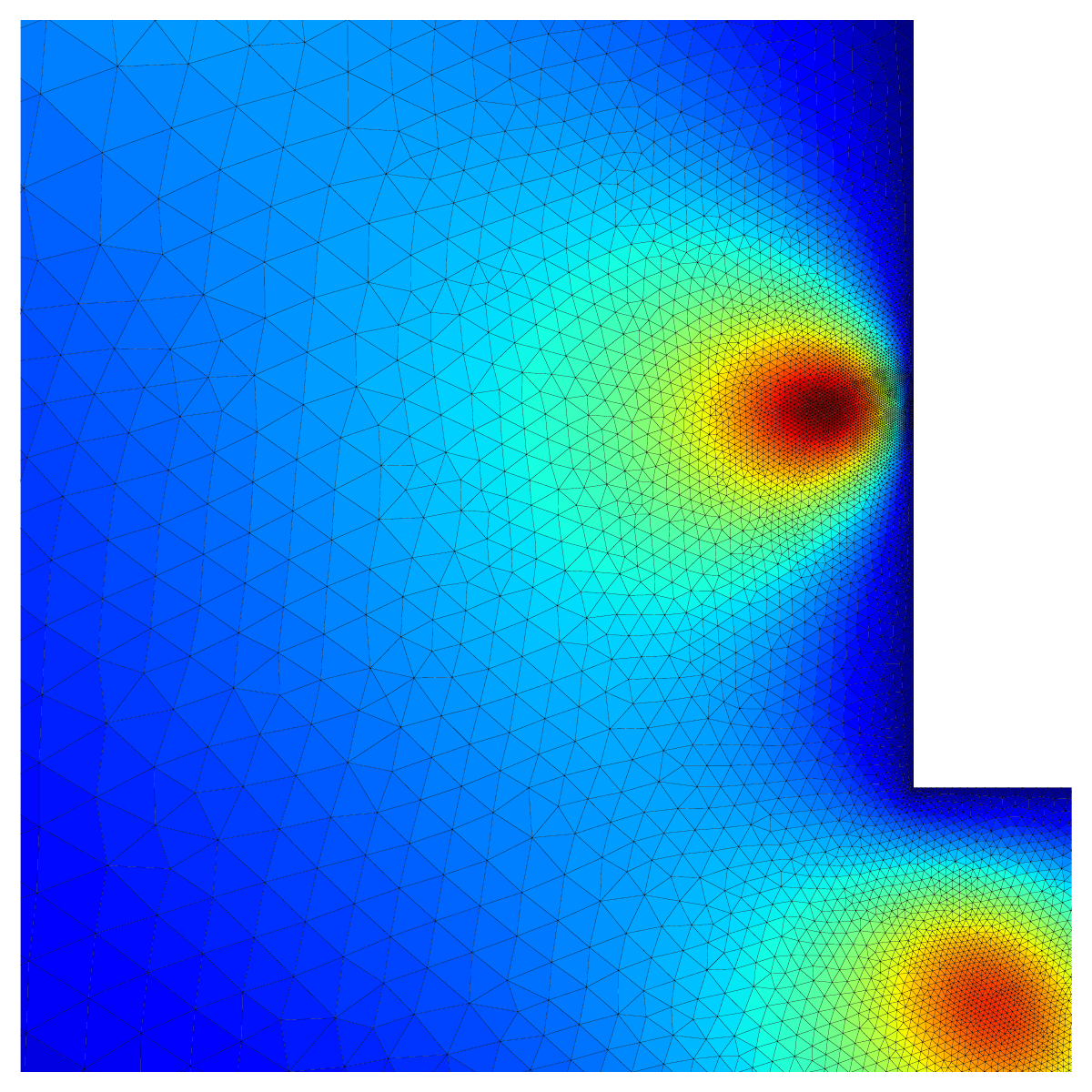}};
        \node at (3*\imagewidth, -2*\imageheight) {\includegraphics[width=\imagewidth, height=\imageheight]{03_appendix/02_figures/matrix_grids/1002/poisson_75/solution_expert.pdf}};
        
        \node at (0*\imagewidth, -3*\imageheight) {\includegraphics[width=\imagewidth, height=\imageheight]{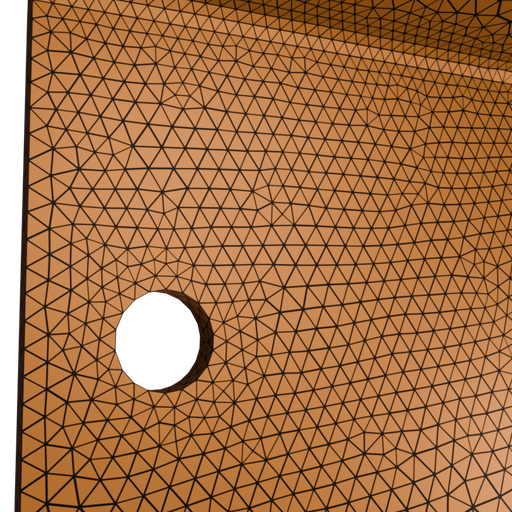}};
        \node at (1*\imagewidth, -3*\imageheight) {\includegraphics[width=\imagewidth, height=\imageheight]{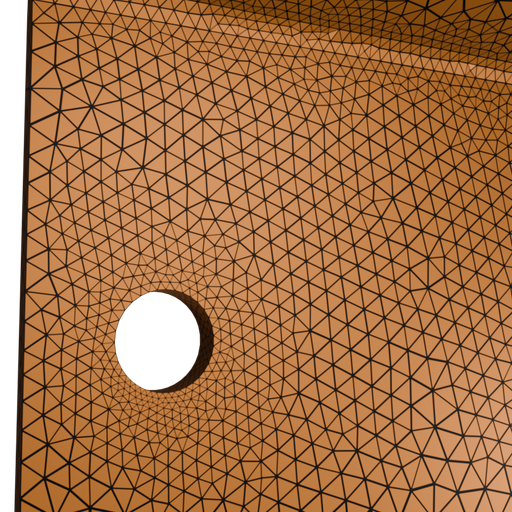}};
        \node at (3*\imagewidth, -3*\imageheight) {\includegraphics[width=\imagewidth, height=\imageheight]{03_appendix/02_figures/matrix_grids/1002/console/expert.png}};
    \end{tikzpicture}
    \caption{
    Meshes generated by the~\gls{cnn} baseline with varying input resolution on all environments. 
    \textbf{Row 1 to 3:} Poisson's Equation with $25$/$50$/$75$ expert refinement steps. 
    \textbf{Row 4:} Console task. 
    Lower image resolutions lead to square-like artifacts in the output mesh which are not present in the expert meshes. 
    The $256\times256\times256$ CNN baseline on the console task could not be computed due to excessive memory usage.}
        \label{app_fig:matrix_cnn}
\end{figure}

We present the qualitative results of all methods applied to all tasks. 
Figure \ref{app_fig:matrix_amber_mean} displays the results for \gls{amber} (Mean), Figure \ref{app_fig:matrix_amber_max} shows the results for \gls{amber} (Max), and Figure \ref{app_fig:matrix_cnn} showcases the predictions for the~\gls{cnn} baseline.

\end{document}